%% file: main_for_arxiv.tex
\documentclass[10pt,twocolumn,letterpaper]{article}

\usepackage[pagenumbers]{cvpr} %

\usepackage{graphicx}
\usepackage{amsmath}
\usepackage{amssymb}
\usepackage{booktabs}

\usepackage{extra_styles}

\usepackage{multirow}
\usepackage{array}
\usepackage{paralist}
\usepackage{comment}        %
\usepackage{bbm}
\usepackage[normalem]{ulem}
\usepackage[accsupp]{axessibility}  %
\usepackage{algorithm}
\usepackage{algorithmic}

\usepackage[pagebackref,breaklinks,colorlinks]{hyperref}
\usepackage[capitalize]{cleveref}
\crefname{section}{Sec.}{Secs.}
\Crefname{section}{Section}{Sections}
\Crefname{table}{Table}{Tables}
\crefname{table}{Tab.}{Tabs.}

\def\cvprPaperID{1264} %
\def\confName{CVPR}
\def\confYear{2022}

\begin{document}

\title{DASO: Distribution-Aware Semantics-Oriented Pseudo-label\\for Imbalanced Semi-Supervised Learning
}
\author{
  Youngtaek Oh\textsuperscript{1} \qquad Dong-Jin Kim\textsuperscript{2} \qquad In So Kweon\textsuperscript{1}\\
  \textsuperscript{1}KAIST, South Korea. \quad \textsuperscript{2}UC Berkeley / ICSI, CA.\\
  \small{
  \textsuperscript{1}\texttt{\{youngtaek.oh, iskweon\}@kaist.ac.kr}\qquad
  \textsuperscript{2}\texttt{\href{mailto:djkim93@berkeley.edu}{\color{black} djkim93@berkeley.edu}}
  }\\
}
\maketitle

\begin{abstract}
The capability of the traditional semi-supervised learning (SSL) methods is far from real-world application due to severely biased pseudo-labels caused by (1) class imbalance and (2) class distribution mismatch between labeled and unlabeled data. This paper addresses such a relatively under-explored problem. First, we propose a general pseudo-labeling framework that class-adaptively blends the semantic pseudo-label from a similarity-based classifier to the linear one from the linear classifier, after making the observation that both types of pseudo-labels have complementary properties in terms of bias. We further introduce a novel semantic alignment loss to establish balanced feature representation to reduce the biased predictions from the classifier. We term the whole framework as \textbf{D}istribution-\textbf{A}ware \textbf{S}emantics-\textbf{O}riented (DASO) Pseudo-label. We conduct extensive experiments in a wide range of imbalanced benchmarks: CIFAR10/100-LT, STL10-LT, and large-scale long-tailed Semi-Aves with open-set class, and demonstrate that, the proposed DASO framework reliably improves SSL learners with unlabeled data especially when both (1) class imbalance and (2) distribution mismatch dominate.
\end{abstract}
\vspace{-1mm}
\addtocontents{toc}{\protect\setcounter{tocdepth}{0}}
\input{main/intro}
\input{main/related}
\input{main/method}
\input{main/exp}
\input{main/conclusion}

{\small
\bibliographystyle{ieee_fullname}
\bibliography{egbib}
}
\appendix
\onecolumn
\begin{center}
    {\vspace*{-.3in}}{\vskip .375in}
    {\Large \bf
        Supplementary Materials for \\DASO: Distribution-Aware Semantics-Oriented Pseudo-Label\\for Imbalanced Semi-Supervised Learning\par
    }
\end{center}
{
  \hypersetup{linkcolor=black}
  \tableofcontents
}
\clearpage
\addtocontents{toc}{\protect\setcounter{tocdepth}{2}}
\input{supple/notations}
\input{supple/algorithm}
\input{supple/detailed_setup}
\input{supple/additional_exp}
\input{supple/detailed_analysis}
\input{supple/overall_framework}

\end{document}

%% file: main/intro.tex
\section{Introduction}
\label{sec:intro}
Semi-supervised learning (SSL)~\cite{chapelle2009semi} has shown to be promising for leveraging unlabeled data to reduce the cost of constructing labeled data~\cite{berthelot2019mixmatch,berthelot2020remixmatch,kuo2020featmatch,sohn2020fixmatch,li2020comatch} and even boost the performance at scale~\cite{kim2019image,yalniz2019billion,xie2020self,pham2020meta}. 
The common approach of these algorithms is to produce \emph{pseudo-labels} for unlabeled data based on model’s predictions and utilize them for regularizing model training~\cite{kim2019image,lee2013pseudo,sohn2020fixmatch}. 
Although adopted in a variety of tasks, these algorithms often assume class-balanced data, while many real-world datasets exhibit \emph{long-tailed} distributions~\cite{bengio2015sharing,dong2018imbalanced,kim2020detecting,kim2021acp++}. 
With class-imbalanced data, the class distribution of pseudo-labels from unlabeled data becomes severely biased to the majority classes due to confirmation bias~\cite{arazo2020pseudo}. Such biased pseudo-labels can further bias the model during training.

\begin{figure}[t]
  \centering
   \includegraphics[width=0.95\columnwidth]{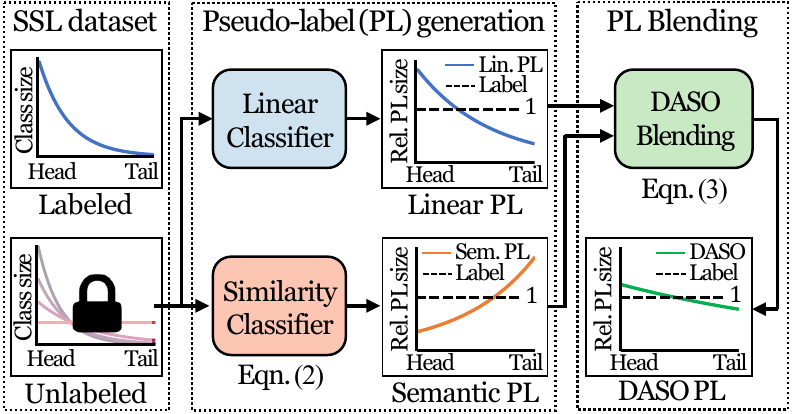}
   \caption{Glimpse of the DASO framework. 
   DASO reduces the overall bias in pseudo-labels (PL) from unlabeled data by blending two complementary PLs from different classifiers.
   Note that bias is conceptually illustrated as relative PL size (Rel. PL size), meaning that pseudo-label size is normalized by actual label size.
   }
   \label{fig:overall_daso}
   \vspace{-3mm}
\end{figure}
Many methods of handling class-imbalanced labels have been proposed in the supervised learning community, but little interest has been made in re-balancing pseudo-labels in SSL.  
Recent studies have explored this \emph{imbalanced SSL} setting, where as a reference to the class distribution of unlabeled data, it is often assumed that it is the same as the class distribution of labels~\cite{kim2020distribution,wei2021crest}, or a separate distribution estimate is required~\cite{kim2020distribution}.
However, the actual class distribution of unlabeled data is unknown without the labels.
For example, unlabeled data may have large class distribution gap from labeled data, including many samples in novel classes not defined in the label set~\cite{su2021semisupervised}. 
As we elaborate in~\cref{sec:exp}, the bias of pseudo-labels also depends on such class distribution mismatch between labeled and unlabeled data, and 
using inaccurate estimates or wrong assumptions about the unlabeled data cannot be helpful under imbalanced SSL. 

In this work, we present a new imbalanced SSL method specifically tailored for alleviating the bias in pseudo-labels under class-imbalanced data, while discarding the common assumption that the class distribution of unlabeled data is the same with the label distribution.
To this end, as shown in \cref{fig:overall_daso}, we observe that semantic pseudo-labels~\cite{han2020unsupervised} obtained from a similarity-based classifier~\cite{snell2017prototypical} are biased towards minority classes as opposed to linear classifier-based pseudo-labels~\cite{lee2013pseudo,sohn2020fixmatch} being biased towards head classes.
As illustrated in~\cref{sec:mtd_bias}, we draw the key inspiration from those complementary properties of two different types of pseudo-labels to develop a new pseudo-labeling scheme. 

In this regard, we introduce a generic imbalanced SSL framework termed Distribution-Aware Semantics-Oriented (DASO) Pseudo-label in~\cref{sec:mtd_daso}.
Building upon the existing SSL learner, we propose to blend the linear and semantic pseudo-labels in different proportions for each class to reduce the overall bias.
This blending strategy can provide a more balanced supervision than simply using either of the pseudo-label.
The primary novelty comes from the scheduling of the weights for mixing the pseudo-labels.
Specifically, we dynamically adjust the relative weights of semantic pseudo-labels to be blended so that linear pseudo-labels are less biased according to the current class distribution of pseudo-labels.
By virtue of such mechanism, without resorting to any class priors for the unlabeled data, DASO reliably brings performance gain even with substantial class distribution mismatch between labeled and unlabeled data.

We further propose a simple yet effective semantic alignment loss to establish balanced feature representation via \emph{balanced} class prototypes,
which is the extension of the consistency regularization framework in \cite{xie2019unsupervised,sohn2020fixmatch} onto feature space. 
We align the unlabeled data onto each of the similar prototypes, by consistently assigning two different views of an unlabeled sample in \emph{feature space} to the same prototype.
These enhanced feature representations not only help linear classifier produce less biased predictions, but can also be reused for semantic pseudo-labels from similarity-based classifier.
We validate the semantic alignment loss is useful under imbalanced SSL, especially helpful for DASO.

The efficacy of DASO is extensively justified with the imbalanced versions of benchmarks: 
CIFAR-10/100~\cite{krizhevsky2009learning} and STL-10~\cite{coates2011analysis} in~\cref{sec:exp}. 
We even test DASO with large-scale long-tailed Semi-Aves~\cite{su2021semisupervised} with open-set classes in unlabeled data, closely related to real-world scenarios. 
As such, DASO consistently benefits under various distributions of unlabeled data and degrees of imbalance, demonstrating to be a truly generic framework that works well on top of diverse frameworks such as existing SSL learners and even other re-balancing frameworks for labels and SSL.

The key contributions in our work can be summarized as follows:
(1) We propose a novel pseudo-labeling framework, DASO, for debiasing pseudo-labels by class-adaptively blending two complementary types of pseudo-labels observing current class distribution of pseudo-labels.
(2) DASO introduces semantic alignment loss to further alleviate the bias from high-quality feature representation, by aligning each unlabeled example to the similar prototype. 
(3) DASO readily integrates with other frameworks to show significant performance improvements under diverse imbalanced SSL setup, including the most practical scenario.  

%% file: main/related.tex
\section{Related Work}
\label{sec:related}
\noindent{\textbf{Class-imbalanced learning.}} 
Datasets that well capture the dynamic nature of \emph{real-world} exhibit \textit{class-imbalanced}, or \textit{long-tailed}  distributions~\cite{van2018inaturalist,gupta2019lvis}. Learning on such datasets has been a great challenge to deep neural networks, since they cannot generalize well to the rare classes~\cite{bengio2015sharing}. Conventional approaches to combat the imbalance include data re-sampling~\cite{chawla2002smote,ando2017deep,kim2020m2m}, cost-sensitive re-weighting~\cite{cui2019class,cao2019learning,park2021influence}, and decoupling the representation and the classifier~\cite{Kang2020Decoupling,zhou2020bbn}. Recently, learning expert models across classes~\cite{xiang2020learning,wang2021longtailed} and re-balancing with the data distribution in loss computation phase~\cite{ren2020balanced,menon2021longtail,hong2021disentangling} are also shown to be effective.
On the other hand, \cite{liu2020selectnet,yang2020rethinking} leveraged unlabeled data for class-imbalanced learning.
Unlike all the aforementioned methods, we focus on alleviating the bias of pseudo-labels in semi-supervised learning due to class imbalanced labels and distribution mismatch between labeled and unlabeled data.

\noindent{\textbf{Semi-supervised learning (SSL).}}
SSL aims to learn from both labeled and unlabeled data. For unlabeled data, SSL generates targets (\eg, pseudo-labels) from model predictions via \textit{pseudo-labeling}~\cite{kim2019image,lee2013pseudo}, \textit{consistency regularization}~\cite{tarvainen2017mean,miyato2018virtual}, and combinations of them~\cite{berthelot2019mixmatch, berthelot2020remixmatch,kim2018disjoint,kuo2020featmatch} under \textit{cluster assumption}~\cite{chapelle2009semi}. 
However, pseudo-labels can be biased with class-imbalanced data~\cite{kim2020distribution}, which harm the model when utilized.
Some works deal with such issue via loss re-weighting~\cite{hyun2020class,lee2021auxiliary,kim2019image}, optimization~\cite{kim2020distribution}, data re-sampling~\cite{wei2021crest}, and meta-learning sample importance~\cite{ren2018learning,ren2020not}. 
However, class distribution of unlabeled data either unknown or different from the labeled one
can also exacerbate the bias, limiting the applicability of such methods. 
In this aspect, we devise a new pseudo-labeling method that handles such challenging but practical scenarios.

%% file: main/method.tex
\section{Proposed Method}
\label{sec:method}
\begin{figure*}[!t]
\centering
\subfloat[][Recall of pseudo-labels.]{
    \includegraphics[width=.27\linewidth]{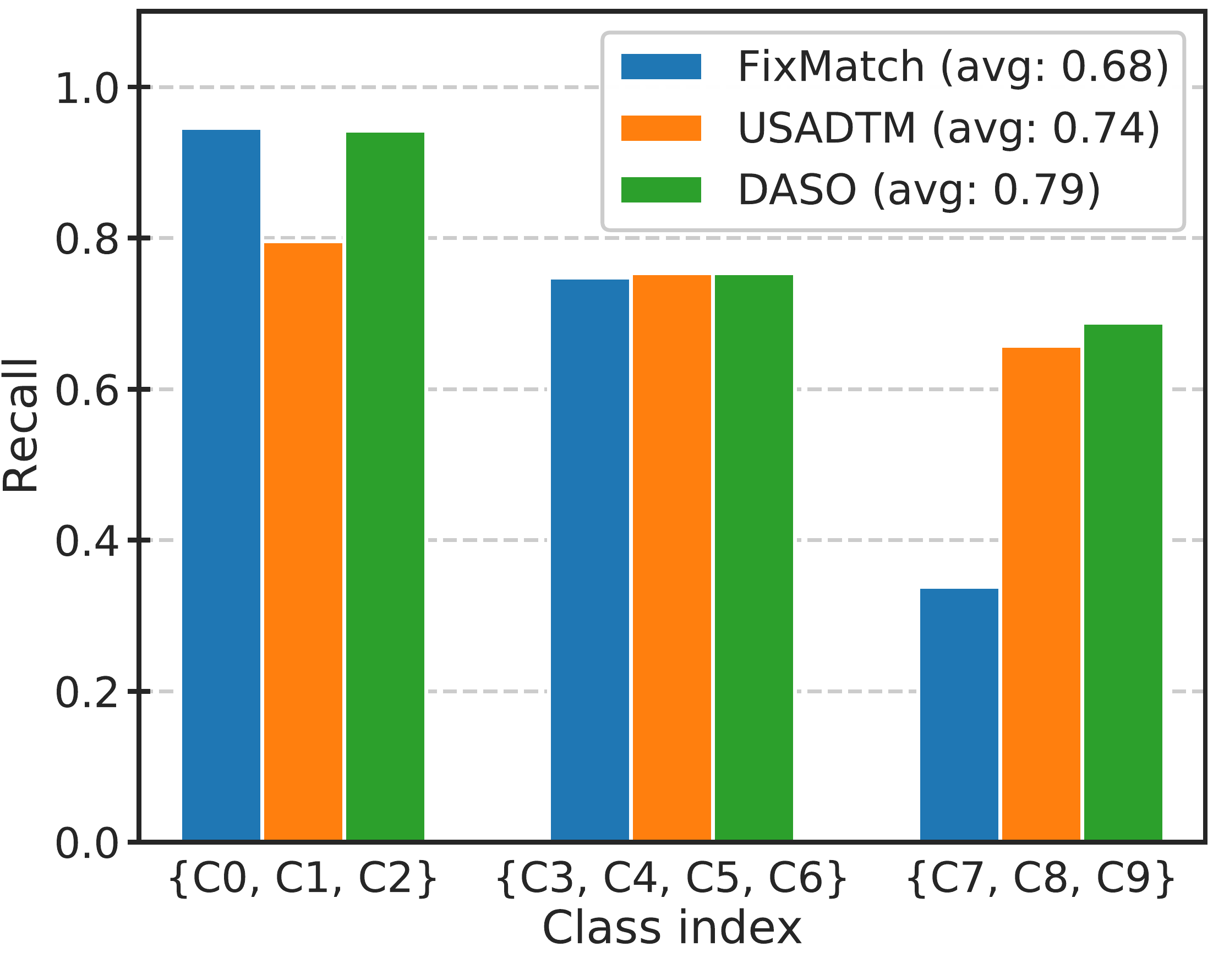}
    \label{fig:bias1_recall}
}\qquad
\subfloat[][Precision of pseudo-labels.]{
    \includegraphics[width=.27\linewidth]{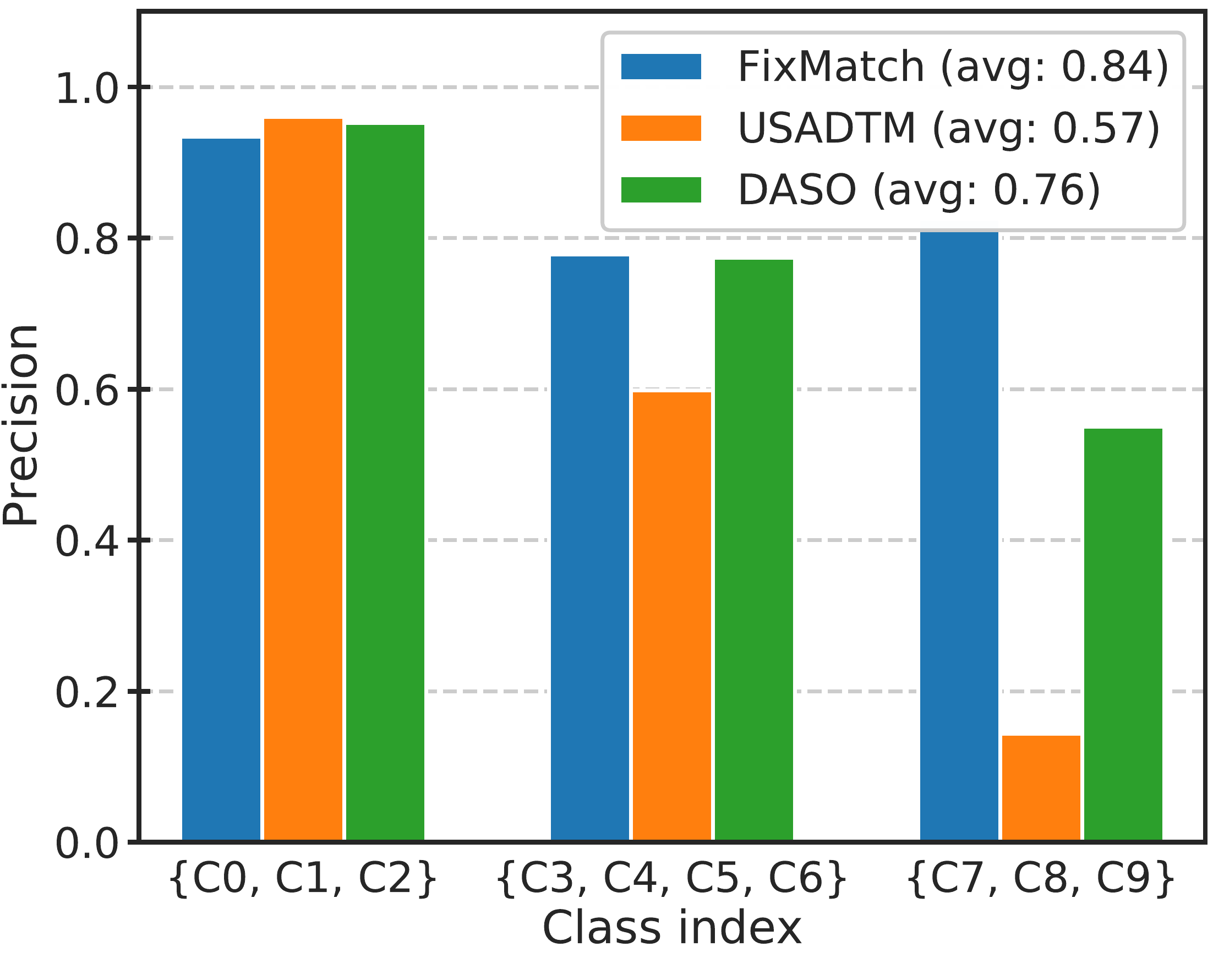}
    \label{fig:bias1_prec}
}\qquad
\subfloat[][Class-wise test accuracy.]{
    \includegraphics[width=.27\linewidth]{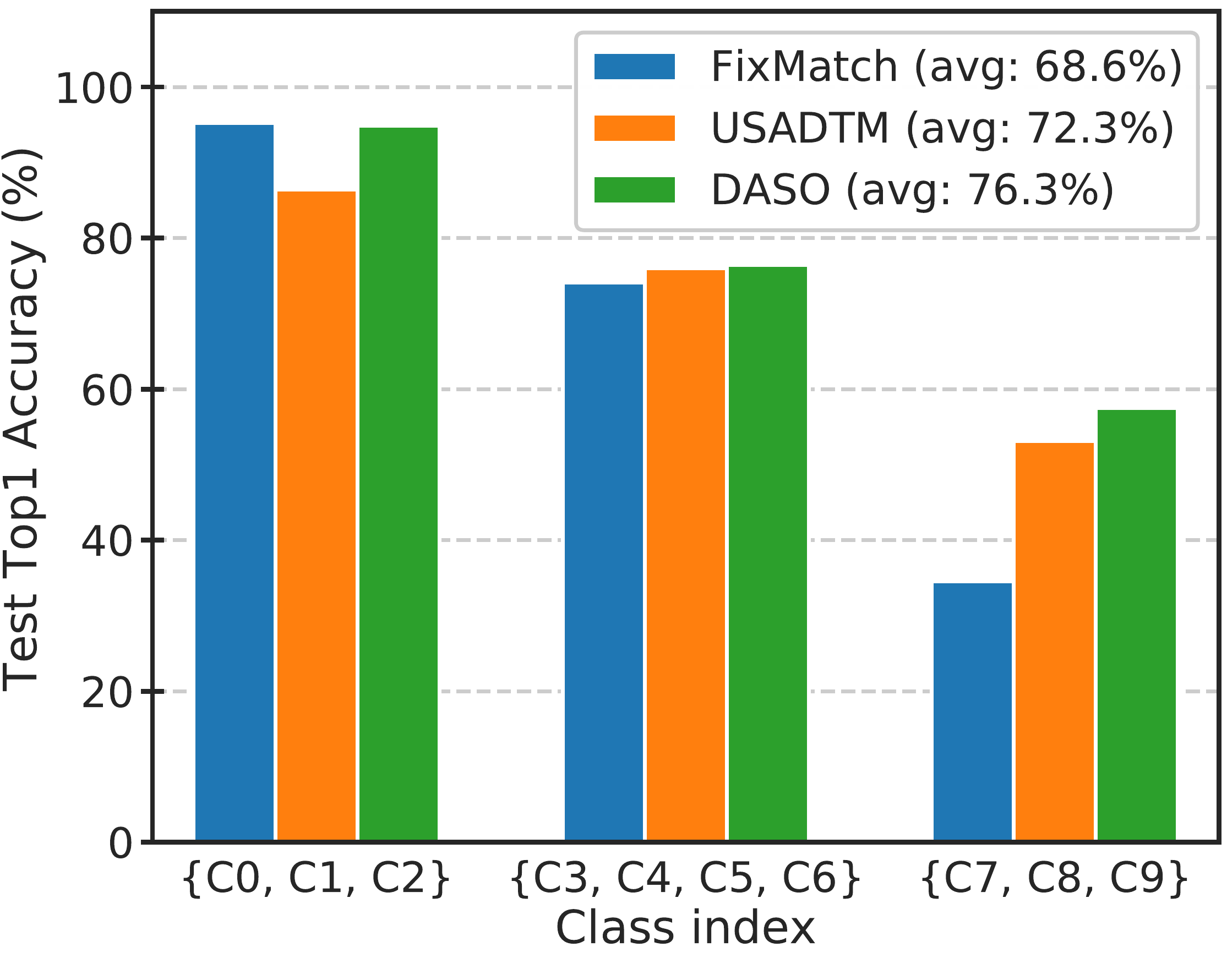}
    \label{fig:bias1_test}
}
\vspace{-2mm}
\caption{
Analysis on recall and precision of pseudo-labels and the corresponding test accuracy.
Note that the class index from x-axis is sorted by the class size; \texttt{C0} and \texttt{C9} are the head and tail classes, respectively. 
Although USADTM~\cite{han2020unsupervised} improves the recall of minority classes, the precision of those classes is significantly reduced. In contrast, DASO improves the recall of minority classes while sustaining the precision, which leads to higher test accuracy of those classes. More analyses with various SSL methods are provided in~\cref{sec:sup_rec_prec_analysis}.
}
\label{fig:pl_bias1}
\vspace{-4mm}
\end{figure*}

\subsection{Preliminaries}
\label{sec:mtd_setup}
\noindent \textbf{Problem setup.}
We consider $K$-class semi-supervised image classification that leverages both labeled data $\cX=\{ (x_{n}, y_{n}) \}^{N}_{n=1}$ and unlabeled data $\cU=\{ u_{m} \}^{M}_{m=1}$ to train a model $f$.
Note that the model $f = f_\phi^\text{cls} \circ f^{\text{enc}}_\theta$ consists of a feature encoder $f^{\text{enc}}_\theta$ followed by a linear classifier $f_\phi^\text{cls}$, where $\theta$ and $\phi$ are the set of parameters of $f^{\text{enc}}_\theta$ and $f_\phi^\text{cls}$.
The input image $x$ is paired with the label $y$ to learn $\cL_{\text{cls}}$ (\eg, cross-entropy) from the prediction $f(x)$. 
For the unlabeled data, a pseudo-label\footnote{In this work, we assume it includes both one-hot form and soft form cases: $\Sigma_k \hat{p}_k = 1$ where $\hat{p}_k \in \left[0, 1\right]$.} $\hat{p} \in \Rbb^K$ is assigned to learn the unsupervised loss $\cL_{u}=\Phi_{u} \left(\hat{p}, f(u) \right)$, where $\Phi_{u}$ can be implemented via entropy~\cite{grandvalet2005semi} or consistency regularization~\cite{laine2016temporal,tarvainen2017mean}, depending on the SSL learner.

For FixMatch~\cite{sohn2020fixmatch} as an example, the pseudo-label $\hat{p}=\textnormal{OneHot}\left(\argmax_k p^{{(w)}}_k \right)$ with $p^{{(w)}}=f\left(\cA_w(u) \right)$ provides the target for the prediction $p^{{(s)}}=f\left(\cA_s(u) \right)$ with some confident ones to the cross-entropy loss $\cH$ as follows: 
\begin{equation}
    \label{eqn:fixmatch}
    \Phi_u(\hat{p},\,p^{{(s)}}) = \mathbbm{1}\left(\max_k p^{{(w)}}_k \geq \tau\right)\,\cH \left(\hat{p},\,p^{{(s)}} \right), 
\end{equation}
where $\cA_w$ and $\cA_s$ correspond to weak augmentation (\eg, random flip and crop) and advanced augmentation (\eg, RandAugment~\cite{cubuk2020randaugment} followed by Cutout~\cite{devries2017improved}), respectively. 

\noindent \textbf{Imbalanced semi-supervised learning.}
Let us denote $N_k$ and $M_k$ as the number of labeled and unlabeled examples respectively in class $k$. 
The degree of imbalance for each data is characterized by the imbalance ratio, $\gamma_l$ or $\gamma_u$, where we assume $\gamma_l=\frac{\max_{k} N_k} {\min_{k} N_k} \gg 1$ under imbalanced SSL. $\gamma_u$ is specified in the same way using the actual labels without access during training. It is worth noting that the class distribution of $\cU$ (\eg, $\gamma_u$) may be either similar to $\cX$, or significantly divergent in practice, and such varying distributions greatly affect the SSL performances with the same $\cX$ as shown in~\Cref{tab:other_alg}.
In this regard, our goal is to produce debiased pseudo-labels with class-imbalanced data, while maintaining the performances of SSL algorithms with various, but still \emph{unknown} class distribution of unlabeled data. 

\subsection{Motivation}
\label{sec:mtd_bias}
\noindent{\textbf{Linear and semantic pseudo-label.}} 
Pseudo-labeling based on linear classifier (\ie, fc layer), which has been widely adopted by pseudo-label-based algorithms~\cite{cho2021dealing,kim2018disjoint,kim2020detecting,kim2021acp++} especially for SSL~\cite{lee2013pseudo,sohn2020fixmatch}, %
can produce biased pseudo-labels towards majority classes with class-imbalanced data. 
We abbreviate this type of pseudo-labels as linear pseudo-labels.
Instead, pseudo-labels can be obtained from similarity-based classifier~\cite{datta1997symbolic, salakhutdinov2007learning} by measuring the similarity of a given representation (\eg, prototypes~\cite{snell2017prototypical}) to an unlabeled sample in feature space, which we call simply \emph{semantic pseudo-labels}. As note, similarity-based classifier has been widely adopted for reducing biased predictions~\cite{rebuffi2017icarl,Kang2020Decoupling,li2021mopro}.
In SSL, USADTM~\cite{han2020unsupervised} utilizes semantic pseudo-labeling method. %
As following, %
we conduct a simple experiment to explore each aspect of linear and semantic pseudo-labels. %

\noindent{\textbf{Trade-offs between linear and semantic pseudo-label.}} 
As shown in~\cref{fig:pl_bias1}, we compare FixMatch~\cite{sohn2020fixmatch} and USADTM~\cite{han2020unsupervised} using linear and semantic pseudo-label respectively, under \emph{imbalanced} SSL setup.
From \cref{fig:bias1_recall,fig:bias1_prec}, the linear pseudo-labels from FixMatch achieve high recall in majority classes while low recall but high precision in the minorities, suggesting that actual minority class examples are biased towards head classes. 
In contrast, for semantic pseudo-labels from USADTM, the actual majorities are biased towards minority classes. This is because the precision of tail classes has decreased significantly in~\cref{fig:bias1_prec}, while the recall has increased in sacrifice of the recall from head classes in~\cref{fig:bias1_recall}.
Comparing the test accuracy from~\cref{fig:bias1_test}, 
USADTM shows relatively increased overall test accuracy compared to FixMatch by virtue of more abundant minority pseudo-labels, while losing the accuracy on the head. In other words, the overall increase in accuracy is limited when only using semantic pseudo-labels. 
We provide two lessons from the simple experiment in \cref{fig:pl_bias1}, as summarized by:
{\setdefaultleftmargin{3mm}{}{}{}{}{}
\begin{enumerate}
  \item Semantic pseudo-labels are \emph{reversely biased towards the tail side}, which lead to the limited accuracy gain.
  \item The linear and semantic pseudo-labels
  have the \emph{complementary properties} useful for reducing the overall bias.
\end{enumerate}
}
These empirical findings motivate us to exploit the linear and semantic pseudo-labels \emph{differently} in different classes for debiasing.
For example, as the linear pseudo-label for a sample $u$ points to the majorities, more semantic pseudo-label component should contribute to the final pseudo-label to prevent the false positives towards the head, and the vice versa when the linear pseudo-label predicts $u$ as minority. 

We also present the result of our solution, DASO, in \cref{fig:pl_bias1}, %
where the recall of the final pseudo-label has increased but the overall pseudo-labels are still not biased towards the minority classes, unlike USADTM. 
Thanks to such unbiased pseudo-labels between the head and tail classes obtained by properly blending two pseudo-labels, the overall test accuracy also increased a lot from~\cref{fig:bias1_test}.

\subsection{DASO Pseudo-label Framework}
\label{sec:mtd_daso}
We propose DASO, a generic framework for imbalanced SSL with two novel contributions as (1) distribution-aware blending for the linear and semantic pseudo-labels %
and (2) semantic alignment loss%
, which are described as follows.

\noindent{\textbf{Framework overview.}}
Without loss of generality, we consider DASO built on top of FixMatch~\cite{sohn2020fixmatch} for convenience in notations, while DASO can easily integrate with other SSL learners as shown in~\Cref{tab:main,tab:other_alg}.
First, the linear and semantic pseudo-label, $\hat{p}$ and $q^{(w)}$ are produced with a feature $z^{(w)}=f^{\text{enc}}_{\theta}(\cA_w(u))$ from the linear and similarity-based classifier, respectively. 
Then the final pseudo-label $\hat{p}^{\prime}$ is obtained from the distribution-aware blending process using $\hat{p}$ and $q^{(w)}$, and it provides the target to $\cL_u=\Phi_u(\hat{p}^{\prime},\,p)$ instead of linear pseudo-label in the existing SSL learner. 
In case of FixMatch, the prediction of $u$ corresponds to $p=p^{(s)}=f(\cA_{s}(u))$.  
For the semantic alignment loss, the semantic pseudo-label $q^{(w)}$ provides the target for $q^{(s)}$ to the cross-entropy, where $q^{(s)}$ is the result of the similarity-based classifier with $z^{(s)}=f^{\text{enc}}_{\theta}(\cA_s(u))$.
Note that we denote $q^{(w)}$ as $\hat{q}$ for simplicity, unless confusion arises.

\noindent{\textbf{Balanced prototype generation.}}
To execute a similarity-based classifier for obtaining the semantic pseudo-label, we first build a set of class prototypes $\mC=\{c_{k} \}^{K}_{k=1}$ from $\cX$, similar to~\cite{han2020unsupervised}. %
In detail, we build a dictionary of memory queue $\mQ=\{Q_{k}\}^{K}_{k=1}$ where each key corresponds to the class and $Q_k$ denotes a memory queue for class $k$ with the fixed size $|Q_k|$. 
The class prototype $c_k$ for every class $k$ is efficiently calculated by averaging the feature points in the queue $Q_k$,
where we update $Q_k$ for all $k$ at every step by pushing new features from labeled data in the batch and discarding the most old ones when $Q_k$ is full. %

The prototype representation can also be imbalanced using class-imbalanced labeled data. 
To prevent such biased prototypes, we additionally propose \emph{balancing} the prototypes compared to \cite{han2020unsupervised} in two ways.  
First, instead of the size of $Q_k$ in proportional to the class frequency, we fix the size of $Q_k$ for all $k$ to the same amount as $L$. 
By averaging the same number of features from each class, we can compensate for the prototypes especially for the minority classes, with earlier samples remaining in $Q_k$.
Secondly, we adopt momentum encoder $f^{\text{enc}}_{\theta^{\prime}}$ when extracting the features for prototype generation inspired by~\cite{he2020momentum}. Note that $f^{\text{enc}}_{\theta^{\prime}}$ has the same architecture with $f^{\text{enc}}_{\theta}$, but $\theta^{\prime}$ is the exponential moving average (EMA) of $\theta$ with momentum ratio $\rho$, \ie, $\theta^{\prime} \leftarrow \rho \theta^{\prime} + (1-\rho )\theta$. This stabilizes the movement of each prototype in feature space across iteration by slowing the pace of network parameter updates. We will verify the effectiveness of balanced prototypes in~\Cref{tab:abl_proto_ema}.

\noindent{\textbf{Linear and semantic pseudo-label generation.}}
We obtain the linear pseudo-label $\hat{p}$ using the linear classifier followed by softmax activation: $\hat{p}=\sigma(f^{\text{cls}}_{\phi}(z^{(w)}))$. The semantic pseudo-label $\hat{q}$ is obtained from the similarity-based classifier that measures the per-class similarity of a query feature point $z$ of either $z^{(w)}$ or $z^{(s)}$ to the \emph{balanced prototypes} $\mC$:
\begin{equation}
    \label{eqn:sim_pl}
    q = \sigma\left(\text{sim}(z, \mC) \mathbin{/} T_\text{proto}\right),
\end{equation}
where $\text{sim}(\cdot, \cdot)$ corresponds to cosine similarity, and $T_{\text{proto}}$ is a temperature hyper-parameter for the classifier. Note that $\hat{p}$ is biased towards head classes while $\hat{q}$ is the vice versa. 

\noindent{\textbf{Distribution-aware blending.}}
To obtain class-specific unbiased pseudo-label $\hat{p}^{\prime}$, the semantic pseudo-label $\hat{q}$ should be exploited \emph{differently} across the class. 
To this end, we propose a novel blending method for pseudo-labels, where we increase the exposure of the component of $\hat{q}$ when $\hat{p}$ is more biased to the head classes.
Formally, we blend them with a set of distribution-aware weights $\upsilon=\{\upsilon_k \}^K_{k=1}$ to reduce the bias that might occur when using either $\hat{p}$ or $\hat{q}$:
\begin{equation}
    \label{eqn:mixup}
    \hat{p}^{\prime} = (1-\upsilon_{k^{\prime}})\, \hat{p} + \upsilon_{k^{\prime}} \hat{q},
\end{equation}
where $k^{\prime}$ is the class prediction from $\hat{p}$, and each $\upsilon_k$ is derived as $\upsilon_k=\frac{1}{\max_{k} \hat{m}_{k}^{ {1}\mathbin{/}{T_{\text{dist}}}}} \left(\hat{m}_k^{ {1}\mathbin{/}{T_{\text{dist}}} } \right)$.
Note that $\hat{m}$ is the normalized class distribution of the current pseudo-labels, which is the accumulation of $\hat{p}^{\prime}$ over a few previous iterations
and $T_{\text{dist}}$ is a hyper-parameter that intercedes the optimal trade-offs between $\hat{p}$ and $\hat{q}$.
Overall, in terms of the linear pseudo-label, the minority pseudo-labels will remain as minority, while pseudo-labels predicted as majority will be likely to recover the original classes thanks to large $\upsilon_{k^{\prime}}$. 

Note that we dynamically adjust the set of weights $\upsilon$ that determines relative intensity of $\hat{q}$ in~\cref{eqn:mixup}, based on the current bias of pseudo-labels $\hat{m}$.
This makes DASO flexible to various distributions of $\cU$ without resorting to any pre-defined distribution. 
For example, even under the same prediction of $\hat{p}$ for a head class, more $\hat{q}$ is blended when the current model is more biased. Similarly, a concurrent work~\cite{wang2022debiased} accumulates predictions for adaptive debiasing.

\noindent{\textbf{Semantic alignment loss.}}
To establish more balanced feature representations, we propose new semantic alignment loss for regularizing the feature encoder $f^{\text{enc}}_{\theta}$. It extends the consistency training framework with two asymmetric augmentations $\cA_{w}$ and $\cA_{s}$ like~\cite{xie2019unsupervised,sohn2020fixmatch} onto feature space. In high-level, we align each unlabeled sample $u$ to the most similar prototype used in the similarity-based classifier, by imposing \emph{consistent assignment} for two augmented views $\cA_w(u)$ and $\cA_s(u)$ to the same $c_k$ in feature space. 
Note $\hat{q}$ is reused to provide the target for $q^{(s)}$ with the cross-entropy loss $\cH$:
\begin{equation}
    \label{eqn:alignment}
    \cL_{\text{align}} = \cH\left(\hat{q},\, q^{(s)} \right),
\end{equation}
where $q^{(s)}$ is from the similarity-based classifier by passing through $z^{(s)}=f^{\text{enc}}_{\theta}(\cA_s(u))$ to~\cref{eqn:sim_pl}.
Since $\cL_{\text{align}}$ relates unlabeled data to the label space through consistently assigning to $\mC$ constructed from labeled features, such enhanced representation can implicitly guide the classifier $f_\phi^\text{cls}$ to produce less biased predictions in general, %
where we validate the efficacy of $\cL_{\text{align}}$ in~\cref{sec:exp_abl,sec:qualitative} respectively.

\noindent{\textbf{Total objective.}}
DASO is a generic framework that can easily couple with other SSL algorithms with the modified pseudo-label, where the final DASO objective is as below:
\begin{equation}
    \label{eqn:total}
    \cL_{\text{DASO}} = \cL_{\text{cls}} + \lambda_u \cL_{u} + \lambda_{\text{align}} \cL_{\text{align}},
\end{equation}

where both $\cL_{\text{cls}}$ and $\cL_{u}$ with $\lambda_u$ come from the base SSL learner, and $\cL_{\text{align}}$ is newly introduced from DASO. 
Note that $\cL_u$ takes the proposed blended pseudo-label in~\cref{eqn:mixup} instead of the original linear pseudo-label of the learner. 
We emphasize that DASO is also applicable to traditional SSL algorithms for performance gain without $\cL_{\text{align}}$ due to the absence of $\cA_s$ in the algorithm, as validated in~\Cref{tab:other_alg}.

%% file: main/exp.tex
\section{Experiments}
\label{sec:exp}
\input{main/tables/tab_main}

\input{main/tables/tab_imb}

\subsection{Experimental Setup}
\label{sec:exp_setup}
To ensure reproducibility\footnote{Code is available at: \url{https://github.com/ytaek-oh/daso}.}, all the settings of DASO and other baseline methods are clarified in~\cref{sec:impl_detail}.

\noindent{\textbf{Datasets.}}
We conduct SSL experiments with various scenarios where the class distribution of unlabeled data is not just limited to the class distribution of labeled data. 
To accommodate such conditions, we adopt CIFAR-10/100~\cite{krizhevsky2009learning} and STL-10~\cite{coates2011analysis} typically adopted in SSL literature~\cite{sohn2020fixmatch}. 
We make the imbalanced versions by exponentially decreasing the amount of samples per class~\cite{cui2019class}.
Following~\cite{kim2020distribution}, we denote the head class size as $N_1$ ($M_1$), and the imbalance ratio as $\gamma_l$ ($\gamma_u$) for the labeled (unlabeled) data respectively. Note that $\gamma_l$ and $\gamma_u$ can vary independently, and we specify `LT' for those imbalanced variants. We also consider Semi-Aves benchmark~\cite{su2021semisupervised} for practical setup, which is the large-scale collection of bird species with natural long-tailed distribution. 
Its unlabeled data also show long-tailed distribution, and include large portion of examples in broader categories compared to samples in labeled data (\eg, \emph{open-set}).
For more details, see~\cref{sec:sup_data_detail}. 

\noindent{\textbf{Baseline methods.}} 
We consider \emph{Supervised} baseline, learning cross-entropy with only labeled data. 
For using unlabeled data, we mainly adopt \emph{FixMatch}~\cite{sohn2020fixmatch} for its simplicity and powerful performances. 
To extensively validate our proposed method in terms of re-balancing, we mainly compare it with the following re-balancing algorithms on top of FixMatch.
Note that the results with other baseline SSL algorithms are provided in~\Cref{tab:other_alg} and the~\cref{sec:sup_remixmatch}.
We consider \emph{logit adjustment} (LA)~\cite{menon2021longtail} for balancing labels. Note that LA can also be applied to SSL methods for re-balancing using labels.
For re-balancing in unlabeled data similar to our framework, \emph{DARP}~\cite{kim2020distribution}
and \emph{CReST}~\cite{wei2021crest} are compared. %
We also experiment with the recently proposed \emph{ABC}~\cite{lee2021auxiliary} that performs single unified re-balancing using both labeled and unlabeled data simultaneously.

\noindent{\textbf{Training and evaluation.}}
We have re-implemented all the baseline methods using PyTorch~\cite{paszke2019pytorch} and conducted experiments under the same codebase for fair comparison, as suggested by \cite{oliver2018realistic}. We train Wide ResNet-28-2~\cite{zagoruyko2016wider} on CIFAR10/100-LT and STL10-LT as a backbone. For training Semi-Aves, we fine-tune the ResNet-34~\cite{he2016deep} pre-trained on ImageNet~\cite{deng2009imagenet}. 
To evaluate, we use the EMA network with the parameters updating every steps, following~\cite{berthelot2019mixmatch,kim2020distribution}.
As note, 
the class score is measured via learned linear classifier at inference time. We measure the top-1 accuracy on the test data every epoch and finally obtain the median of the accuracy values during the last 20 evaluations~\cite{berthelot2019mixmatch}. When reporting the results, we compute the mean and standard deviation of three independent runs.

\subsection{Results on CIFAR10/100-LT and STL10-LT.}
\label{sec:exp_results}
As the main results, we first consider the case when the distribution of labeled data and unlabeled data is the same (\eg, $\gamma=\gamma_l=\gamma_u$) in~\Cref{tab:main}, which is the ideal case for SSL. In~\Cref{tab:imb_exp}, we relax such assumption and test imbalanced SSL methods under practical yet challenging scenarios with diverse unlabeled data distributions (\eg, $\gamma_l \neq \gamma_u$).

\noindent \textbf{In case of $\gamma_l=\gamma_u$.} 
We compare the proposed DASO with several baseline methods, with or without class re-balancing in~\Cref{tab:main}. For \emph{Supervised} case, even if Logit Adjustment (LA)~\cite{menon2021longtail} is applied, the performances are rather limited compared to even na\"ive SSL method (\ie, FixMatch~\cite{sohn2020fixmatch}). 

We then compare imbalanced SSL methods: DARP~\cite{kim2020distribution} and CReST+~\cite{wei2021crest} with the proposed DASO on FixMatch. Remarkably, 
DASO shows comparable or even better results in most setups with significant gains compared to baseline FixMatch, 
although DARP and CReST+ even push the predictions of unlabeled data to the label distribution using the assumption $\gamma_l = \gamma_u$ (\ie, distribution alignment~\cite{berthelot2020remixmatch}). This verifies the efficacy of DASO for debiasing pseudo-labels, even without resorting to the label distribution.

To validate DASO can reliably benefit from re-balancing labels for debiasing pseudo-labels, we further compare imbalanced SSL methods on label re-balancing FixMatch via LA~\cite{menon2021longtail} (noted as FixMatch + LA). The results show DASO performs the best in most of the setups. 
It is noticeable that LA with DASO always improves performances compared to both FixMatch w/ DASO and FixMatch + LA cases. 

Finally, we consider ABC~\cite{lee2021auxiliary} in the bottom of~\Cref{tab:main}. 
It jointly trains the SSL learner and the auxiliary balanced classifier (ABC) using both labeled and unlabeled data with \emph{linear pseudo-labels}, while the ABC is opted for evaluation. 
We find that training ABC can readily be extended by just replacing the \emph{linear pseudo-label} for ABC with DASO pseudo-label~\eqref{eqn:mixup}. 
Finally, DASO can be significantly pushed by combining with ABC \cite{lee2021auxiliary} (\ie, 13\% gain upon FixMatch for CIFAR-10). It verifies the flexibility of DASO on any baselines regardless of re-balancing methods.

\noindent \textbf{In case of $\gamma_l \neq \gamma_u$.} 
The class distribution of unlabeled data could be either unknown or arguably different from that of the labeled data in real-world (\eg, $\gamma_l \neq \gamma_u$). 
To simulate such scenarios, for CIFAR10-LT, we consider two extreme cases 
for the class distribution of unlabeled data: uniform ($\gamma_u=1$) and flipped long-tail ($\gamma_u=1/100$) with respect to the labeled data.
For STL10-LT, since we cannot control the size and imbalance of unlabeled data due to unknown labels, we instead set $\gamma_l \in \{10, 20\}$ with the whole fixed unlabeled data.
\Cref{tab:imb_exp} summarizes the results of imbalanced SSL methods under the setups. 
Note that more comparisons of SSL methods with different re-balancing techniques (\ie, LA~\cite{menon2021longtail} and ABC~\cite{lee2021auxiliary}) are presented in~\cref{sec:sup_stl}. 

Surprisingly, DASO outperforms other baselines by significant margins in most cases. 
For example, DASO shows 13.6\% and 18.1\% of absolute gain from FixMatch upon CIFAR-10 ($\gamma_u=1$) and STL-10 ($\gamma_l=20$), respectively. 
Though DARP~\cite{kim2020distribution} estimates the distribution of unlabeled data in advance as the prior, the estimation accuracy decreases as using less labels for training.
Under $\gamma_l \neq \gamma_u$, we evaluate both CReST with self-training only and CReST+ with progressive distribution alignment~\cite{wei2021crest}.
Clearly, resorting to the label distributions as the prior for unlabeled data in CReST+ rather harms the accuracy compared to CReST, since the assumption of $\gamma_l=\gamma_u$ is violated. In particular, when the class distribution of unlabeled data is completely inverted ($\gamma_u=1/100$), the accuracy loss becomes more severe, resulting in little gain over FixMatch.

By virtue of debiased pseudo-labels from DASO, the abundant minority-class unlabeled samples are correctly used despite class-imbalanced labels. 
Consequently, the results confirm that conditioning on a certain distribution for unlabeled data (\eg, $\gamma_u=\gamma_l$) is undesirable in imbalanced SSL, and DASO greatly reduces the bias in presence of distribution mismatch, even without access to the distribution.
\input{main/tables/tab_otheralg_semiaves}
\noindent \textbf{DASO on other SSL learner.}
To verify DASO is a generic pseudo-labeling framework, we evaluate DASO based on other SSL algorithms including MeanTeacher~\cite{tarvainen2017mean}, MixMatch~\cite{berthelot2019mixmatch}, and ReMixMatch~\cite{berthelot2020remixmatch} in \Cref{tab:other_alg}. 
As note, MeanTeacher and MixMatch only perform pseudo-label blending~\eqref{eqn:mixup} without semantic alignment loss~\eqref{eqn:alignment} due to the absence of $\cA_{s}$. 
For CIFAR10-LT, we set $\gamma_l=100$ and for CIFAR100-LT and STL10-LT, we set $\gamma_l=10$. 
We observe that DASO greatly improves the performances for all the setups, and notably, it achieves 2.05$\times$ accuracy compared to MixMatch and brings 29.1\% absolute gain in ReMixMatch on CIFAR10-LT under $\gamma_u=1$. 
This implies that DASO noticeably helps SSL algorithms in general to benefit from unlabeled data under imbalanced SSL setup. As note, we show the comparison of imbalanced SSL methods built on other SSL learner (\eg, ReMixMatch~\cite{berthelot2020remixmatch}) in~\cref{sec:sup_remixmatch}.

\subsection{Results on Large-Scale Semi-Aves}
\label{sec:exp_large}
We test DASO on a realistic {Semi-Aves} benchmark~\cite{su2021semisupervised}. Both labeled data ($\cX$) and unlabeled data ($\cU$) show long-tailed distributions, while $\cU$ contains large \emph{open-set} examples ($\cU_\text{out}$) that do not belong to any of the classes in $\cX$.
The results are shown in \Cref{tab:semi_aves}. We report both cases: $\cU=\cU_\text{in}$ and $\cU=\cU_\text{in}+\cU_\text{out}$, where $\cU_\text{in}$ contains examples that share the class of $\cX$. We measure the performances by top-1 accuracy, reporting the one in the final (Last Top1) and the median values in last 20 epochs (Med20 Top1), following~\cite{oliver2018realistic}. More details on this dataset can be found in~\cref{sec:sup_data_detail}.

\noindent{\textbf{In case of $\cU=\cU_\text{in}$.}}
As it has the distribution gap between $\cX$ and $\cU$, baseline DARP~\cite{kim2020distribution} and CReST~\cite{wei2021crest} with inadequate class prior from $\cX$ show only a slight gain or even unsatisfactory performances compared to FixMatch~\cite{sohn2020fixmatch}. In contrary, DASO shows the best performance among the baselines with favorable improvements upon FixMatch.

\noindent{\textbf{In case of $\cU=\cU_{\text{in}} + \cU_\text{out}$.}}
Since $\cU$ contains large amount of \emph{open-set} class examples, performance drop is observed consistently across all baselines, as similar observations are made in~\cite{chen2020semi,guo2020safe,park2021opencos}. 
Among the baselines, DASO shows the best performance with favorable gain.
The results suggest that DARP~\cite{kim2020distribution} is slightly helpful when both $\cU_{\text{in}}$ and $\cU_{\text{out}}$ are considered altogether for optimization. 
Concerning CReST and CReST+~\cite{wei2021crest} with self-training, due to noisy predictions from $\cU_{\text{out}}$ for constructing datasets for the next generation, they rather performs poorly than FixMatch. 
As such, DASO has superiority in the challenging but practical scenario of long-tailed distributions, even in presence of large amount of open-set examples.
To understand this, we further provide the analyses on the confidence plots with or without DASO using each of $\cU_{\text{in}}$ and $\cU_{\text{out}}$ in~\cref{sec:sup_saves_conf}.

\subsection{Ablation Study}
\label{sec:exp_abl}
We conduct ablation studies to understand why DASO reliably provides improvements to baseline methods. 
To accommodate both $\gamma_l=\gamma_u$ and $\gamma_l \neq \gamma_u$ cases, we consider FixMatch on CIFAR10-LT with $N_1=500$, $\gamma=100$ (noted as C10) and STL10-LT with $N_1=150$, $\gamma_l=10$ (noted as STL10) respectively to evaluate each aspect of DASO. 

\noindent \textbf{Component analysis.}
\Cref{tab:abl_sa_loss} studies the two major components of DASO: distribution-aware pseudo-label blending and the semantic alignment loss. 
From the table, both blending mechanism and $\cL_{\text{align}}$ provides significant gain over FixMatch. For example, the blending and $\cL_{\text{align}}$ achieve about 6\% and 3\% absolute gain, respectively, and combining both shows 15.7\% gain in total on STL10. The results confirm that both class-adaptively blending linear and semantic pseudo-labels and the semantic alignment loss are important for reducing bias under imbalanced SSL. %

\noindent{\textbf{Effect of pseudo-label blending.}} 
\Cref{tab:abl_blending} studies the different way of pseudo-label blending on DASO with \emph{constant} weights.
Due to the bias in the pseudo-labels, using either linear ($\upsilon_k=0$) or semantic ($\upsilon_k=1$) pseudo-label leads to a marginal gain.
In addition, blending them with the same ratio ($\upsilon_k=0.5$) shows the lower performance compared to our final DASO, which demonstrates that distribution-aware class-adaptive blending is crucial for imbalanced SSL.

\noindent{\textbf{Effect of balanced prototype.}}
\Cref{tab:abl_proto_ema} studies the different design choices of DASO in prototype generation: balanced prototypes (noted as bal.) with EMA encoder (noted as EMA).
When generating class prototypes, using class-imbalanced queue without EMA encoder leads to worse performance.
In contrary, DASO with both balanced queue using EMA encoder shows the best performance, showing that both correspond to the valid components for the balanced prototypes from imbalanced labeled data. 

\noindent{\textbf{Ablation study on $T_{\text{dist}}$.}} 
In \Cref{tab:temp}, we study the effect of the temperature hyper-parameter $T_{\text{dist}}$ to compute the weights for pseudo-label blending described in \cref{eqn:mixup}.
We empirically find that, for CIFAR-10 and STL-10, $T_{\text{dist}}=1.5$ and $T_{\text{dist}}=0.3$ show the best performance respectively.

\begin{figure}[!t]
    \centering
    \includegraphics[width=.99\columnwidth]{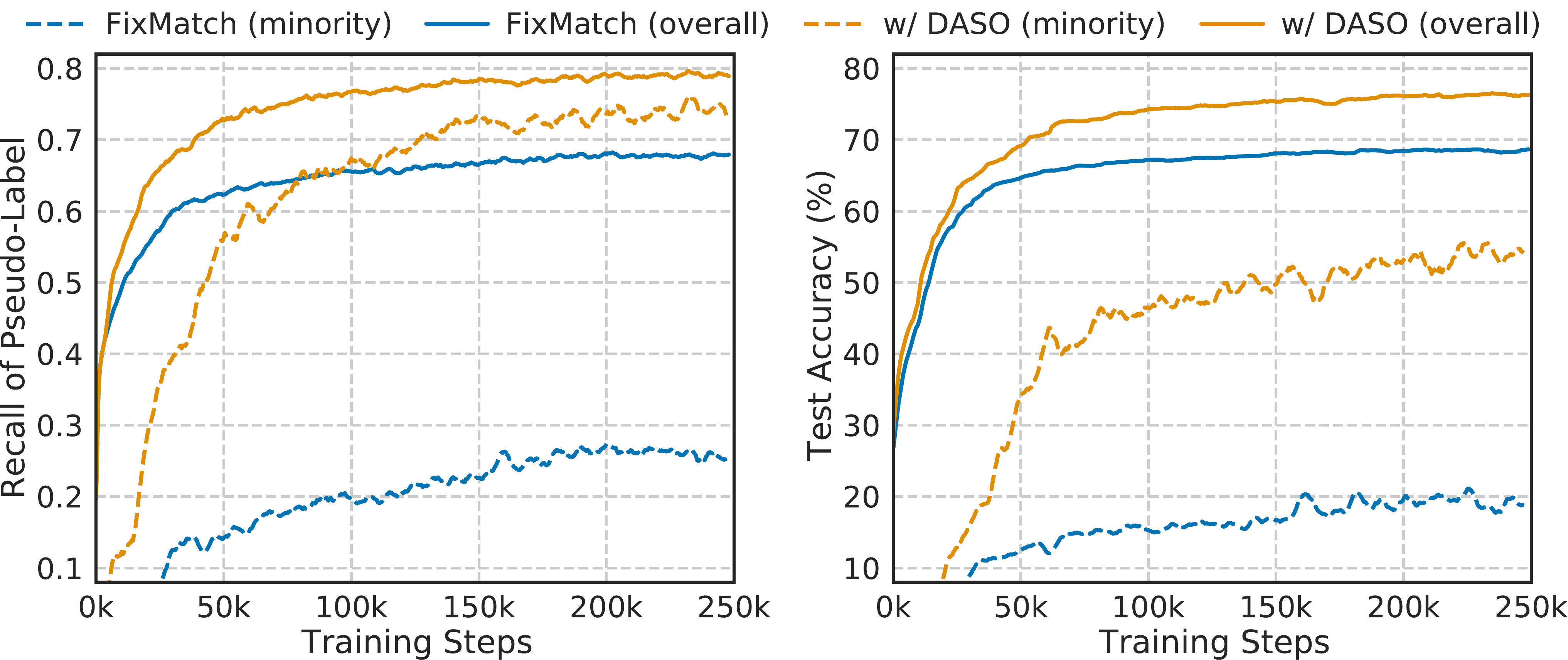}
    \caption{Train curves for the recall of pseudo-labels (left) and the test accuracy (right) on CIFAR10-LT.
    DASO significantly remedies the bias of pseudo-labels on minority classes, and such unbiased pseudo-labels lead to large gains on the test accuracy.
    }
    \label{fig:c10_main_analysis}
    \vspace{-2mm}
\end{figure}

\subsection{Detailed Analysis}
\label{sec:qualitative}
In this section, we qualitatively analyze how DASO improves the performance under imbalanced SSL setup. We consider FixMatch~\cite{sohn2020fixmatch} without and with DASO trained on CIFAR10-LT with $\gamma=100$ and $N_1=500$. Note that \aref{E} includes analyses in more various setups. 

\noindent \textbf{Unbiased pseudo-label improves test accuracy.} 
We visualize the train curves for the recall of pseudo-labels and the test accuracy values in \cref{fig:c10_main_analysis}. We denote those for the minorities (\eg, last 20\% classes) as dashed lines. From the left of~\cref{fig:c10_main_analysis}, DASO significantly raises the final recall for the tail classes, which is 3$\times$ compared to that of FixMatch. From the right, both minority and overall test accuracy values in final greatly improved by virtue of the less biased pseudo-labels towards the head classes, which are nearly 3$\times$ and 9\% compared to those of FixMatch, respectively.

\begin{figure}[!t]
    \centering
    \includegraphics[width=0.89\columnwidth]{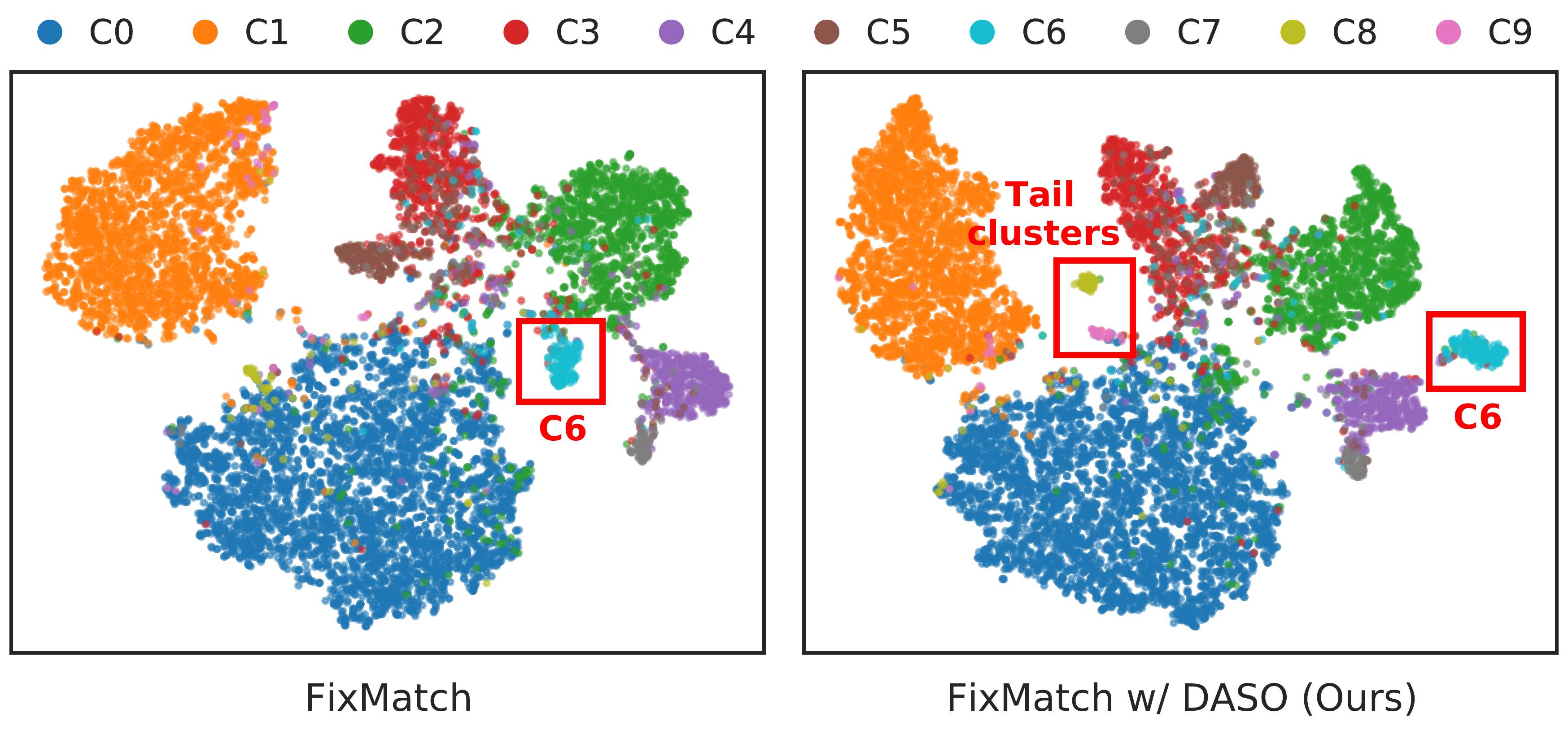}
    \caption{Comparison of t-SNE visualization of unlabeled data from FixMatch (left) and FixMatch w/ DASO (right). 
    Learning with DASO helps the model to establish tail-class clusters in feature space, which can further reduce the biases from the classifier. 
    }
    \label{fig:main_tsne_viz}
    \vspace{-2mm}
\end{figure}

\noindent \textbf{Tail-class clusters are better identified.} 
To verify the efficacy of reducing the bias, we present t-SNE~\cite{van2008visualizing} visualizations of the encoders' outputs on $\cU$ from FixMatch and w/ DASO respectively. As shown in~\cref{fig:main_tsne_viz}, tail class examples (\eg, C8 and C9) from FixMatch are scattered to the majority classes. From the right, however, the clusters of tail are clearly recognized as indicated. In addition, the separability of C6 is improved. Thanks to such well identified tail-class clusters from DASO, the actual minority unlabeled examples are correctly leveraged to learn the unbiased model. 

%% file: main/tables/tab_main.tex
\begin{table*}[!t]
\centering
 \resizebox{0.87\linewidth}{!}{%
    \begin{tabular}{lcccccccc}
         \toprule
         & \multicolumn{4}{c}{CIFAR10-LT} & \multicolumn{4}{c}{CIFAR100-LT} \\
         & \multicolumn{2}{c}{$\gamma=\gamma_l=\gamma_u=100$} & \multicolumn{2}{c}{$\gamma=\gamma_l=\gamma_u=150$}  & \multicolumn{2}{c}{$\gamma=\gamma_l=\gamma_u=10$} & \multicolumn{2}{c}{$\gamma=\gamma_l=\gamma_u=20$} \\
         \cmidrule(lr){2-3} \cmidrule(lr){4-5} \cmidrule(lr){6-7} \cmidrule(l){8-9}
         \multirow{2}{*}{Algorithm} & $N_1=500$ & $N_1=1500$ & $N_1=500$ & $N_1=1500$ & $N_1=50$ & $N_1=150$ & $N_1=50$ & $N_1=150$ \\
         & $M_1=4000$ & $M_1=3000$ & $M_1=4000$ & $M_1=3000$ & $M_1=400$ & $M_1=300$ & $M_1=400$ & $M_1=300$ \\

        \cmidrule(r){1-1} \cmidrule(lr){2-3} \cmidrule(lr){4-5} \cmidrule(lr){6-7} \cmidrule(l){8-9}

        Supervised & \ms{47.3}{0.95} & \ms{61.9}{0.41}  & \ms{44.2}{0.33} & \ms{58.2}{0.29}  & \ms{29.6}{0.57} & \ms{46.9}{0.22} &  \ms{25.1}{1.14} & \ms{41.2}{0.15} \\
        ~~ w/ LA~\cite{menon2021longtail} & \ms{53.3}{0.44} & \ms{70.6}{0.21}  & \ms{49.5}{0.40} & \ms{67.1}{0.78}  & \ms{30.2}{0.44} & \ms{48.7}{0.89}  &  \ms{26.5}{1.31} & \ms{44.1}{0.42} \\

        \cmidrule(r){1-1} \cmidrule(lr){2-3} \cmidrule(lr){4-5} \cmidrule(lr){6-7} \cmidrule(l){8-9}
        
        FixMatch~\cite{sohn2020fixmatch} & \ms{67.8}{1.13} & \ms{77.5}{1.32}  & \ms{62.9}{0.36}  & \ms{72.4}{1.03}  & \ms{45.2}{0.55} & \ms{56.5}{0.06} & \ms{40.0}{0.96} & \ms{50.7}{0.25} \\
        ~~w/ DARP~\cite{kim2020distribution} & \ms{74.5}{0.78} & \ms{77.8}{0.63}  & \ms{67.2}{0.32} & \ms{73.6}{0.73}  & \ms{49.4}{0.20} & \ms{58.1}{0.44} & \ms{43.4}{0.87}  & \ms{52.2}{0.66} \\
        ~~w/ CReST+~\cite{wei2021crest} & \msb{76.3}{0.86} & \ms{78.1}{0.42} & \ms{67.5}{0.45} & \ms{73.7}{0.34} & \ms{44.5}{0.94} & \ms{57.4}{0.18} & \ms{40.1}{1.28} & \ms{52.1}{0.21} \\
        ~~w/ DASO (Ours) & \ms{76.0}{0.37} & \msb{79.1}{0.75}   & \msb{70.1}{1.81} & \msb{75.1}{0.77}  & \msb{49.8}{0.24} & \msb{59.2}{0.35} & \msb{43.6}{0.09} & \msb{52.9}{0.42} \\
        \cmidrule(r){1-1} \cmidrule(lr){2-3} \cmidrule(lr){4-5} \cmidrule(lr){6-7} \cmidrule(l){8-9}

        FixMatch + LA~\cite{menon2021longtail} & \ms{75.3}{2.45} & \ms{82.0}{0.36}  & \ms{67.0}{2.49} & \ms{78.0}{0.91}  & \ms{47.3}{0.42} & \ms{58.6}{0.36}  &  \ms{41.4}{0.93} & \ms{53.4}{0.32} \\
        ~~w/ DARP~\cite{kim2020distribution} & \ms{76.6}{0.92} & \ms{80.8}{0.62} & \ms{68.2}{0.94} & \ms{76.7}{1.13} & \ms{50.5}{0.78} & \ms{59.9}{0.32} & \msb{44.4}{0.65} & \ms{53.8}{0.43} \\
        ~~w/ CReST+~\cite{wei2021crest} & \ms{76.7}{1.13} & \ms{81.1}{0.57} & \msb{70.9}{1.18} & \ms{77.9}{0.71} & \ms{44.0}{0.21} & \ms{57.1}{0.55} & \ms{40.6}{0.55} & \ms{52.3}{0.20} \\
        ~~w/ DASO (Ours) & \msb{77.9}{0.88} & \msb{82.5}{0.08}   & \ms{70.1}{1.68} & \msb{79.0}{2.23}  & \msb{50.7}{0.51} & \msb{60.6}{0.71} &  \ms{44.1}{0.61} & \msb{55.1}{0.72} \\
        \cmidrule(r){1-1} \cmidrule(lr){2-3} \cmidrule(lr){4-5} \cmidrule(lr){6-7} \cmidrule(l){8-9}

        FixMatch + ABC~\cite{lee2021auxiliary} & \ms{78.9}{0.82} & \msb{83.8}{0.36} & \ms{66.5}{0.78} & \ms{80.1}{0.45} & \ms{47.5}{0.18} & \ms{59.1}{0
        .21} & \ms{41.6}{0.83} & \ms{53.7}{0.55} \\
        ~~w/ DASO (Ours) & \msb{80.1}{1.16} & \ms{83.4}{0.31} & \msb{70.6}{0.80} & \msb{80.4}{0.56} & \msb{50.2}{0.62} & \msb{60.0}{0.32} & \msb{44.5}{0.25} & \msb{55.3}{0.53} \\
        \bottomrule
    \end{tabular}
}
\caption{
Comparison of accuracy (\%) with combinations of re-balancing methods on CIFAR10/100-LT under $\gamma_l=\gamma_u$ setup.
Our DASO consistently improves the performance over all the baselines without or with re-balancing, even with ABC~\cite{lee2021auxiliary} designed for imbalanced SSL.
We indicate the best results for each division as bold. 
More results including new baseline methods are provided in~\cref{sec:sup_comprehensive}.
}
\vspace{0.2\baselineskip} 
\label{tab:main}
\end{table*}

%% file: main/tables/tab_imb.tex
\begin{table*}[t]
\centering
\resizebox{.88\linewidth}{!}{%
    \begin{tabular}{lcccccccccc}
        \toprule
         & \multicolumn{4}{c}{CIFAR10-LT ($\gamma_l \neq \gamma_u$)} & \multicolumn{4}{c}{STL10-LT ($\gamma_u=$\emph{ N/A})} \\
         & \multicolumn{2}{c}{$\gamma_u=1$ (uniform)} & \multicolumn{2}{c}{$\gamma_u=1/100$ (reversed)}  & \multicolumn{2}{c}{$\gamma_l=10$} & \multicolumn{2}{c}{$\gamma_l=20$} \\
         \cmidrule(lr){2-3} \cmidrule(lr){4-5} \cmidrule(lr){6-7} \cmidrule(l){8-9}
         \multirow{2}{*}{Algorithm}  & $N_1=500$  & $N_1=1500$ & $N_1=500$  & $N_1=1500$ & $N_1=150$  & $N_1=450$ & $N_1=150$  & $N_1=450$ \\
                                     & $M_1=4000$ & $M_1=3000$ & $M_1=4000$ & $M_1=3000$ & $M=100k$ & $M=100k$ & $M=100k$ & $M=100k$ \\
        \cmidrule(r){1-1} \cmidrule(lr){2-3} \cmidrule(lr){4-5} \cmidrule(lr){6-7} \cmidrule(l){8-9}

        FixMatch~\cite{sohn2020fixmatch}        & \ms{73.0}{3.81}  & \ms{81.5}{1.15} & \ms{62.5}{0.94}  & \ms{71.8}{1.70} & \ms{56.1}{2.32}  & \ms{72.4}{0.71} & \ms{47.6}{4.87} & \ms{64.0}{2.27} \\
        ~~w/ DARP~\cite{kim2020distribution}    & \ms{82.5}{0.75}  & \ms{84.6}{0.34} & \ms{70.1}{0.22}  & \ms{80.0}{0.93} & \ms{66.9}{1.66}  & \ms{75.6}{0.45} & \ms{59.9}{2.17} & \ms{72.3}{0.60}\\
        ~~w/ CReST~\cite{wei2021crest}   & \ms{83.2}{1.67}  & \ms{87.1}{0.28} & \ms{70.7}{2.02} & \msb{80.8}{0.39} & \ms{61.7}{2.51} & \ms{71.6}{1.17} & \ms{57.1}{3.67} & \ms{68.6}{0.88} \\
        ~~w/ CReST+~\cite{wei2021crest}  & \ms{82.2}{1.53} & \ms{86.4}{0.42} & \ms{62.9}{1.39} & \ms{72.9}{2.00} & \ms{61.2}{1.27} & \ms{71.5}{0.96} & \ms{56.0}{3.19} & \ms{68.5}{1.88} \\
        ~~w/ DASO (Ours)                        & \msb{86.6}{0.84} & \msb{88.8}{0.59} & \msb{71.0}{0.95} & \ms{80.3}{0.65} & \msb{70.0}{1.19} & \msb{78.4}{0.80} & \msb{65.7}{1.78} & \msb{75.3}{0.44} \\
        
        \bottomrule
    \end{tabular}%
}
\caption{Comparison of accuracy (\%) for imbalanced SSL methods on CIFAR10-LT and STL10-LT under $\gamma_l \neq \gamma_u$ setup. 
For CIFAR10-LT, $\gamma_l$ is fixed to 100, and $\gamma_u$ is unknown for STL10-LT. Our DASO consistently shows significant gains on FixMatch~\cite{sohn2020fixmatch} without resorting to any class prior under diverse class distribution mismatches between labeled and unlabeled data. 
We indicate the best results as bold. 
}
\label{tab:imb_exp}
\vspace{-2mm}
\end{table*}

%% file: main/tables/tab_otheralg_semiaves.tex
\begin{table*}[!t]
\centering
\begin{minipage}[t]{0.47\linewidth}\vspace{0pt}
    \centering
    \resizebox{0.92\linewidth}{!}{%
    \begin{tabular}{lcccc}
        \toprule
         & \multicolumn{2}{c}{C10-LT} & \multicolumn{1}{c}{C100-LT} & \multicolumn{1}{c}{STL10-LT}\\
        \cmidrule(lr){2-3} \cmidrule(lr){4-4} \cmidrule(l){5-5}
         & \multicolumn{2}{c}{$N_1=1500$} & $N_1=150$ & $N_1=450$  \\
         & \multicolumn{2}{c}{$M_1=3000$} & $M_1=300$ & $M=100k$  \\
        \cmidrule(lr){2-3} \cmidrule(lr){4-4} \cmidrule(l){5-5}
        {Algorithm} & \multicolumn{1}{c}{$\gamma_u=100$} & \multicolumn{1}{c}{$\gamma_u=1$} & \multicolumn{1}{c}{$\gamma_u=10$} & \multicolumn{1}{c}{$\gamma_u:$ \emph{N/A}} \\
        \cmidrule(r){1-1} \cmidrule(lr){2-3} \cmidrule(lr){4-4} \cmidrule(l){5-5}
        Mean Teacher~\cite{tarvainen2017mean}    & \ms{68.6}{0.88}           & \ms{46.4}{0.98}           & \ms{52.1}{0.09}           & \ms{54.6}{1.17} \\
        ~~w/ DASO (Ours) & \msb{{70.7}}{0.59}  & \msb{{87.6}}{0.27}  & \msb{{52.5}}{0.37}  & \msb{{78.4}}{0.80} \\
        \cmidrule(r){1-1} \cmidrule(lr){2-3} \cmidrule(lr){4-4} \cmidrule(l){5-5}
        MixMatch~\cite{berthelot2019mixmatch}        & \ms{65.7}{0.23}           & \ms{35.7}{0.69}           & \ms{54.2}{0.47}           &  \ms{52.7}{1.42} \\
        ~~w/ DASO (Ours) & \msb{{70.9}}{1.91}  & \msb{{73.4}}{2.05}  & \msb{{55.6}}{0.49}  &  \msb{{68.4}}{0.71} \\
        \cmidrule(r){1-1} \cmidrule(lr){2-3} \cmidrule(lr){4-4} \cmidrule(l){5-5}
        ReMixMatch~\cite{berthelot2020remixmatch}      & \ms{77.0}{0.55}           & \ms{60.4}{0.70}                    & \ms{61.5}{0.57}           & \ms{71.9}{0.86} \\
        ~~w/ DASO (Ours) & \msb{{80.2}}{0.68}  & \msb{90.5}{0.35}                    & \msb{62.1}{0.69}  & \msb{{80.9}}{0.55} \\
        \bottomrule
    \end{tabular}%
    }
    \caption{Comparison of accuracy (\%) from DASO upon other SSL methods: MeanTeacher~\cite{tarvainen2017mean}, MixMatch~\cite{berthelot2019mixmatch}, and ReMixMatch~\cite{berthelot2020remixmatch}.
    DASO improves the performances in all the setups.
    }
    \label{tab:other_alg}
\end{minipage}
\qquad~
\centering
\begin{minipage}[t]{0.47\linewidth}\vspace{0pt}
    \centering
    \resizebox{.99\columnwidth}{!}{%
        \begin{tabular}{lcccccc}
            \toprule
            \multicolumn{1}{c}{\multirow{2}{*}{Benchmark}} & \multicolumn{4}{c}{Semi-Aves} \\
            &  \multicolumn{2}{c}{$\cU = \cU_{\text{in}}$} & \multicolumn{2}{c}{$\cU = \cU_{\text{in}} + \cU_{\text{out}}$} \\
            \cmidrule(r){1-1} \cmidrule(r){2-3} \cmidrule(l){4-5}
            Method & Last Top1 & Med20 Top1 & Last Top1 & Med20 Top1  \\
            \cmidrule(r){1-1} \cmidrule(r){2-3} \cmidrule(l){4-5}
            Supervised                              & \ms{41.7}{0.32} & \ms{41.7}{0.32}  & \ms{41.7}{0.32} & \ms{41.7}{0.32}  \\
            \cmidrule(r){1-1} \cmidrule(r){2-3} \cmidrule(l){4-5}
            FixMatch~\cite{sohn2020fixmatch}        & \ms{53.8}{0.17} & \ms{53.8}{0.13} & \ms{45.7}{0.89} & \ms{46.1}{0.50}   \\
            ~~w/ DARP~\cite{kim2020distribution}    & \ms{52.3}{0.48} & \ms{52.1}{0.48} & \ms{46.3}{0.70} & \ms{46.4}{0.61}    \\
            ~~w/ CReST~\cite{wei2021crest}          & \ms{52.1}{0.36} & \ms{52.2}{0.27} & \ms{43.6}{0.69} & \ms{43.6}{0.68} \\
            ~~w/ CReST+~\cite{wei2021crest}         & \ms{53.9}{0.38} & \ms{53.8}{0.38} & \ms{45.1}{1.09} & \ms{45.2}{1.00} \\
            ~~w/ DASO (Ours)                        & \msb{{54.5}}{0.08} & \msb{{54.6}}{0.12} & \msb{47.9}{0.41} & \msb{47.9}{0.38}  \\
            \bottomrule
        \end{tabular}%
    }
\caption{Comparison of accuracy (\%) on Semi-Aves benchmark~\cite{su2021semisupervised}. DASO shows the best performance among state-of-the-art imbalanced SSL methods. Moreover, DASO still performs well in presence of massive open-set class examples $\cU_{\text{out}}$.}
\label{tab:semi_aves}
\end{minipage}

\vspace{1.0\baselineskip}
\begin{minipage}{0.24\linewidth}
    \centering
    \resizebox{.99\linewidth}{!}{%
        \begin{tabular}{lccc}
            \toprule
            & $\cL_{\text{align}}$ & C10 & STL10 \\
            \midrule
            FixMatch & \xmark & 68.25 & 55.53 \\
            DASO & \xmark & 70.98 & 61.64 \\
            FixMatch & \cmark & 73.15 & 58.51 \\
            \midrule
            DASO & \cmark & \textbf{75.97} & \textbf{70.21} \\
            \bottomrule
        \end{tabular}%
    }
    \caption{Ablation study on pseudo-label blending and semantic alignment loss $\cL_{\text{align}}$.}
    \label{tab:abl_sa_loss}
\end{minipage}
\quad
\begin{minipage}{0.22\linewidth}
    \centering
    \resizebox{.83\linewidth}{!}{%
        \begin{tabular}{lcc}
            \toprule
             & C10 & STL10 \\
            \midrule
            $\upsilon_k=0$ & 73.15 & 58.51 \\
            $\upsilon_k=1$ & 72.35 & 62.60 \\
            $\upsilon_k=0.5$ & 72.96 & 64.21 \\
            \midrule
            DASO & \textbf{75.97} & \textbf{70.21} \\
            \bottomrule
        \end{tabular}%
    } 
    \caption{Ablation study on the pseudo-label blending strategy with $\cL_{\text{align}}$ applied.}
    \label{tab:abl_blending}
\end{minipage}
\quad
\begin{minipage}{0.23\linewidth}
    \centering
    \resizebox{.88\linewidth}{!}{%
        \begin{tabular}{cccc}
            \toprule
             bal. & EMA & C10 & STL10 \\
            \midrule
            \xmark & \xmark & 74.98 & 68.54 \\
            \cmark & \xmark & 74.54 & 70.01 \\
            \xmark & \cmark & 75.01 & 69.49 \\
            \midrule
            \cmark & \cmark & \textbf{75.97} & \textbf{70.21} \\
            \bottomrule
        \end{tabular}%
    } 
    \caption{Ablation study on balancing prototypes and using EMA encoder on DASO.}
    \label{tab:abl_proto_ema}
\end{minipage}
\quad
\begin{minipage}{0.23\linewidth}
    \centering
    \resizebox{.9\linewidth}{!}{%
        \begin{tabular}{lcc}
            \toprule
             & C10 & STL10 \\
            \cmidrule{1-1} \cmidrule{2-3}
            $T_{\text{dist}}=0.3$ & 73.97 & \textbf{70.21} \\
            $T_{\text{dist}}=0.5$ & 74.47 & 68.35 \\
            $T_{\text{dist}}=1.0$ & 74.82 & 65.96 \\
            $T_{\text{dist}}=1.5$ & \textbf{75.97} & 64.54 \\
            \bottomrule
        \end{tabular}%
    } 
    \caption{Ablation study on $T_{\text{dist}}$ for DASO. We select $T_{\text{dist}}$ by 1.5 and 0.3 each. 
    }
    \label{tab:temp}
\end{minipage}
\vspace{-2mm} 
\end{table*}

%% file: main/conclusion.tex
\section{Discussion}
\label{sec:conclusion}
\noindent \textbf{Conclusion.} 
We proposed a novel distribution-aware semantics-oriented (DASO) pseudo-label for imbalanced semi-supervised learning. 
DASO adaptively blends the linear and semantic pseudo-labels within each class to mitigate the overall bias across the class.
Moreover, we introduced balanced prototypes and semantic alignment loss. 
From extensive experiments, we showed the efficacy of DASO on various challenging and realistic setups, especially when class imbalance and class distribution mismatch dominate.

\noindent \textbf{Potential societal impact.}
The proposed solution can contribute to solving various social problems attributed to imbalance in real-world, such as gender, racial or religious bias, by improving the fairness of classifiers using unlabeled data.
Also, our method can contribute to the active learning research~\cite{cho2021mcdal,kim2021single,shin2021labor}, which can also suffer from the bias.
However, the proposed algorithm should be carefully considered as it can be used to raise other fairness issues such as over-balance or discrimination against minorities.

\noindent \textbf{Limitations.}
This study focused on alleviating the bias of pseudo-labels, treating unlabeled data as \emph{truly unlabeled}.  
DASO modulates the debiased pseudo-labels by introducing a hyper-parameter $T_{\text{dist}}$,
which is effective and efficient 
than
estimating the class distribution of unlabeled data.
However,
$T_{\text{dist}}$ can be highly dependent on each data and distribution.
As mentioned in~\cite{oliver2018realistic}, tuning such hyper-parameter is not straightforward under label-scarce setting, which is the common concern in %
SSL literature. 

\section*{Acknowledgements}
This research was supported by the National Research Foundation of Korea (NRF)’s program of developing and demonstrating innovative products based on public demand funded by the Korean government (Ministry of Science and ICT (MSIT)) (No. NRF-2021M3E8A2100445).

%% file: supple/notations.tex
\section{Notations}
In this section, we clarify all the notations with corresponding descriptions introduced in this work.
\begin{table*}[h]
\begin{center}
    \resizebox{0.95\linewidth}{!}{
    \begin{tabular}{p{0.2\textwidth}p{0.79\textwidth}}
        \toprule
        Notation & Description \\
        \midrule
        DASO & Distribution-Aware Semantic-Oriented (Pseudo-label) \\
        \cmidrule{1-2}
        SSL & Semi-Supervised Learning. \\
        \cmidrule{1-2}
        $K$ & The number of classes in the labeled data. \\
        \cmidrule{1-2}
        $\cX$, \;$\cU$ & Labeled data and unlabeled data. \\
        \cmidrule{1-2}
        $N, \; M$ & Total number of examples in labeled data and unlabeled data. \\
        \cmidrule{1-2}
        $N_k$,\;$M_k$ & Number of examples in class $k$ for labeled data and unlabeled data. \\
        \cmidrule{1-2}
        $\gamma_l,\; \gamma_u$ & Imbalance ratio for labeled data and unlabeled data. \\
        \cmidrule{1-2}
        $\hat{m}$ & Empirical pseudo-label distribution in probability form; $\hat{m} \in \left[0, 1 \right]^K$. \\
        \cmidrule{1-2}
        $\sigma(\cdot)$ & Softmax activation. \\
        \cmidrule{1-2}
        $\cH(y, \;p)$ & Cross-entropy between the target $y$ and prediction $p$. \\
        \cmidrule{1-2}
        $\text{sim}(\cdot, \cdot)$ & Cosine similarity. \\
        \cmidrule{1-2}
        $f$ & A classification model; a feature encoder $f^{\text{enc}}_{\theta}$ followed by a linear classifier $f^{\text{cls}}_{\phi}$. \\
        \cmidrule{1-2}
        $f^{\text{enc}}_{\theta^\prime}$ & An EMA encoder (momentum encoder). \\
        \cmidrule{1-2}
        $\rho$ & Decay ratio for the momentum encoder. \\
        \cmidrule{1-2}
        $\mQ$ & A dictionary of memory queue; $\{Q_k \}^K_{k=1}$. \\
        \cmidrule{1-2}
        L & The maximum queue size for the \emph{balanced} memory queue. \\
        \cmidrule{1-2}
        $\mC$ & A set of class prototypes; $\{c_k \}^{K}_{k=1}$. \\
        \cmidrule{1-2}
        $T_{\text{proto}}$ & A temperature factor for the similarity-based classifier. \\
        \cmidrule{1-2}
        $T_{\text{dist}}$ & A temperature factor for the empirical pseudo-label distribution. \\
        \cmidrule{1-2}
        $\hat{p}, \;q^{(w)}$ or $\hat{q}$ & A linear pseudo-label and semantic pseudo-label. \\
        \cmidrule{1-2}
        $\upsilon$ & Class-specific mixup factor for the linear and semantic pseudo-label; $\{\upsilon_k \}^K_{k=1}$. \\
        \cmidrule{1-2}
        $\hat{p}^{\prime}$ & A blended pseudo-label. \\
        \cmidrule{1-2}
        $\text{PseudoLabel}(\cdot)$ & Pseudo-labeler specified by an SSL algorithm. \\
        \cmidrule{1-2}
        $\Phi_u (\cdot, \cdot)$ & A regularizer for $\cU$, specified by an SSL algorithm. \\
        \cmidrule{1-2}
        $\lambda_{u}$ & The loss weight for $\cL_u$. \\
        \cmidrule{1-2}
        $\cL_{\text{align}}$ & Semantic alignment loss. \\
        \cmidrule{1-2}
        $\lambda_{\text{align}}$ & The loss weight for $\cL_{\text{align}}$. \\
        \cmidrule{1-2}
        $P$ & Pre-train steps for applying pseudo-label blending and $\cL_{\text{align}}$. \\
        \cmidrule{1-2}
        $\cA_{w}$ & A set of weak augmentations; horizontal \emph{flip} and/or \emph{crop}. \\
        \cmidrule{1-2}
        $\cA_{s}$ & A set of strong augmentations; RandAugment~\cite{cubuk2020randaugment} followed by Cutout~\cite{devries2017improved}. \\
        \cmidrule{1-2}
        $\mu$ & Unlabeled batch ratio; multiplied to the labeled batch size $B$. \\
        \bottomrule
        \end{tabular}
    }
    \caption{Notations and their descriptions used throughout this work.}
    \label{tab:Notation}
\end{center}
\end{table*}
\clearpage

%% file: supple/algorithm.tex
\section{Algorithm}
\Cref{alg:blending} summarizes the blending procedure for the linear and semantic pseudo-labels based on the empirical pseudo-label distribution, and \Cref{alg:daso} represents the whole DASO framework built upon a typical SSL algorithm where the regularizer for the SSL algorithm corresponds to $\Phi_{{u}}$. 
\begin{algorithm*}[!ht]
    \centering
    \caption{Distribution-aware pseudo-label blending,\quad $\hat{p}^{\prime} \leftarrow \text{Blend} \left(\hat{p}, \hat{q}, T_{\text{dist}} \right)$.}
    \label{alg:blending}
    \begin{algorithmic}
        \STATE {\bfseries Input:} Linear pseudo-label $\hat{p} \in [0, 1]^K$, semantic pseudo-label $\hat{q} \in [0, 1]^K$, \\
        \qquad \quad Temperature factor for the pseudo-label distribution $T_{\text{dist}}$.
        \STATE {\bfseries Require:} Empirical pseudo-label distribution $\hat{m}=\{\hat{m}_k \}^{K}_{k=1}$.
        \STATE {\bfseries Output:} Blended pseudo-label $\hat{p}^{\prime} \in [0, 1]^K$. \vspace{1mm}
        \FOR{$k=1$ {\bfseries to} $K$}
            \STATE $\upsilon_k \leftarrow \hat{m}^{1/T_{\text{dist}}}_k$ \quad \quad \COMMENT{Temperature scaling for empirical pseudo-label distribution.}
            \STATE $\upsilon_k \leftarrow \upsilon_k / \max_k \upsilon_k$ \COMMENT{Normalization for blending.}
        \ENDFOR
        \STATE $k^{\prime} \leftarrow \argmax_{k} \hat{p}_k$ \quad \quad \quad \COMMENT{Class prediction of the linear pseudo-label.}
        \STATE $\hat{p}^{\prime} \leftarrow (1-\upsilon_{k^{\prime}})\, \hat{p} + \upsilon_{k^{\prime}} \hat{q}$ \quad \COMMENT{Pseudo-label blending.} \vspace{1mm}
    \end{algorithmic}
\end{algorithm*}
\begin{algorithm*}[!ht]
    \centering
    \caption{Distribution-Aware Semantic-Oriented (DASO) Pseudo-label framework.}
    \label{alg:daso}
    \begin{algorithmic} 
        \STATE {\bfseries Input:} A batch of labeled data $\cX_B=\{\left(x_b, y_b \right) \}^{B}_{b=1}$ and unlabeled data $\cU_B=\{u_b \}^{\mu B}_{b=1}$. \\
        \quad \qquad Network for feature encoder $f^{\text{enc} }_{\theta}$, momentum encoder $f^{\text{enc} }_{\theta^{\prime}}$, and linear classifier $f^{\text{cls} }_{\phi}$. \\
        \quad \qquad Dictionary of memory queue $\mQ=\{Q_k \}^{K}_{k=1}$, Momentum decay ratio $\rho$. \\
        \quad \qquad Maximum queue size L, temperature factor for the similarity-based classifier $T_{\text{proto}}$, \\
        \quad \qquad Pre-train steps for pseudo-label blending $P$, current training step t. \vspace{1mm}
        \STATE {\bfseries Require:} A set of weak augmentations $\cA_w$ and strong augmentations $\cA_s$. \vspace{1mm}
        \STATE \vspace{1mm} \COMMENT{\textbf{Balanced Prototype Generation.}} \vspace{1mm}
        \STATE Enqueue $z^{(l)}$ into $Q_k$, where $z^{(l)}=f^{\text{enc}}_{\theta^{\prime}}(x)$ and $k\leftarrow y$, \quad $\forall\, (x, y)\in\cX_B$. \vspace{1mm}
        \STATE Dequeue the earliest elements from~$Q_{k}$ ~\emph{s.t.} $|Q_{k}| = L$, \quad $\forall\, k\in \{1,\ldots, K \}$. \vspace{1mm}
        \STATE $c_k \leftarrow \frac{1} {|Q_k|} \sum_{z_i \in Q_k} z_i$,~$\;\forall\, k\in \{1,\ldots, K \}$, \quad \COMMENT{A set of balanced prototypes $\mC=\{c_k \}^{K}_{k=1}$.} \vspace{2mm}
        \STATE \COMMENT{\textbf{Pseudo-label generation.}}
        \FOR{$u$ {\bfseries in} $\cU_B$}
            \STATE $z^{(w)} \leftarrow f^{\text{enc} }_{\theta} \left(\cA_w(u)\right)$, \quad $z^{(s)} \leftarrow f^{\text{enc} }_{\theta} \left(\cA_s(u)\right)$ \quad \COMMENT{feature extraction}
            \STATE $\hat{p} \leftarrow \text{$\sigma$} \left( f^{\text{cls}}_{\phi} \left(z^{(w)}\right) \right)$, \quad $q^{(w)} \leftarrow \sigma \left( {\text{sim}\left(z^{(w)}, \mC \right)} \mathbbm{/}\,{T_{\text{proto}}} \right)$
            \STATE $\hat{p}^{\prime} \leftarrow \text{Blend} \left(\hat{p},\,q^{(w)},\,T_{\text{dist}} \right)$ \textbf{if} $t \geq P$ \textbf{else} $\hat{p}$ \quad \COMMENT{Blend pseudo-labels after $P$ train steps.}
        \ENDFOR \vspace{2mm}
        \STATE \COMMENT{\textbf{Compute losses.}}
        \vspace{1mm}
        \STATE $\cL_{\text{cls}} \leftarrow \Ebb_{(x, y) \in \cX_B} \left[ \cH \left( y,\,\sigma \left( f(x)\right)\right) \right] $
        \vspace{1mm}
        \STATE $\cL_{\text{align}} \leftarrow \Ebb_{u \in \cU_B} \left[\mathbbm{1}\left(t \geq P \right) \cdot  \cH \left(q^{(w)},\,q^{(s)} \right)\right]$ $\,$ where $\,q^{(s)} \leftarrow \sigma \left( {\text{sim}\left(z^{(s)}, \mC \right)}\mathbbm{/}\,{T_{\text{proto}}} \right)$.
        \vspace{1mm}
        \STATE $\cL_u \leftarrow \Ebb_{u \in \cU_B} \left[ \Phi_{{u}}\left( {\hat{p}^{\prime},\,p^{\left(s \right)}}\right) \right]$ where $p^{(s)} \leftarrow f^{\text{cls}}_{\phi} \left( z^{\text{(s)}} \right)$.
        \vspace{1mm}
        \STATE $\cL_{\text{DASO}} \leftarrow \cL_{\text{cls}} + \lambda_{u}\cL_{u} + \lambda_{\text{align}}\cL_{\text{align}}$ \vspace{2mm}
        \STATE \COMMENT{\textbf{Update parameters.}} \vspace{1mm}
        \STATE Update $\theta$ and $\phi$ to minimize $\cL_{\text{DASO}}$ via SGD optimizer. 
        \STATE $\theta^{\prime} \leftarrow \rho\theta^{\prime} + (1-\rho)\theta$ \quad \COMMENT{Update the parameters of momentum encoder.}
        \STATE $t \leftarrow t+1$ \vspace{1mm}
    \end{algorithmic}
\end{algorithm*}

%% file: supple/detailed_setup.tex
\section{Detailed Experimental Setup}
\label{sec:sup_setup}
\subsection{Benchmarks}
\label{sec:sup_data_detail}
In this work, we evaluate both cases of $(i)$ labeled data and unlabeled data shares the same class distribution (\eg, $\gamma_l=\gamma_u$), and $(ii)$ the class distribution of unlabeled data can be different from the labeled data in various degree (\eg, $\gamma_l \neq \gamma_u$). 

\vspace{1mm}
\noindent \textbf{CIFAR-10 and CIFAR-100.} 
CIFAR benchmarks~\cite{krizhevsky2009learning} originally have the same number of examples per class; 5000 and 500 examples in $32\times 32$ sized image for CIFAR-10 and CIFAR-100, respectively. We use the head class size $N_1$ and imbalance ratio of labels $\gamma_l$ to craft the \emph{synthetically long-tailed} variants across the level of imbalance and total amount of labels, following the protocol from~\cite{kim2020distribution}. The number of examples other than the head class is calculated by $N_k=N_1\cdot \gamma_l^{-\frac{k-1} {K-1}}$ as proposed by~\cite{cui2019class}. Note that each $N_k$, the number of examples in class $k$ is sorted in a descending order (\ie, $N_1 \geq \cdots \geq N_K$). Similarly, the number of examples per class for the unlabeled data can be determined by: $M_k=M_1\cdot \gamma_u^{-\frac{k-1} {K-1}}$ using the labels, and the true labels are thrown away before training. 
We call those variants as CIFAR10/100-LT, which consist of labeled and unlabeled splits. We measure the performance on the test data, which have $10k$ examples in total for both data. 

\vspace{1mm}
\noindent \textbf{STL-10.} 
To generate STL10-LT: a \emph{long-tailed} variant of STL-10~\cite{coates2011analysis}, we follow the same process as explained in above. Besides the $5k$ labeled examples, STL-10 contains additional $100k$ unlabeled examples from a similar but broader distribution compared to the labeled data. Since the information about the class distribution of the unlabeled data is not known, we only construct the imbalanced labeled data and use the whole $100k$ unlabeled examples for training. 

\vspace{1mm}
\noindent \textbf{Semi-Aves.}
We also consider Semi-Aves benchmark~\cite{su2021semisupervised} for more realistic scenarios. \emph{Semi-Aves} includes $1k$ species of birds sampled from the \emph{iNaturalist-2018}~\cite{van2018inaturalist} with \emph{long-tailed} class distribution. Moreover, only 200 species are considered \emph{in-class}, and the other 800 species correspond to the \emph{out-of-class} (\ie, novel, open-set) categories for the unlabeled data. For \emph{in-class} examples, about $4k$ examples are labeled ($\cX$), and the other $27k$ examples are unlabeled ($\cU_{\text{in}}$). Note that the class distribution of labeled data does not match that of $\cU_{\text{in}}$ ($\gamma_l \neq \gamma_u$), as illustrated in~\cite{su2021semisupervised}. The \emph{out-of-class} unlabeled data ($\cU_{\text{out}}$) have $122k$ examples in total. \emph{Semi-Aves} benchmark provides $2k$ images and $8k$ images (\ie, 10 images and 40 images per class) for the validation and test data, respectively. 
We combine the labeled training data and validation data, $6k$ in total, for the labeled training data in our experiments, following~\cite{su2021realistic}. As note, we do not make any distinction between $\cU_{\text{in}}$ and $\cU_{\text{out}}$ when learning on the whole unlabeled data ($\cU=\cU_{\text{in}}+\cU_{\text{out}}$). 

\subsection{Training Details}
\label{sec:train_detail}
\noindent \textbf{CIFAR10/100-LT and STL10-LT.} 
Following the training protocol in~\cite{kim2020distribution}, we train a Wide ResNet-28-2~\cite{zagoruyko2016wider} with 1.5M parameters for $250k$ iterations. 
We set the batch size of the labeled data as 64, and the network is optimized via Nesterov SGD with momentum 0.9 and weight decay 5e-4. For the methods with using only labels, the base learning rate is set to 0.1 with linear warm-up applied during the first 2.5\% of the total train steps, and it decays after 80\% and 90\% of the training phase by a factor of 100, respectively, following~\cite{cao2019learning}. 
For SSL methods, we set the base learning rate as 0.03, which is fixed during the training. 
For the exponential moving average (EMA) network parameters for evaluation, the decay ratio $\rho$ is set to 0.999.
We further clarify the details for each method, such as hyper-parameters in~\cref{sec:impl_detail}. 
We measure the performance every 500 iterations (\eg, considered as 1 epoch), and report the median value in last 20 evaluations.  
\vspace{1mm}

\noindent \textbf{Semi-Aves.}
We train ResNet-34~\cite{he2016deep} with 21.3M parameters pre-trained on ImageNet~\cite{deng2009imagenet}. 
For the Supervised method, we train for 90 epochs of the labeled data, while we train 90 epochs of unlabeled data for SSL methods, using SGD optimizer with momentum 0.9. 
The base learning rate is set to 0.1 and 0.04 for the Supervised and SSL method each, with the linear warm-up for the first 5 epochs and it decays after 30 and 60 epochs, by a factor of 10. 
We set the labeled batch size as 256. 
All training images are randomly cropped and re-scaled to $224 \times 224$ size with random horizontal flip. 
The EMA decay ratio is $\rho=0.9$. The hyper-parameters of the individual method is described in~\cref{sec:impl_detail}.

\subsection{Implementation Details}
\label{sec:impl_detail}

\noindent \textbf{DASO.} 
$T_{\text{dist}}$, for scaling the empirical pseudo-label distribution, is chosen out of $\{0.3, 0.5, 1.0, 1.5 \}$. 
Specifically, for CIFAR10-LT, $T_{\text{dist}}=1.5$ in case of $\gamma_l=\gamma_u$, while $T_{\text{dist}}=0.3$ in the case of $\gamma_l \neq \gamma_u$.
For the other hyper-parameters, $T_{\text{proto}}=0.05$, $L=256$, and $\lambda_{\text{align}}=1$, which are kept unchanged during experiments.
The ablation study for those parameters is provided in~\cref{sec:more_abl}. 
We start applying DASO with $\cL_{\text{align}}$ after a few pre-training steps $P=5000$ to avoid unconfident predictions in the early stage of training. 
For empirical pseudo-label distribution $\hat{m}$, we accumulate the class predictions of the final pseudo-labels $\hat{p}^{\prime}$ every 100 iterations on CIFAR10/100-LT and STL10-LT. 
For Semi-Aves, we set $P=20$ epochs and update $\hat{m}$ every epoch. 
For the EMA decay ratio $\rho$ for prototype generation, we simply use the same parameter of the one for evaluation. 
\Cref{tab:daso_param} summarizes the training details of DASO. 
\input{supple/tables/daso_param}
\vspace{1mm}

\noindent \textbf{Supverised.}
The only labeled data is trained via standard cross-entropy loss $\cH$. The training protocol and hyper-parameters (total iterations, learning rate, optimizer, and etc.) are described in~\cref{sec:train_detail}. 
\vspace{1mm}

\noindent \textbf{Re-weighting with the Effective Number of Samples}~\cite{cui2019class}. 
The per-class weights are applied to the cross-entropy loss based on the effective number of samples.
\begin{equation}
    E_{N_k} = \frac{1 - \beta^{N_k}} {1 - \beta},
\end{equation}
where $N_k$ corresponds to the number of samples in class $k$, and then the weight for class $k$ is set to be proportional to the inverse of the effective number $E_{N_k}$. $\beta$ is a hyper-parameter, which is set to 0.999 during the experiments. 
\vspace{1mm}

\noindent \textbf{LDAM-DRW}~\cite{cao2019learning}. 
Decision boundary of the classifier takes up more margin in rare classes, using LDAM loss:
\begin{equation}
\label{eqn:ldam}
        \cL_{LDAM}   = -\, \log\, \frac{ e^{ z_{y_k} - \Delta_{y_k} } } {  e^{ z_{y_k} - \Delta_{y_k} } + \sum_{j \neq {y_k}} e^{z_j} },\; \textnormal{where } \Delta_k \propto \frac{1} {N^{1/4}_k}.
\end{equation}
Then it adopts deferred re-weighting scheme (DRW) to apply re-balancing algorithm in later stage of training. Following DRW scheme, we apply re-weighting objective~\cref{eqn:ldam} after 200k iterations. 

\vspace{1mm}
\noindent \textbf{cRT}~\cite{Kang2020Decoupling}. 
After training the entire network under imbalanced distribution, the classifier is re-trained with the parameters of the feature encoder fixed for a balanced objective. We first train a model with cross-entropy loss. In classifier re-training phase, we simply re-weight the cross-entropy loss with the weights based on the effective number of samples~\cite{cui2019class} for $100k$ iterations. The learning rate schedule under re-training phase is proportionally adjusted.

\vspace{1mm}
\noindent \textbf{Logit Adjustment (LA)}~\cite{menon2021longtail}. 
Logits are adjusted by enforcing a large margin for the minority classes compared to the majority ones in either two ways: \emph{post-hoc adjustment} or \emph{logit-adjusted} cross-entropy, based on the class frequency of labels. In this work, we adopt the latter strategy. Before measuring cross-entropy for the labeled data, each logit is adjusted by:
\begin{equation}
    p_k \leftarrow p_k + \tau \log {n}_k,
\end{equation}
where $p=f(x)$ and ${n}_k$ denotes the class label frequency value in class $k$. $\tau=1$ is a temperature scaling factor.

\vspace{1mm}
\noindent \textbf{PseudoLabel}~\cite{lee2013pseudo}. 
The one-hot pseudo-label $\hat{p}$ from $p=f(u)$ regularizes the unlabeled example. Only the predictions with the highest probability value above a certain threshold $\tau$ contribute to the regularizer. We set $\tau$ to 0.95. 
\begin{equation}
    \Phi_u(\hat{p},\,p) = \mathbbm{1}\left(\max_k p_k \geq \tau\right)\,\cH \left(\hat{p},\,p \right), 
\end{equation}
where $\hat{p} = \text{OneHot}(\argmax_k p_k)$. We set the loss weight $\lambda_u=1$ and apply linear ramp-up with the ratio of 0.4; $\lambda_u$ linearly increases starting from 0 and attains the maximum value ($\lambda_u=1$) at 40\% of the total iterations. 
\vspace{1mm}

\noindent \textbf{MeanTeacher}~\cite{tarvainen2017mean}. 
The momentum encoder $f^{\text{EMA}}=f^{\text{cls}}_{\phi^\prime} \circ f^{\text{enc}}_{\theta^\prime}$ generates the target for the prediction of unlabeled data, where $\phi^\prime$ and $\theta^\prime$ are the momentum-updating network parameters of linear classifier and feature encoder, respectively.
\begin{equation}
    \Phi_u(\hat{p},\, p) = \norm{\sigma(\hat{p}) - \sigma(p)}^2,\;\textnormal{where}\; \hat{p} = f^{\text{EMA}}(\cA_w(u))\; \textnormal{and}\; p=f(\cA_w(u)). 
\end{equation}
We set the EMA decay ratio $\rho=0.999$. $\lambda_u$ is set to 50, applying the linear ramp-up with the ratio of 0.4.
\vspace{1mm}

\noindent \textbf{MixMatch}~\cite{berthelot2019mixmatch}.
Pseudo-label is produced from the multiple augmentations of the same image with entropy regularization. Then the model learns mixup~\cite{zhang2018mixup} images and (pseudo-) labels over the whole labeled and unlabeled data. We use the number of augmentations as 2, temperature scaling factor as 0.5, and the sampling hyper-parameter for mixup regularization $\alpha$ as 0.5. We also apply linear ramp-up strategy for $\lambda_u$, where it attains its maximum value 100 with the ratio of 0.016. 
\vspace{1mm}

\noindent \textbf{ReMixMatch}~\cite{berthelot2020remixmatch}. 
It adds up two techniques of \emph{Augmentation Anchoring} and \emph{Distribution Alignment} over MixMatch~\cite{berthelot2019mixmatch}. 
We use the advanced augmentation as RandAugment~\cite{cubuk2020randaugment} followed by Cutout~\cite{devries2017improved}. Considering the computational cost, we set the number of advanced augmentations as $\mu=2$. For the others, we set the temperature scaling factor for pseudo-labels as 0.5, and $\alpha$ as 0.75. The weights for pre-mixup loss and rotation loss are both set to 0.5. For $\lambda_u$, the linear ramp-up ratio is set to 0.016 with $\lambda_u=1.5$. We apply weak augmentations for convenience for the labeled data, instead of advanced augmentation. 

\vspace{1mm}
\noindent \textbf{FixMatch}~\cite{sohn2020fixmatch}.
One-hot pseudo-labels are generated from weakly augmented images as the same with PseudoLabel~\cite{lee2013pseudo}, then they provide the targets for the predictions from strong augmentations of the same images to the cross-entropy loss $\cH$:
\begin{equation}
    \Phi_u(\hat{p},\,p^{{(s)}}) = \mathbbm{1}\left(\max_k p^{{(w)}}_k \geq \tau\right)\,\cH \left(\hat{p},\,p^{{(s)}} \right), 
\end{equation}
where $\hat{p}=\textnormal{OneHot}\left(\argmax_k p^{{(w)}}_k \right)$ with $p^{{(w)}}=f\left(\cA_w(u) \right)$ and $p^{{(s)}}=f\left(\cA_s(u) \right)$. 
We use RandAugment~\cite{cubuk2020randaugment} for the advanced augmentation. For fair comparisons to ReMixMatch~\cite{berthelot2020remixmatch}, we use the unlabeled batch ratio $\mu$ as 2. For the other hyper-parameters, $\lambda_u$ is set to 1 without applying linear ramp-up strategy.
\vspace{1mm}

\noindent \textbf{USADTM}~\cite{han2020unsupervised}.
It combines \emph{unsupervised semantic aggregation} (USA); a clustering objective in unlabeled data and \emph{deformable template matching} (DTM); assigning a semantic pseudo-label to each unlabeled example solely from feature-space. 
The semantic pseudo-label is determined by the agreement of two different distance measure from a sample to each class prototypes constructed from the labeled data. 
In our experiments, we use the loss weight for the mutual information loss $\alpha=0.1$ and $\tau=0.85$ for the confidence threshold, following~\cite{han2020unsupervised}. 
We note that \cite{han2020unsupervised} keeps some confident unlabeled examples to treat them as labeled examples to enforce cross-entropy loss due to the limited labels (\ie, 4 labels per class). 
This would also help generally in \emph{imbalanced} SSL, but we do not adopt this strategy in our experiments in order to fairly comparing with other SSL methods focusing on the aspect of \emph{pseudo-labeling} method. 
\vspace{1mm}

\noindent \textbf{BOSS}~\cite{smith2020building}.
This originally proposes to apply three techniques altogether on FixMatch~\cite{sohn2020fixmatch} to achieve state-of-the-art performance on CIFAR-10 benchmark under one label per class: \emph{prototype (single-example per class) refining}, \emph{pseudo-label re-balancing}, and \emph{self-training iterations}. We only adopt \emph{pseudo-label re-balancing} method from the original paper for fairly comparing under \emph{imbalanced} SSL. \emph{Pseudo-label re-balancing} includes adjusting loss weights and confidence thresholds based on the class distribution of predicted pseudo-labels on top of the FixMatch loss: 
\begin{equation}
    \label{eqn:sup_boss}
    \Phi_u(\hat{p},\,p^{{(s)}}) = \mathbbm{1}\left(\max_k p^{{(w)}}_k \geq \tau_k \right)\,\frac{1} {Z \cdot \hat{c}_k} \cH\left(\hat{p},\,p^{{(s)}} \right),
\end{equation}
where $\tau_k$ is the class-dependent confidence threshold defined as:
\begin{equation}
    \tau_k=\tau - \Delta \cdot \left(1 - \frac{\hat{c}_k} {\max_k \hat{c}_k} \right),
\end{equation}
and $\hat{c}_k$ is the number of predicted pseudo-labels in the current batch for class $k$. We fix $\Delta=0.25$ during the experiments. 
Note that the scale of $\Phi_u$ is adjusted by a factor of $Z$ to consistently maintain the relative scale of $\lambda_u$. 
\vspace{1mm}

\noindent \textbf{DARP}~\cite{kim2020distribution}.
The class distribution of the predicted pseudo-labels is explicitly adjusted to the \emph{given} class priors via solving a convex optimization problem. In our experiments, we use the class prior as the class label frequency in case of $\gamma_l=\gamma_u$ for CIFAR10-LT and CIFAR100-LT, and in case of Semi-Aves benchmark. In other cases, \ie, $\gamma_l \neq \gamma_u$, we estimate the distribution of the unlabeled data (\eg, $M_k$) using held-out validation set, following~\cite{kim2020distribution}. We start applying DARP at $100k$ iterations of training with refining pseudo-labels every 10 steps. We use $\alpha=2.0$ for removing the noisy entries.
\vspace{1mm}

\noindent \textbf{CReST}~\cite{wei2021crest}.
Self-training is adopted where a SSL algorithm is \emph{iteratively re-trained} with adding some acceptable pseudo-labeled samples to the labeled data. 
The relative ratio of pseudo-labeled samples that will be added to the labeled set in next generation for each class $k$ is defined as: $\mu_k=\left(N_{K+1-k} \mathbbm{/} N_1 \right)^{\alpha}$, where $N_k$ is the label size for class $k$, suggesting that minority-class pseudo-labels are more likely to be added.
In CReST+, it adds the progressive distribution alignment (PDA) to the CReST method. 
To fairly compare with other baselines with 250$k$ of the maximum iterations in total, we divide the whole iterations to 5 generations, where each generation trains 50$k$ iterations for CIFAR10/100-LT and STL10-LT. 
For Semi-Aves, we divide the whole 90 epochs to 3 generations of 30 epochs. 
For CIFAR10/100-LT and STL10-LT, we set $\alpha=1/3$ and $t_{\text{min}}=0.5$, and $\alpha=0.7$ and $t_{\text{min}}=0.5$ for Semi-Aves respectively similar to~\cite{wei2021crest}.

\vspace{1mm}
\noindent \textbf{ABC}~\cite{lee2021auxiliary}.
It trains an auxiliary balanced classifier (ABC) built upon a whole SSL learner (\eg, FixMatch~\cite{sohn2020fixmatch}). 
In particular, ABC shares the feature extractor with the existing pipeline, and learns the re-weighted versions of both cross-entropy with labels and \emph{consistency regularization} from unlabeled data.
The re-weight mechanism is performed by the balanced batch of labeled data and unlabeled data, where the batched images corresponding to each labels and predicted pseudo-labels are dropped with a probability sampled from Bernoulli distribution. Here, the parameter for Bernoulli is inversely proportional to the class frequency of the labels and pseudo-labels respectively. 
The ABC classifier is opted during inference. 

%% file: supple/tables/daso_param.tex
\begin{table}[ht]
\centering
\resizebox{0.55\linewidth}{!}{%
\begin{tabular}{ccccc}
    \toprule
    parameter & CIFAR10-LT & CIFAT100-LT & STL10-LT & Semi-Aves \\
    \midrule
    \emph{lr} & \multicolumn{3}{c}{0.03} & 0.04 \\
    \cmidrule(r){1-1} \cmidrule(lr){2-4} \cmidrule(l){5-5}
    $B$ & \multicolumn{3}{c}{64} & 256 \\
    \cmidrule(r){1-1} \cmidrule(lr){2-4} \cmidrule(l){5-5}
    $\mu$ & \multicolumn{3}{c}{2} & 5 \\
    \cmidrule(r){1-1} \cmidrule(lr){2-4} \cmidrule(l){5-5}
    SGD momentum & \multicolumn{3}{c}{0.9} & 0.9 \\
    \cmidrule(r){1-1} \cmidrule(lr){2-4} \cmidrule(l){5-5}
    Nesterov & \multicolumn{3}{c}{True} & True \\
    \cmidrule(r){1-1} \cmidrule(lr){2-4} \cmidrule(l){5-5}
    weight decay & \multicolumn{3}{c}{5e-4} & 3e-4 \\
    \cmidrule(r){1-1} \cmidrule(lr){2-4} \cmidrule(l){5-5}
    $L$ & \multicolumn{3}{c}{256} & 256   \\
    \cmidrule(r){1-1} \cmidrule(lr){2-4} \cmidrule(l){5-5}
    $\rho$ &  \multicolumn{3}{c}{0.999} & 0.9         \\
    \cmidrule(r){1-1} \cmidrule(lr){2-4} \cmidrule(l){5-5}
    $T_{\text{proto}}$ & \multicolumn{3}{c}{$0.05$} & 0.05 \\
    \cmidrule(r){1-1} \cmidrule(lr){2-4} \cmidrule(l){5-5}
    $\lambda_{\text{align}}$ & \multicolumn{3}{c}{1.0} & 1.0 \\
    \cmidrule(r){1-1} \cmidrule(lr){2-4} \cmidrule(l){5-5}
    $P$ & \multicolumn{3}{c}{5000 \emph{steps}} & 20 \emph{epochs} \\
    \cmidrule(r){1-1} \cmidrule(lr){2-4} \cmidrule(l){5-5}
    $T_{\text{dist}}$ & $\{1.5, 0.3\}$ & 0.3 & 0.3 & 0.5 \\
    \bottomrule
\end{tabular}%
}
\caption{A complete list of training details for DASO framework.}
\label{tab:daso_param}
\end{table}

%% file: supple/additional_exp.tex
\input{supple/tables/new_rw}
\section{Additional Experiments}
\label{sec:sup_exp}
\subsection{Comprehensive Comparison with More Baselines}
\label{sec:sup_comprehensive}
Experiments from the main paper evaluated DASO and other baseline methods specifically designed for \emph{re-balancing the biased pseudo-labels} under class-imbalanced labels and distribution mismatch between $\cX$ and $\cU$. 
In~\Cref{tab:new_rw}, we introduce more diverse baseline methods for comparisons across different benchmarks including both $\gamma_l=\gamma_u$ and $\gamma_l \neq \gamma_u$ cases. 
As following, we term SSL methods as SSL, label re-balancing methods as LB, and the re-balancing methods for pseudo-labels as PB from~\Cref{tab:new_rw}.
We consider \emph{LDAM-DRW}~\cite{cao2019learning}, \emph{classifier re-training (cRT)}~\cite{Kang2020Decoupling}, and \emph{class re-weighting with effective number of samples (CB re-weight)}~\cite{cui2019class} for LB, respectively. 
For SSL methods, we additionally introduce \emph{PseudoLabel}~\cite{lee2013pseudo} and \emph{USADTM}~\cite{han2020unsupervised}.
We further consider \emph{BOSS}~\cite{smith2020building} as PB.
The implementation details on those methods are explained in~\cref{sec:impl_detail}. 
Note that we extensively compare PB methods based on other than FixMatch in~\Cref{tab:compare_remix}. 

We observe in~\Cref{tab:new_rw} that applying LB improves the performance for \emph{Supervised} and semi-supervised (SSL, PB) learning methods in general. 
This suggests that the bias of pseudo-label can be reduced by LB methods. 
In particular, the performance of DASO can be further pushed by additionally applying LB methods, as noted from \emph{CB re-weight + DASO} and \emph{LA + DASO}. 
This verifies that DASO is complementary to the existing LB methods, where the source for the performance improvement of DASO itself comes from the ability to \emph{truly} alleviate the bias of pseudo-labels, not just re-balancing the labels.  
\vspace{1mm}

\subsection{DASO with Label Re-Balancing when $\gamma_l \neq \gamma_u$}
\label{sec:sup_stl}
We further evaluate DASO combined with other re-balancing techniques: LA~\cite{menon2021longtail} and ABC~\cite{lee2021auxiliary}, when the class distribution of unlabeled data significantly differs from the labeled data (\eg, $\gamma_l \neq \gamma_u$). 
In this setup, we conduct experiments with STL10-LT, as shown in~\Cref{tab:rebal_stl}. 
\input{supple/tables/rebalancing_stl}

We observe that both LA~\cite{menon2021longtail} and ABC~\cite{lee2021auxiliary}, are beneficial upon baseline FixMatch.
Moreover, the performance can be further pushed when DASO is applied on top of those methods.
However, the performances show marginal improvements compared to the FixMatch w/ DASO. 
This opens a new challenge that calls for the design of an unified re-balancing approach of labels and unlabeled data, which can also well address the potentially unknown unlabeled data.

\subsection{Comparison based on ReMixMatch}
\label{sec:sup_remixmatch}
To verify the efficacy of DASO as a \emph{generic framework}, we further compare the pseudo-label re-balancing (PB) methods based on  ReMixMatch~\cite{berthelot2020remixmatch}. 
In particular, we provide the results as the same way when DASO is integrated with FixMatch~\cite{sohn2020fixmatch} from the main paper. 
\Cref{tab:compare_remix} shows the results. We compare each method on CIFAR10/100-LT and STL10-LT, varying the imbalance ratio while the amount of labels used are fixed by $N_1$. Note that for CIFAR benchmarks, $\gamma=\gamma_l=\gamma_u$.   

\input{supple/tables/compare_remix}

As can be seen, DASO achieves the best results among the baselines for comparison.
From CIFAR benchmarks (\eg, $\gamma_l=\gamma_u$), DASO outperforms both DARP~\cite{kim2020distribution} and CReST+~\cite{wei2021crest} that leverages the assumption of $\gamma_l=\gamma_u$ explicitly; for example, they utilize the actual class distribution of unlabeled data. 
As note, while CReST+ is beneficial for ReMixMatch when trained on CIFAR10-LT, but it performs worse in CIFAR100-LT results. 
This might come from the limited amount of labels and the repeated training with re-initializing models via self-training. 
For STL10-LT cases, the improvements from both DARP and CReST+ can be limited due to the mismatch of class distributions between the labeled data and unlabeled data.
In contrary, DASO significantly surpasses the other methods without the access to the class distribution of either labels or unlabeled data. 
To summarize, DASO can improve typical baseline SSL methods under imbalanced data \emph{in general}.

\subsection{Results on Test-Time Logit Adjustment}
In the main paper, we have considered Logit Adjustment (LA)~\cite{menon2021longtail} as applying logit-adjusted cross-entropy loss during training. 
This point is also explained in \cref{sec:impl_detail}. 
On the other hand, we also consider adjusting the logits during inference also present in~\cite{menon2021longtail}; we denote this type of LA as \emph{LA (inf)}.
In \Cref{tab:test_la}, we report the results obtained from LA~\cite{menon2021longtail} by this strategy when the class distribution of labeled data and unlabeled data are identical ($\gamma=\gamma_l=\gamma_u$).
\input{supple/tables/test_la}

\subsection{More Ablation Study}
\label{sec:more_abl}
We conduct several ablation studies on the hyper-parameters in DASO framework. As the same with the ablation study conducted from the main paper, we consider FixMatch~\cite{sohn2020fixmatch} with DASO on CIFAR10-LT with $N_1=500$, $\gamma=100$ (denoted as C10) and STL10-LT with $N_1=150$, $\gamma_l=10$ (denoted as STL10) respectively. \Cref{tab:abl_queue_size} compares different values of the queue size $L$ for constructing the \emph{balanced} prototypes. \Cref{tab:abl_temp_proto} tests different temperature factor $T_{\text{proto}}$ for the similarity-based classifier. Finally, \Cref{tab:abl_lamb_align} shows the effect of different loss weights $\lambda_{\text{align}}$ for the semantic alignment loss. We shaded rows that correspond to the hyper-parameter of the complete DASO framework. We also indicate the best results in bold. 

\input{supple/tables/new_abl}

As note, we do not tune the hyper-parameters above ($L$, \,$T_{\text{proto}}, \,\lambda_{\text{align}}$) depending on different benchmarks across different imbalance ratio. For example, in STL10-LT case, using $\lambda_{\text{align}}$ value higher than 1 seems effective, but the result of 70.21\% obtained from $\lambda_{\text{align}}=1$ already performs well. 

%% file: supple/tables/new_rw.tex
\begin{table*}[t]
\centering
 \resizebox{0.97\linewidth}{!}{%
    \begin{tabular}{lccccccccc}
    \toprule
         & \multicolumn{3}{c}{\multirow{2}{*}{Method type}} & \multicolumn{2}{c}{CIFAR10-LT} & \multicolumn{2}{c}{CIFAR100-LT} & \multicolumn{2}{c}{STL10-LT} \\
        &  &  &  & \multicolumn{2}{c}{$\gamma=\gamma_l=\gamma_u=100$}  & \multicolumn{2}{c}{$\gamma=\gamma_l=\gamma_u=10$} & \multicolumn{2}{c}{$\gamma_l=10$, $\gamma_u$: \emph{unknown}} \\
        \cmidrule(lr){2-4} \cmidrule(lr){5-6} \cmidrule(lr){7-8} \cmidrule(l){9-10}
        \multirow{2}{*}{Algorithm} & \multirow{2}{*}{SSL} & \multirow{2}{*}{LB} & \multirow{2}{*}{PB} & $N_1=500$ & $N_1=1500$ & $N_1=50$ & $N_1=150$ & $N_1=150$ & $N_1=450$ \\
        &  &  &  & $M_1=4000$ & $M_1=3000$ & $M_1=400$ & $M_1=300$ & $M=100k$ & $M=100k$ \\
       \cmidrule(r){1-1} \cmidrule(lr){2-4} \cmidrule(lr){5-6} \cmidrule(lr){7-8} \cmidrule(l){9-10}
        Supervised &  &  &  & \ms{47.3}{0.95} & \ms{61.9}{0.41} & \ms{29.6}{0.57} & \ms{46.9}{0.22} & \ms{40.2}{1.80} & \ms{60.4}{1.91} \\
        ~~w/ LDAM-DRW~\cite{cao2019learning} &  &  \checkmark &  & \ms{50.1}{1.55} & \ms{65.7}{1.49} & \ms{28.4}{0.32} & \ms{46.2}{0.46} & \ms{41.8}{3.05} & \ms{62.1}{1.39}\\
        ~~w/ cRT~\cite{Kang2020Decoupling} &  &  \checkmark &  & \ms{49.5}{1.05} & \ms{65.8}{0.47} & \ms{30.1}{0.50} & \ms{48.0}{0.43} & \ms{40.8}{1.95} & \ms{61.6}{1.83} \\
        ~~w/ LA~\cite{menon2021longtail} &  &  \checkmark &  & \ms{53.3}{0.44} & \ms{70.6}{0.21} & \ms{30.2}{0.44} & \ms{48.7}{0.89}  & \ms{42.8}{1.78} & \ms{63.1}{1.13} \\
        \cmidrule(r){1-1} \cmidrule(lr){2-4} \cmidrule(lr){5-6} \cmidrule(lr){7-8} \cmidrule(l){9-10}
        PseudoLabel~\cite{lee2013pseudo}  & \checkmark &  &  & \ms{47.8}{1.06} & \ms{63.4}{0.81} & \ms{30.7}{0.18} & \ms{47.8}{0.40} & \ms{42.3}{0.83} & \ms{60.4}{1.11} \\
        USADTM~\cite{han2020unsupervised} & \checkmark &  &  & \ms{72.9}{0.74} & \ms{73.3}{0.39} & \ms{48.7}{1.00} & \ms{58.2}{0.79} & \ms{68.9}{1.83} & \ms{77.1}{0.74} \\
        \cmidrule(r){1-1} \cmidrule(lr){2-4} \cmidrule(lr){5-6} \cmidrule(lr){7-8} \cmidrule(l){9-10}
        FixMatch~\cite{sohn2020fixmatch} & \checkmark &  &  & \ms{67.8}{1.13} & \ms{77.5}{1.32} & \ms{45.2}{0.55} & \ms{56.5}{0.06} & \ms{56.1}{2.32} & \ms{72.4}{0.71} \\
        ~~w/ CB re-weight~\cite{cui2019class} & \checkmark & \checkmark &  & \ms{72.2}{1.28} & \ms{80.9}{1.52} & \ms{46.0}{0.27} & \ms{58.3}{0.46} & \ms{58.9}{2.79} & \ms{74.7}{0.55} \\
        ~~w/ LA~\cite{menon2021longtail} & \checkmark & \checkmark &  & \ms{75.3}{2.45} & \msbb{82.0}{0.36} & \ms{47.3}{0.42} & \ms{58.6}{0.36}  & \ms{63.4}{2.99} & \ms{75.9}{1.25}  \\
        \cmidrule(r){1-1} \cmidrule(lr){2-4} \cmidrule(lr){5-6} \cmidrule(lr){7-8} \cmidrule(l){9-10}
        ~~w/ BOSS~\cite{smith2020building} & \checkmark &  & \checkmark & \ms{70.3}{0.87} & \ms{76.5}{0.66} & \ms{50.0}{0.39} & \ms{59.3}{0.22} & \ms{66.4}{2.09} & \ms{76.0}{0.85} \\
        ~~w/ DARP~\cite{kim2020distribution} & \checkmark &  & \checkmark & \ms{74.5}{0.78} & \ms{77.8}{0.63} & \ms{49.4}{0.20} & \ms{58.1}{0.44} & \ms{66.9}{1.66} & \ms{75.6}{0.45} \\
        ~~w/ CReST~\cite{wei2021crest} & \checkmark &  & \checkmark & \ms{73.4}{3.10} & \ms{76.6}{1.23} & \ms{44.3}{0.77} & \ms{57.1}{0.58} & \ms{61.7}{2.51} & \ms{71.6}{1.17} \\
        ~~w/ CReST+~\cite{wei2021crest} & \checkmark &  & \checkmark & \ms{76.3}{0.86} & \ms{78.1}{0.42} & \ms{44.5}{0.94} & \ms{57.1}{0.65} & \ms{61.2}{1.27} & \ms{71.5}{0.96} \\
        ~~w/ DASO (Ours) & \checkmark &  & \checkmark & \ms{76.0}{0.37} & \ms{79.1}{0.75} & \ms{49.8}{0.24} & \ms{59.2}{0.35} & \ms{70.0}{1.19} & \msbb{78.4}{0.80} \\
        \cmidrule(r){1-1} \cmidrule(lr){2-4} \cmidrule(lr){5-6} \cmidrule(lr){7-8} \cmidrule(l){9-10}
        ~~w/ CB re-weight + DASO (Ours) & \checkmark & \checkmark & \checkmark & \msbb{77.3}{0.86} & \ms{81.2}{0.77} & \msbb{50.3}{0.18} & \msbb{60.1}{0.12} & \msbb{70.2}{1.05} & \ms{77.8}{0.58} \\
        ~~w/ LA + DASO (Ours) & \checkmark & \checkmark & \checkmark & \msb{77.9}{0.88} & \msb{82.5}{0.08} & \msb{50.7}{0.51} & \msb{60.6}{0.71} & \msb{71.3}{1.81} & \msb{79.0}{0.58} \\
        \bottomrule
    \end{tabular}%
}
\caption{Comparison of accuracy (\%) with different methods and their combinations on CIFAR10-LT, CIFAR100-LT, and STL10-LT under different label sizes with class imbalance. %
SSL denotes semi-supervised learning. LB and PB correspond to re-balancing for labels and pseudo-labels, respectively. 
Our DASO shows consistent performance gain over the baseline FixMatch~\cite{sohn2020fixmatch}, and adding label re-balancing to our method shows the best performance among the baselines. CIFAR10/100-LT benchmarks represent the $\gamma_l=\gamma_u$ setup, and STL10-LT corresponds to $\gamma_l \neq \gamma_u$ setup. We indicate the best results in bold and the second-best results with underlined. 
}
\label{tab:new_rw}
\end{table*}

%% file: supple/tables/rebalancing_stl.tex
\begin{table}[ht]
\centering
 \resizebox{0.55\linewidth}{!}{%
    \begin{tabular}{lcccc}
    \toprule
     & \multicolumn{4}{c}{STL10-LT ($M=100k$)} \\
     & \multicolumn{2}{c}{$\gamma_l=10$} & \multicolumn{2}{c}{$\gamma_l=20$} \\
    \cmidrule(lr){2-3} \cmidrule(l){4-5} 
    Algorithm & $N_1=150$ & $N_1=450$ & $N_1=150$ & $N_1=450$ \\
    \cmidrule(r){1-1} \cmidrule(lr){2-3} \cmidrule(l){4-5}
    FixMatch~\cite{sohn2020fixmatch}  & \ms{56.1}{2.32} & \ms{72.4}{0.71} & \ms{47.6}{4.87} & \ms{64.0}{2.27} \\
    ~~w/ DASO (Ours) & \ms{70.0}{1.19} & \ms{78.4}{0.80} & \msb{65.7}{1.78} & \ms{75.3}{0.44} \\
    \cmidrule(r){1-1} \cmidrule(lr){2-3} \cmidrule(l){4-5}

    FixMatch w/ LA~\cite{menon2021longtail} & \ms{64.4}{1.35} & \ms{75.9}{1.25} & \ms{51.5}{3.23} & \ms{67.4}{1.04} \\
    ~~w/ DASO (Ours) & \msb{71.7}{1.09} & \msb{79.0}{0.58} & \ms{65.6}{1.43} & \msb{75.8}{0.81} \\    
    \cmidrule(r){1-1} \cmidrule(lr){2-3} \cmidrule(l){4-5}

    FixMatch + ABC~\cite{lee2021auxiliary} & \ms{66.3}{1.00} & \ms{77.1}{0.56} & \ms{59.3}{2.66} & \ms{73.0}{0.91} \\
    ~~w/ DASO (Ours) & \ms{69.6}{0.94} & \ms{77.9}{0.89} & \ms{64.5}{2.81} & \ms{74.7}{0.16} \\
    \bottomrule
    \end{tabular}%
}
\caption{
    Comparison of accuracy (\%) with the combination of various re-balancing methods on $\gamma_l \neq \gamma_u$ setup. 
    DASO somewhat obtains performance gain when even combined with either LA~\cite{menon2021longtail} or ABC~\cite{lee2021auxiliary} on FixMatch. 
    We indicate the best results as bold.  
}
\label{tab:rebal_stl}
\end{table}

%% file: supple/tables/compare_remix.tex
\begin{table}[ht]
\centering
 \resizebox{0.7\linewidth}{!}{%
    \begin{tabular}{lcccccc}
    \toprule
     & \multicolumn{2}{c}{CIFAR10-LT} & \multicolumn{2}{c}{CIFAR100-LT} & \multicolumn{2}{c}{STL10-LT} \\
     & \multicolumn{2}{c}{$N_1=500, M_1=4000$.} & \multicolumn{2}{c}{$N_1=50, M_1=400$.} & \multicolumn{2}{c}{$N_1=150$.} \\
    \cmidrule(lr){2-3} \cmidrule(lr){4-5} \cmidrule(l){6-7} 
    Algorithm & $\gamma=100$ & $\gamma=150$ & $\gamma=10$ & $\gamma=20$ & $\gamma_l=10$ & $\gamma_l=20$ \\
    \cmidrule(r){1-1} \cmidrule(lr){2-3} \cmidrule(lr){4-5} \cmidrule(l){6-7} 
    ReMixMatch~\cite{berthelot2020remixmatch}   & \ms{70.9}{2.37} & \ms{64.7}{0.95} & \ms{52.3}{0.91} & \ms{46.5}{0.30} & \ms{54.4}{2.15} & \ms{46.5}{1.93} \\
    ~~w/ DARP~\cite{kim2020distribution}        & \ms{72.2}{2.72} & \ms{65.7}{1.20} & \ms{52.8}{0.65} & \ms{47.0}{0.17} & \ms{61.2}{2.62} & \ms{59.5}{2.56} \\
    ~~w/ CReST+~\cite{wei2021crest}             & \ms{75.6}{1.60} & \ms{65.9}{2.20} & \ms{49.9}{0.80} & \ms{44.5}{1.04} & \ms{64.1}{1.68} & \ms{49.2}{0.90} \\
    ~~w/ DASO (Ours)                            & \msb{76.8}{0.81} & \msb{68.5}{0.98} & \msb{53.6}{0.81} & \msb{47.8}{0.69} & \msb{75.0}{0.95} & \msb{68.5}{5.14} \\
     \bottomrule
    \end{tabular}%
}
\caption{Comparison of accuracy (\%) with various pseudo-label re-balancing (PB) methods upon different baseline SSL learner, ReMixMatch~\cite{berthelot2020remixmatch}. DASO outperforms all the other methods by a significant margin, which is consistent with the results when the baseline SSL learner was FixMatch from the main paper. We indicate the best results as bold. 
}
\label{tab:compare_remix}
\end{table}

%% file: supple/tables/test_la.tex
\begin{table}[ht]
\centering
 \resizebox{0.65\linewidth}{!}{%
    \begin{tabular}{lcccc}
    \toprule
     & \multicolumn{2}{c}{CIFAR10-LT} & \multicolumn{2}{c}{CIFAR100-LT} \\
     & \multicolumn{2}{c}{$N_1=1500$, $M_1=3000$} & \multicolumn{2}{c}{$N_1=50$, $M_1=300$} \\
    \cmidrule(lr){2-3} \cmidrule(l){4-5} 
    Algorithm & $\gamma=100$ & $\gamma=150$ & $\gamma=10$ & $\gamma=20$ \\
    \cmidrule(r){1-1} \cmidrule(lr){2-3} \cmidrule(l){4-5}
    FixMatch~\cite{sohn2020fixmatch}                    & \ms{77.5}{1.32} &  \ms{72.4}{1.03}  & \ms{56.5}{0.06} & \ms{50.7}{0.25} \\
    \cmidrule(r){1-1} \cmidrule(lr){2-3} \cmidrule(l){4-5}
    FixMatch w/ LA~\cite{menon2021longtail}             & \ms{82.0}{0.36}  & \ms{78.0}{0.91}  & \ms{58.6}{0.36} & \ms{53.4}{0.32} \\
    FixMatch w/ LA + CReST+~\cite{wei2021crest}         & \ms{81.1}{0.57}  & \ms{77.9}{0.71}  & \ms{57.1}{0.55} & \ms{52.3}{0.20} \\
    FixMatch w/ LA + DASO (Ours)                        & \ms{82.5}{0.08}  & \ms{79.0}{2.23}  & \msb{60.6}{0.71} & \ms{55.1}{0.72} \\
    \cmidrule(r){1-1} \cmidrule(lr){2-3} \cmidrule(l){4-5}
    FixMatch w/ LA (inf)~\cite{menon2021longtail}       & \ms{82.8}{1.43}  & \ms{79.2}{1.15}  & \ms{58.7}{0.63}  & \ms{53.3}{0.43} \\
    FixMatch w/ LA (inf) + CReST+~\cite{wei2021crest}   & \ms{82.9}{0.24}  & \ms{80.3}{0.56}  & \ms{57.8}{0.47}  & \ms{53.3}{0.83} \\
    FixMatch w/ LA (inf) + DASO (Ours)                  & \msb{84.5}{0.55} & \msb{81.8}{0.83} & \ms{60.5}{0.49} & \msb{55.2}{0.47} \\
    \bottomrule
    \end{tabular}%
}
\caption{Comparison of accuracy (\%) with different strategies of applying Logit Adjustment (LA)~\cite{menon2021longtail}: either train-time (noted as LA) or during inference (noted as LA (inf)). We observe large gains compared to baseline FixMatch when LA is applied during inference.
}
\label{tab:test_la}
\end{table}

%% file: supple/tables/new_abl.tex
\definecolor{Gray}{gray}{0.9}
\begin{table}[ht]
\centering
\begin{minipage}{0.29\columnwidth}
    \centering
    \resizebox{.73\columnwidth}{!}{%
        \begin{tabular}{lcc}
            \toprule
             & C10 & STL10 \\
            \cmidrule{1-1} \cmidrule{2-3}
            FixMatch & 68.25 & 55.53 \\
            \cmidrule{1-1} \cmidrule{2-3}
            $L=128$  & 73.77 & 69.17  \\
            \rowcolor{Gray}
            $L=256$  & \textbf{75.97} & \textbf{70.21}  \\
            $L=512$  & 75.03 & 69.96  \\
            $L=1024$ & 74.36 & 69.64 \\
            $L=2048$ & 73.50 & 69.99 \\
            \bottomrule
        \end{tabular}%
    }
    \caption{Ablation study on $L$, the \emph{balanced} queue size.}
    \label{tab:abl_queue_size}
\end{minipage}\qquad
\begin{minipage}{0.29\columnwidth}
    \centering
    \resizebox{.8\columnwidth}{!}{%
        \begin{tabular}{lcc}
            \toprule
             & C10 & STL10 \\
            \cmidrule{1-1} \cmidrule{2-3}
            FixMatch & 68.25 & 55.53 \\
            \cmidrule{1-1} \cmidrule{2-3}
            $T_{\text{proto}}=0.02$ & 73.84 & 68.19 \\
            \rowcolor{Gray}
            $T_{\text{proto}}=0.05$ & \textbf{75.97} & \textbf{70.21} \\
            $T_{\text{proto}}=0.2$  & 70.53 & 66.62 \\
            $T_{\text{proto}}=0.5$  & 52.36 & 60.92 \\
            $T_{\text{proto}}=1.0$  & 46.47 & 57.40 \\
            \bottomrule
        \end{tabular}%
    }
    \caption{Ablation study on $T_{\text{proto}}$ for semantic pseudo-label.}
    \label{tab:abl_temp_proto}
\end{minipage}\qquad
\begin{minipage}{0.29\columnwidth}
    \centering
    \resizebox{0.76\columnwidth}{!}{%
        \begin{tabular}{lcc}
            \toprule
             & C10 & STL10 \\
            \cmidrule{1-1} \cmidrule{2-3}
            FixMatch & 68.25 & 55.53 \\
            \cmidrule{1-1} \cmidrule{2-3}
            $\lambda_{\text{align}}=0$    & 70.98 & 61.64 \\
            $\lambda_{\text{align}}=0.5$  & 73.78 & 69.01 \\
            \rowcolor{Gray}
            $\lambda_{\text{align}}=1$    & \textbf{75.97} & {70.21} \\
            $\lambda_{\text{align}}=1.5$  & 74.59 & \textbf{71.51} \\
            $\lambda_{\text{align}}=2$    & 74.57 & 71.12 \\
            \bottomrule
        \end{tabular}%
    }
    \caption{Ablation study on $\lambda_{\text{align}}$, which is a weight for $\cL_{\text{align}}$.}
    \label{tab:abl_lamb_align}
\end{minipage}
\end{table}

%% file: supple/detailed_analysis.tex
\section{Detailed Analysis}
\label{sec:sup_analysis}
\subsection{Recall and Precision Analysis}
\label{sec:sup_rec_prec_analysis}
\subsubsection{Detailed comparison for linear pseudo-label and semantic pseudo-label methods}
We first take a closer look at the bias of pseudo-labels of each method by analyzing per-class recall and precision. 
We then compare the class-wise test accuracy of each model to evaluate the capability for each class, as done in the main paper. \cref{fig:analysis_c10_fud} provides the comparison of FixMatch w/ DASO (ours) and USADTM~\cite{han2020unsupervised} over FixMatch~\cite{sohn2020fixmatch} trained on CIFAR10-LT.
\input{supple/figures/analysis_c10_fud}

Compared to the linear pseudo-labels, the recall of semantic pseudo-labels on minority classes significantly increased in~\cref{fig:c10_recall_fud}. 
However, their precision values are degraded on the minorities, which means that the semantic pseudo-labels have the bias towards the minorities, leading to performance drop on the majority classes. 

In contrary, the pseudo-labels generated from our DASO maintain high precision while the recall on the minority classes increased, encouraging high performance on both of majority and minority classes. 
From the analyses, pseudo-labels from DASO find the trade-off between linear and semantic pseudo-labels with respect to the bias that performs well on test data. 
Since DASO also aims to keep the prediction of majority classes, the test accuracy drop on the head classes is well addressed. 

Note that \cref{fig:analysis_c100_fud} shows the same analysis on the models trained on CIFAR100-LT. 
\input{supple/figures/analysis_c100_fud}

\subsubsection{DASO with class distribution mismatch on traditional SSL learner}
We present the analyses of bias in pseudo-labels for the other \emph{classic} SSL algorithms: MeanTeacher~\cite{tarvainen2017mean} and MixMatch~\cite{berthelot2019mixmatch} in \cref{fig:analysis_c10_mt,fig:analysis_c10_mm}, respectively, in case of uniform distribution of unlabeled data; \ie, $\gamma_u=1$. 
In such a case, class distribution mismatch (\ie, $\gamma_l \neq \gamma_u$) can damage the accuracy of the model.

From the recall curves in~\cref{fig:mt_recall,fig:mm_recall} and the precision curves in~\cref{fig:mt_prec,fig:mm_prec}, the pseudo-labels of the baseline SSL learners are {severely} biased towards the head classes, since most of the minority class examples are collapsed to the majority class ones. 
The unlabeled data with $\gamma_u=1$ rather significantly accelerated the bias, to the point where the precision curve is completely reversed; precision values in the majority classes significantly degraded, compared to the recall curve. 
Thereby, the model rarely predicts some of the minority class examples for the test dataset in~\cref{fig:mt_acc,fig:mm_acc}. 
\input{supple/figures/analysis_c10_mt}
\input{supple/figures/analysis_c10_mm}

In contrast, we demonstrate that DASO can even \emph{completely} mitigate such a devastating bias, by just coupling the linear pseudo-labels with the semantic pseudo-labels obtained from the similarity-based classifier. 
In this case, the semantic alignment loss $\cL_{\text{align}}$ is not applied, due to the absence of advanced augmentation $\cA_{s}$ for MeanTeacher and MixMatch. 
Surprisingly, in MeanTeacher (MT) with DASO, the recall and precision values become uniform, resulting in a \emph{uniform} per-class test accuracy in~\cref{fig:mt_acc}. 
When combined with MixMatch~\cite{berthelot2019mixmatch}, DASO also recovers the minority-class pseudo-labels significantly. 
In final, the averaged test accuracy can be more than doubled (\ie, $37.3\% \to 77.2\%$), as shown in~\cref{fig:mm_acc}. 

As such, DASO helps alleviate the bias in pseudo-labels, even when the class distributions between labeled and unlabeled data substantially differ, without accessing the knowledge about the underlying distribution of unlabeled data. 

\begin{figure}[ht]
  \centering
   \includegraphics[width=0.8\linewidth]{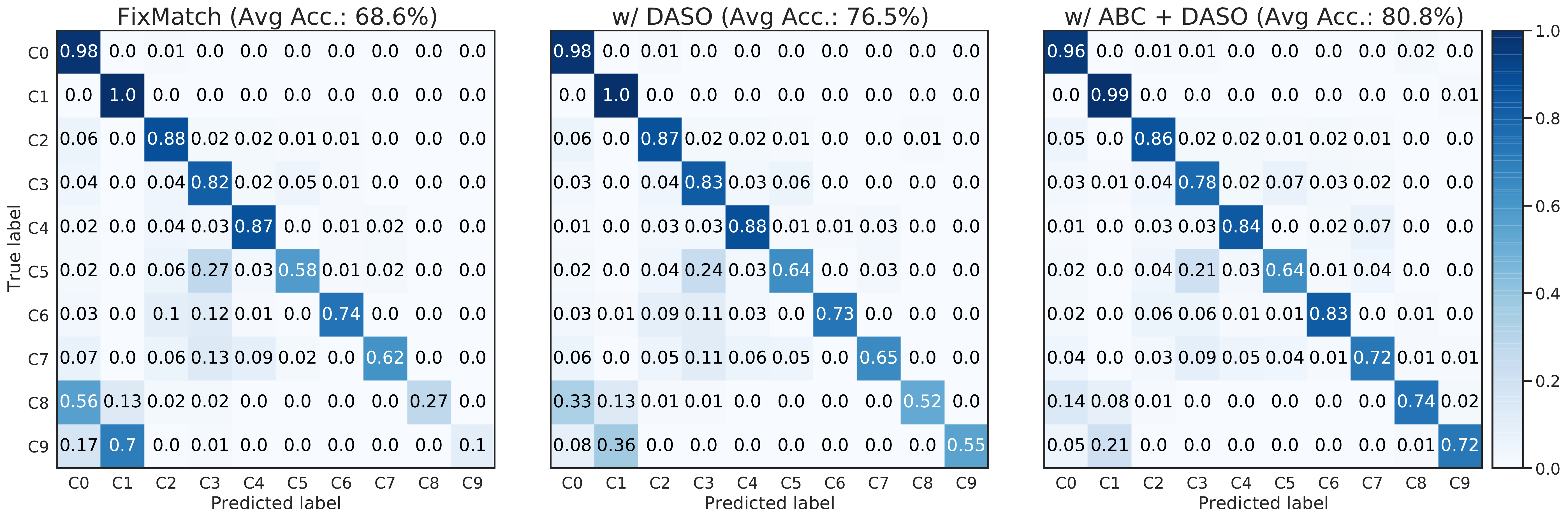}
   \caption{Analysis of predictions from test data via confusion matrix. All the methods are trained on CIFAR10-LT with $\gamma=100$ and $N_1=500$ upon the same fixed random seed. 
   DASO greatly recovers the predictions on the actual minority class examples in test data. 
   }
   \label{fig:cm_c10}
\end{figure}
\subsection{Confusion Matrix on Test Data}
We compare the confusion matrices of the predictions from the test data. From the baseline FixMatch~\cite{sohn2020fixmatch}, we further apply our DASO on both FixMatch and FixMatch w/ ABC~\cite{lee2021auxiliary}. 
As shown in~\cref{fig:cm_c10}, the predictions on the tail classes (\eg, C8 and C9) in FixMatch are severely biased towards the majority classes (\eg, C1). This limits the overall performance, which is carried by the non-minority classes (68.6\%). 
On the other hand, from the center of~\cref{fig:cm_c10}, DASO significantly alleviates the bias towards the head classes observing C8 and C9 classes, while the performances on the other classes are well maintained. 
When DASO is integrated with ABC~\cite{lee2021auxiliary} in the right figure, the accuracy values are further improved. 

\subsection{Train Curves for Recall and Accuracy}
We compare the train curves of recall and test accuracy values from FixMatch~\cite{sohn2020fixmatch} and FixMatch w/ DASO (Ours) trained on CIFAR10/100-LT respectively in~\cref{fig:curve_c10,fig:curve_c100}. 
Here, we plot those from majority classes (\eg, first 20\% classes) and minority classes (\eg, last 20\% classes), in addition to the overall values. 
From both CIFAR10/100-LT benchmarks, DASO significantly improves the recall and test accuracy values on the minority classes, while relatively maintaining those from the majority classes. 
This verifies the efficacy of DASO that specifically handles the biased minority classes in unlabeled data.
\input{supple/figures/train_curves}

\subsection{Further Comparison of Feature Representations}
\begin{figure}[ht]
  \centering
   \includegraphics[width=0.7\linewidth]{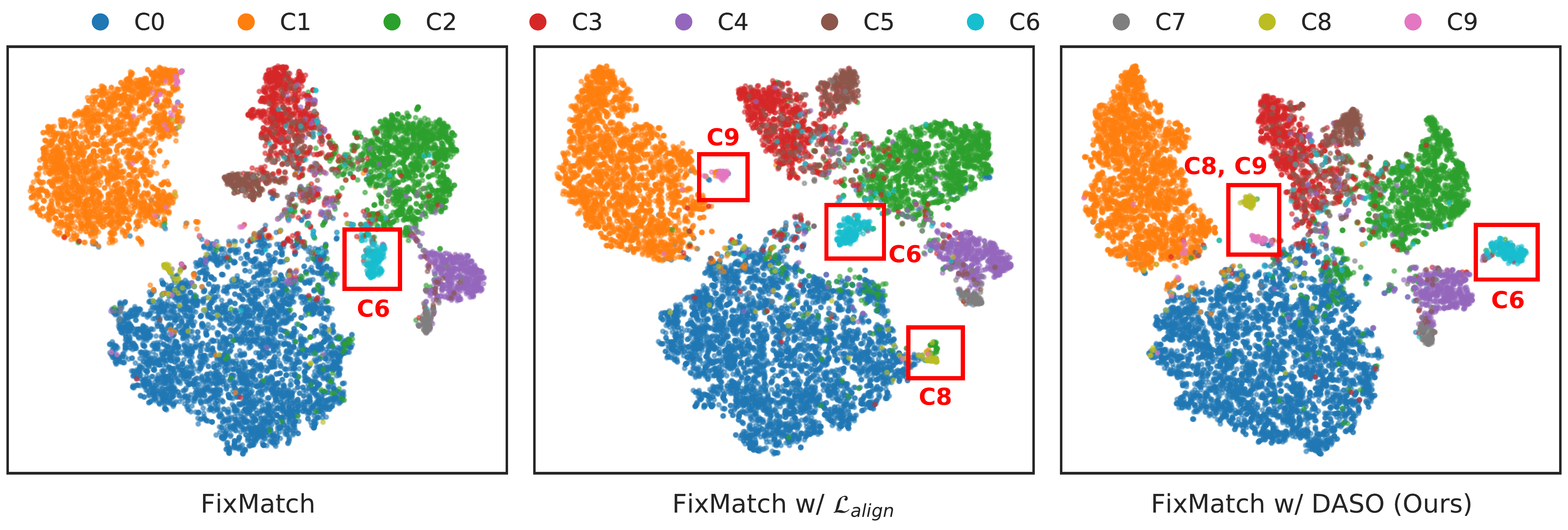}
   \caption{Comparison of t-SNE~\cite{van2008visualizing} visualizations of feature representations. 
   We additionally compare the model trained with FixMatch w/ $\cL_{\text{align}}$ between the original FixMatch~\cite{sohn2020fixmatch} and FixMatch w/ DASO (Ours). 
   Note that both of the semantic alignment loss $\cL_{\text{align}}$ and our class-adaptive pseudo-label blending contribute to alleviating the bias in pseudo-labels in perspective of feature representation. 
   }
   \label{fig:sa_tsne}
\end{figure}

To verify the efficacy of the proposed semantic alignment loss ($\cL_{\text{align}}$), we further visualize the t-SNE~\cite{van2008visualizing} of the feature encoder outputs from FixMatch w/ $\cL_{\text{align}}$ in the center of~\cref{fig:sa_tsne}. 
Compared to FixMatch, applying $\cL_{\text{align}}$ without the class-adaptive pseudo-label blending can already cluster the minority classes (\eg, C6, C8, and C9) in the center of the figure. 
However, those indicated clusters lie nearby the head-class clusters (\eg, C0 and C1), where the classifier can still be confused. 
In that sense, the complete DASO from the right figure further improves the separability of the tail classes from the head classes. 
This demonstrates that while applying the semantic alignment loss $\cL_{\text{align}}$ could be helpful for the minority classes, both class-adaptive pseudo-label blending and $\cL_{\text{align}}$ are the essential components for our DASO framework.

\subsection{Confidence Analysis from Out-of-class Examples}
\label{sec:sup_saves_conf}
To investigate the efficacy of DASO pseudo-label, we analyze the confidence of predictions of unlabeled data after training model with $\cU=\cU_{\text{in}}+\cU_{\text{out}}$ under Semi-Aves benchmark~\cite{su2021semisupervised}. 
\cref{fig:analysis_saves_ent} visualizes the histograms of entropy values obtained from either FixMatch~\cite{sohn2020fixmatch} or FixMatch w/ DASO, respectively. 
Note that since both models do not explicitly learn how to distinguish \emph{in-class} and \emph{out-of-class} categories at all, those samples cannot be completely separated in confidence plot. 

\input{supple/figures/analysis_s_aves_ent}

FixMatch w/ DASO, which learned the blending of linear and semantic pseudo-labels can be effective in that the \emph{out-of-class} examples in $\cU_{\text{out}}$ are further pushed towards the low-confidence region (\ie, higher entropy) compared to the \emph{in-class} unlabeled examples in $\cU_{\text{in}}$. 
For example, about $8k$ out-of-class examples correspond to the most confident samples in~\cref{fig:saves_ent_fm}, 
while they reduced to $4k$ with DASO in~\cref{fig:saves_ent_daso}.
We suppose 
DASO has the \emph{implicit} ability to push more examples corresponding to \emph{out-of-class} that can cause degradation, towards the low-confident area. 
This point implies the potential application of DASO towards an open-set SSL scenario, where SSL algorithms also observe unlabeled data in a broader class distribution compared to the labels, and learning without \emph{harmful} out-of-class examples would be important.

%% file: supple/figures/analysis_c10_fud.tex
\begin{figure}[ht]
\centering
\subfloat[][Recall of pseudo-labels]{
    \includegraphics[width=.29\linewidth]{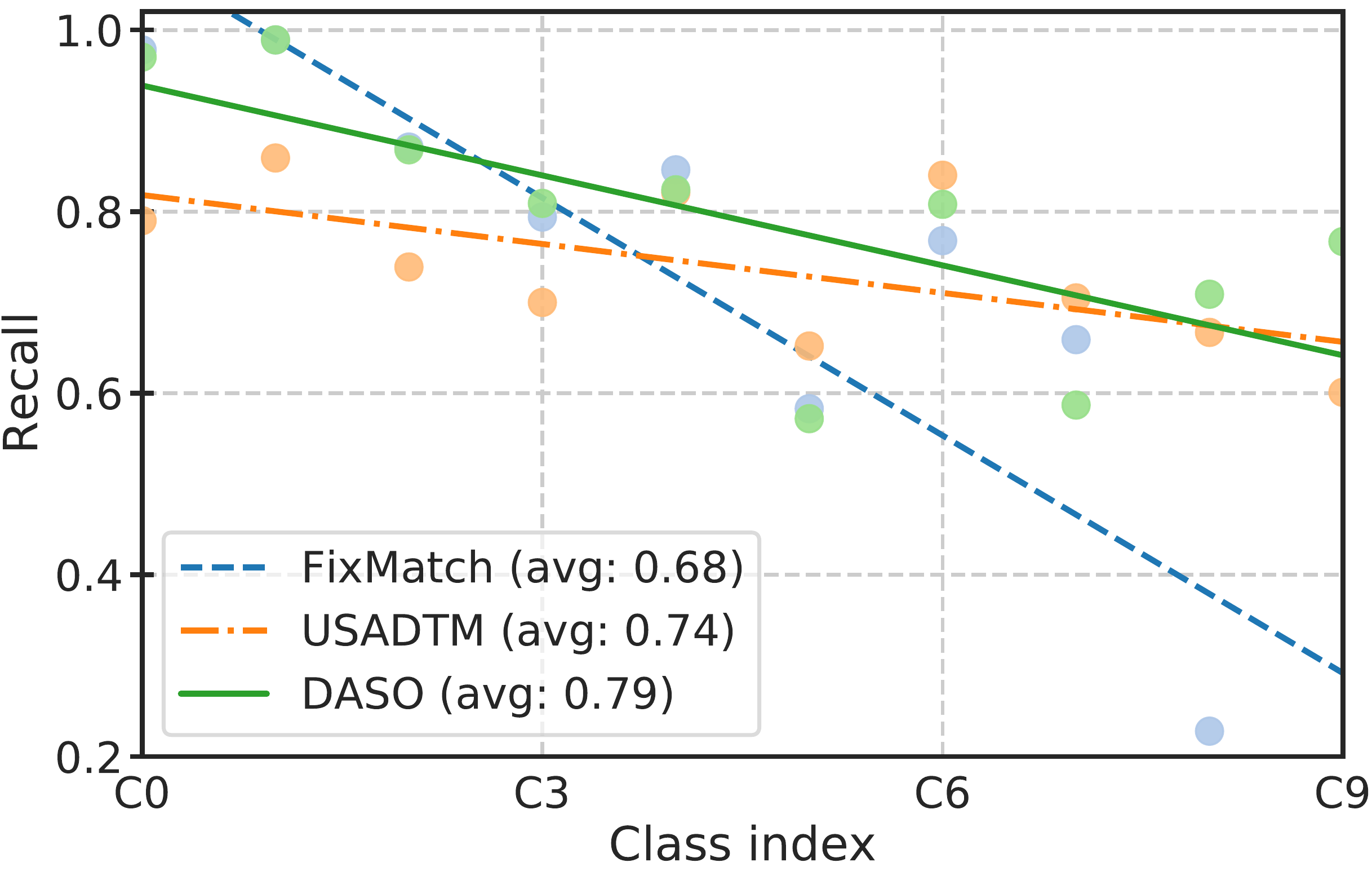}
    \label{fig:c10_recall_fud}
}
\quad
\subfloat[][Precision of pseudo-labels]{
    \includegraphics[width=.29\linewidth]{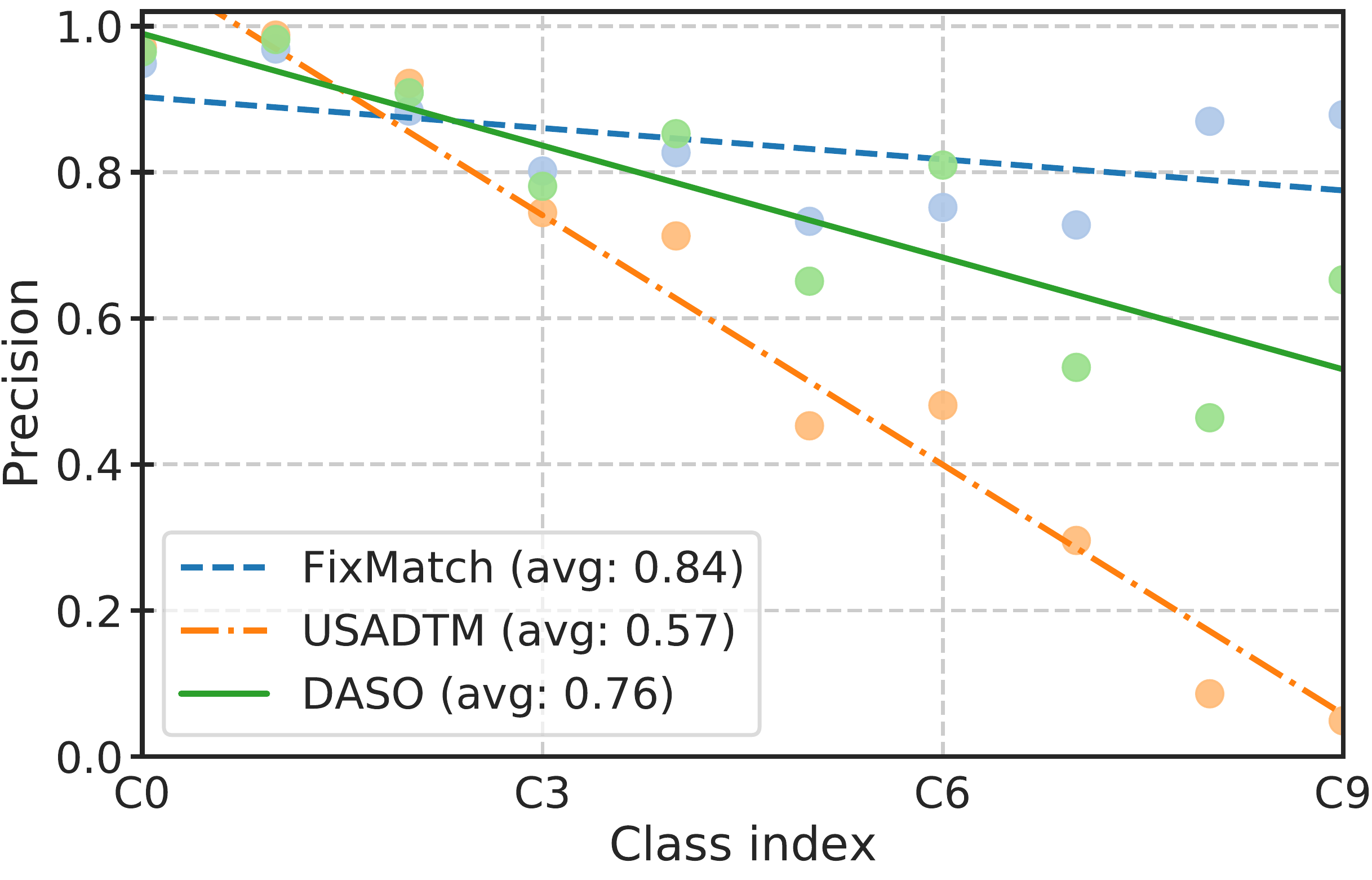}
    \label{fig:c10_prec_fud}
}
\quad
\subfloat[][Class-wise test accuracy]{
    \includegraphics[width=.29\linewidth]{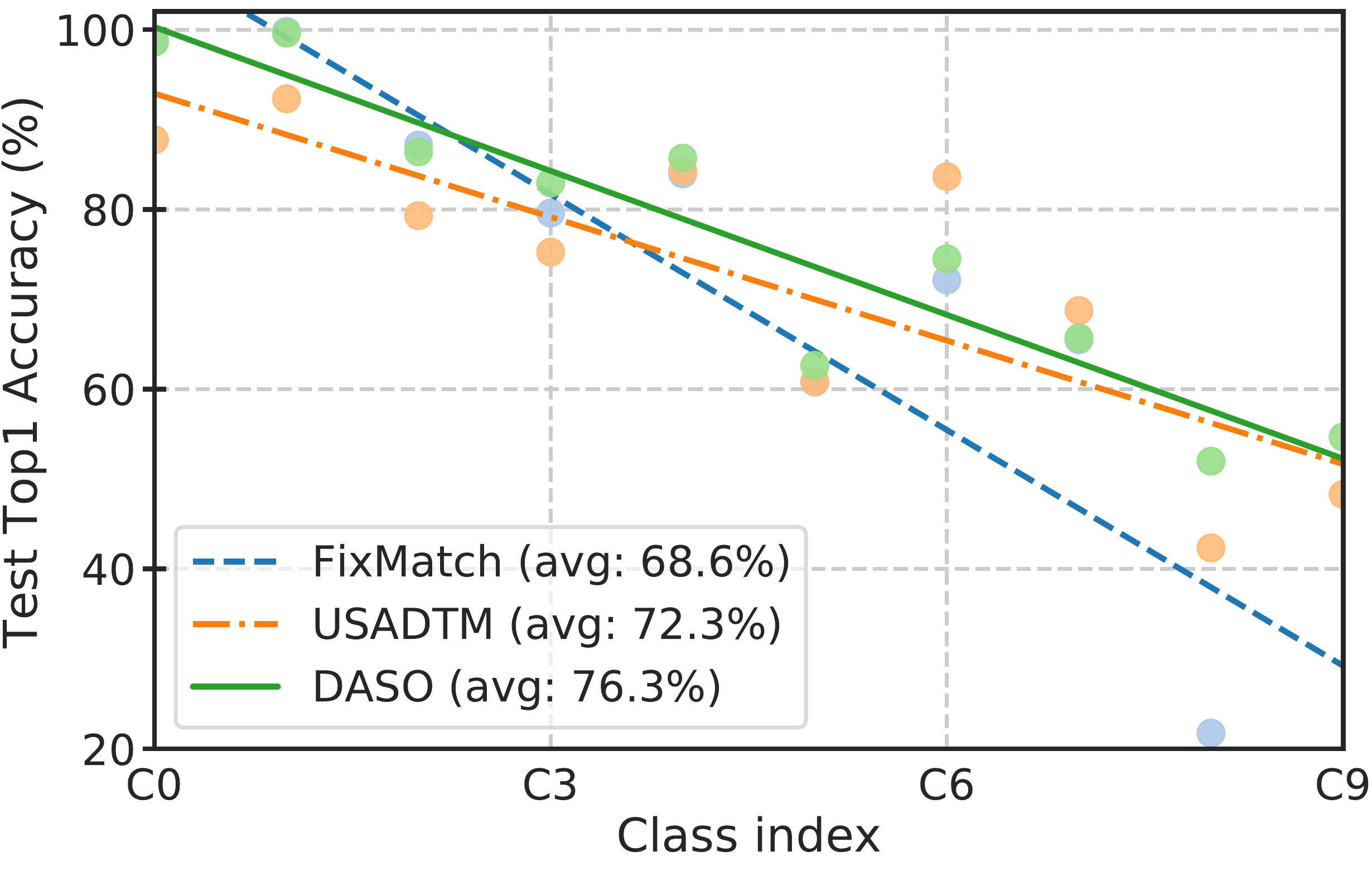}
    \label{fig:c10_acc_fud}
}
\caption{Analysis of bias in pseudo-labels and test accuracy. We consider FixMatch~\cite{sohn2020fixmatch} for linear pseudo-labels, USADTM~\cite{han2020unsupervised} for semantic pseudo-labels, and the proposed FixMatch w/ DASO trained on CIFAR10-LT with $N_1=500$ with $\gamma_l=\gamma_u=100$.}
\label{fig:analysis_c10_fud}
\end{figure}

%% file: supple/figures/analysis_c100_fud.tex
\begin{figure}[ht]
\centering
\subfloat[][Recall of pseudo-labels]{
    \includegraphics[width=.29\linewidth]{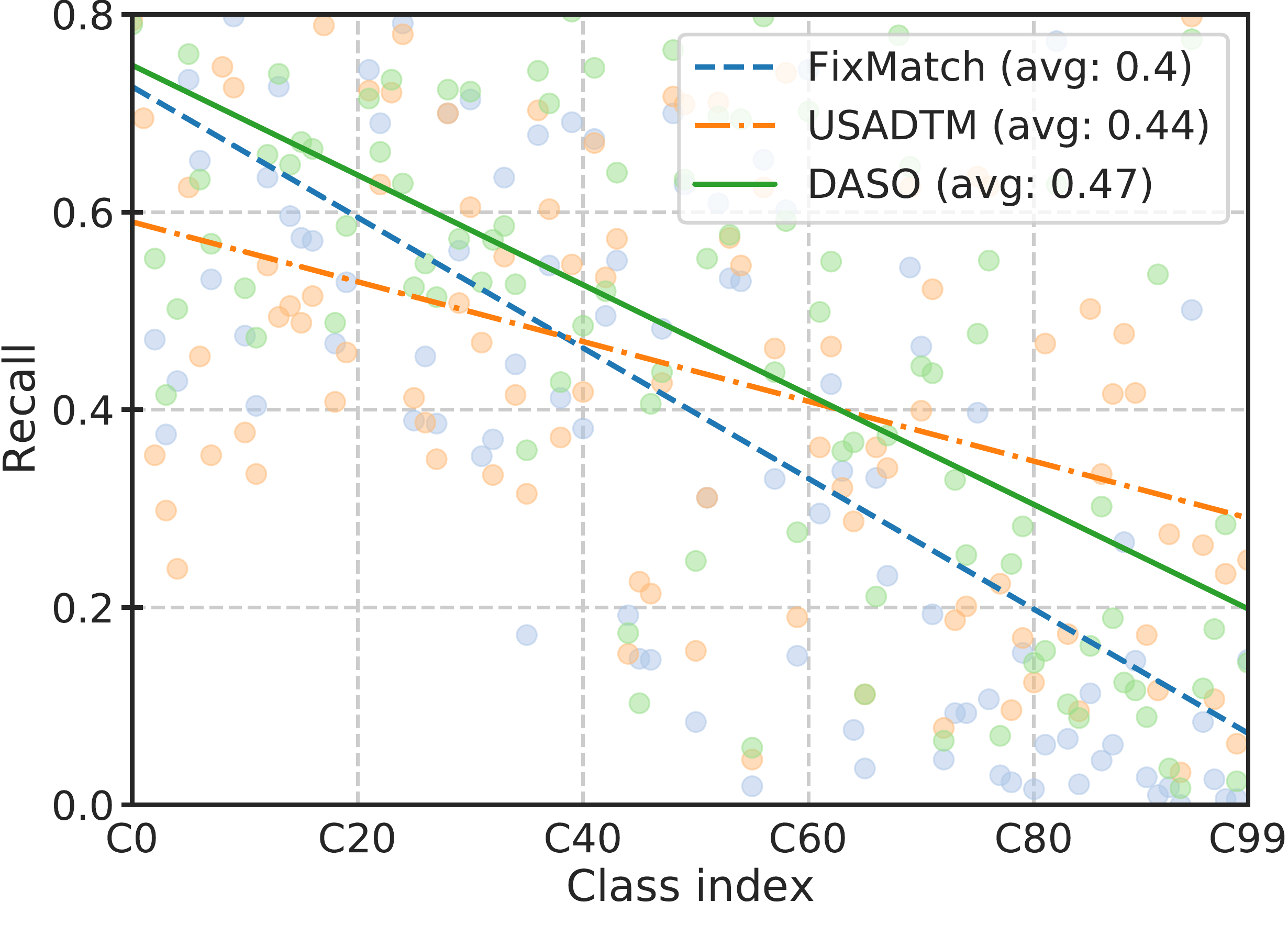}
    \label{fig:c100_recall_fud}
}
\quad
\subfloat[][Precision of pseudo-labels]{
    \includegraphics[width=.29\linewidth]{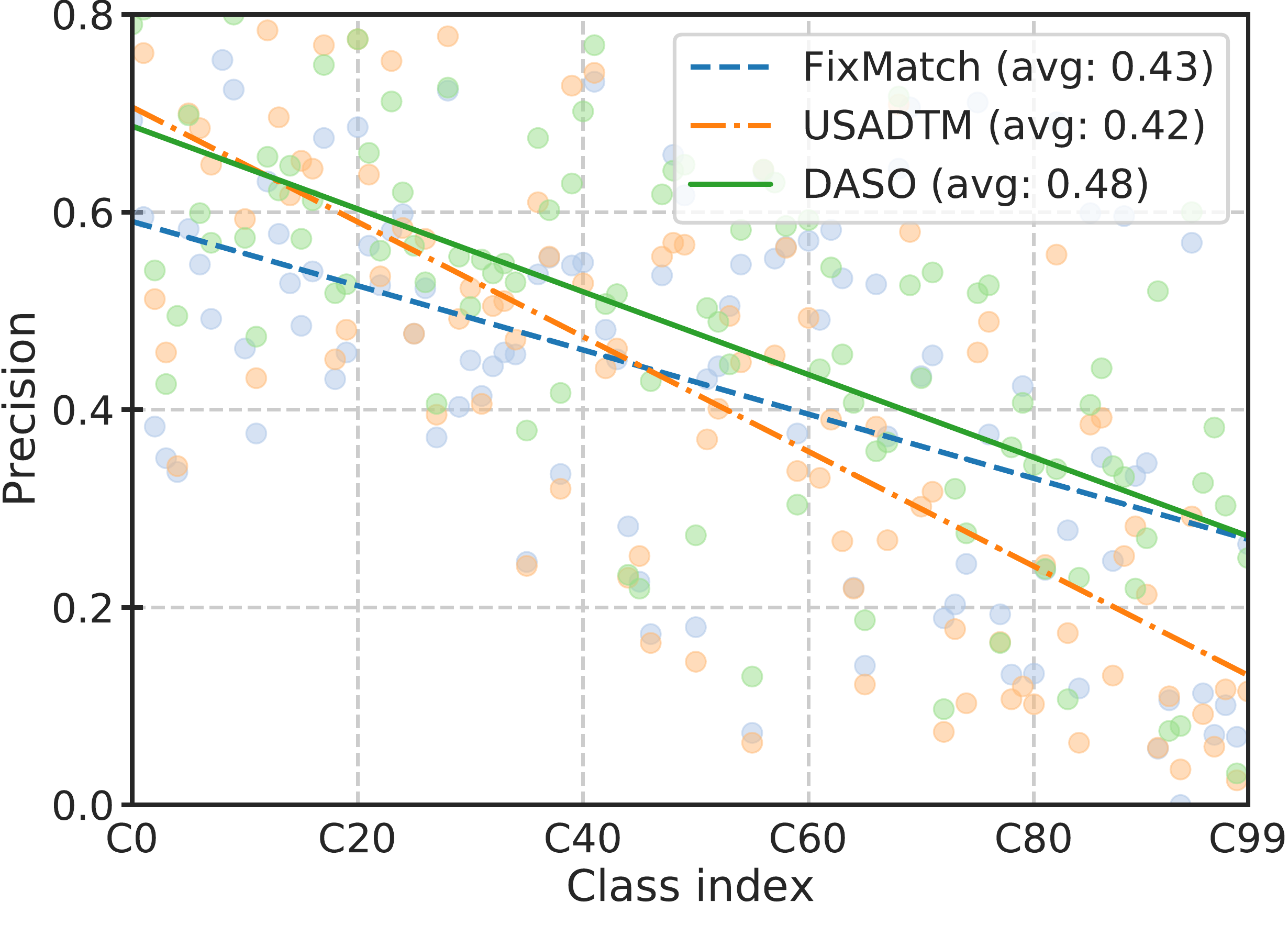}
    \label{fig:c100_prec_fud}
}
\quad
\subfloat[][Class-wise test accuracy]{
    \includegraphics[width=.29\linewidth]{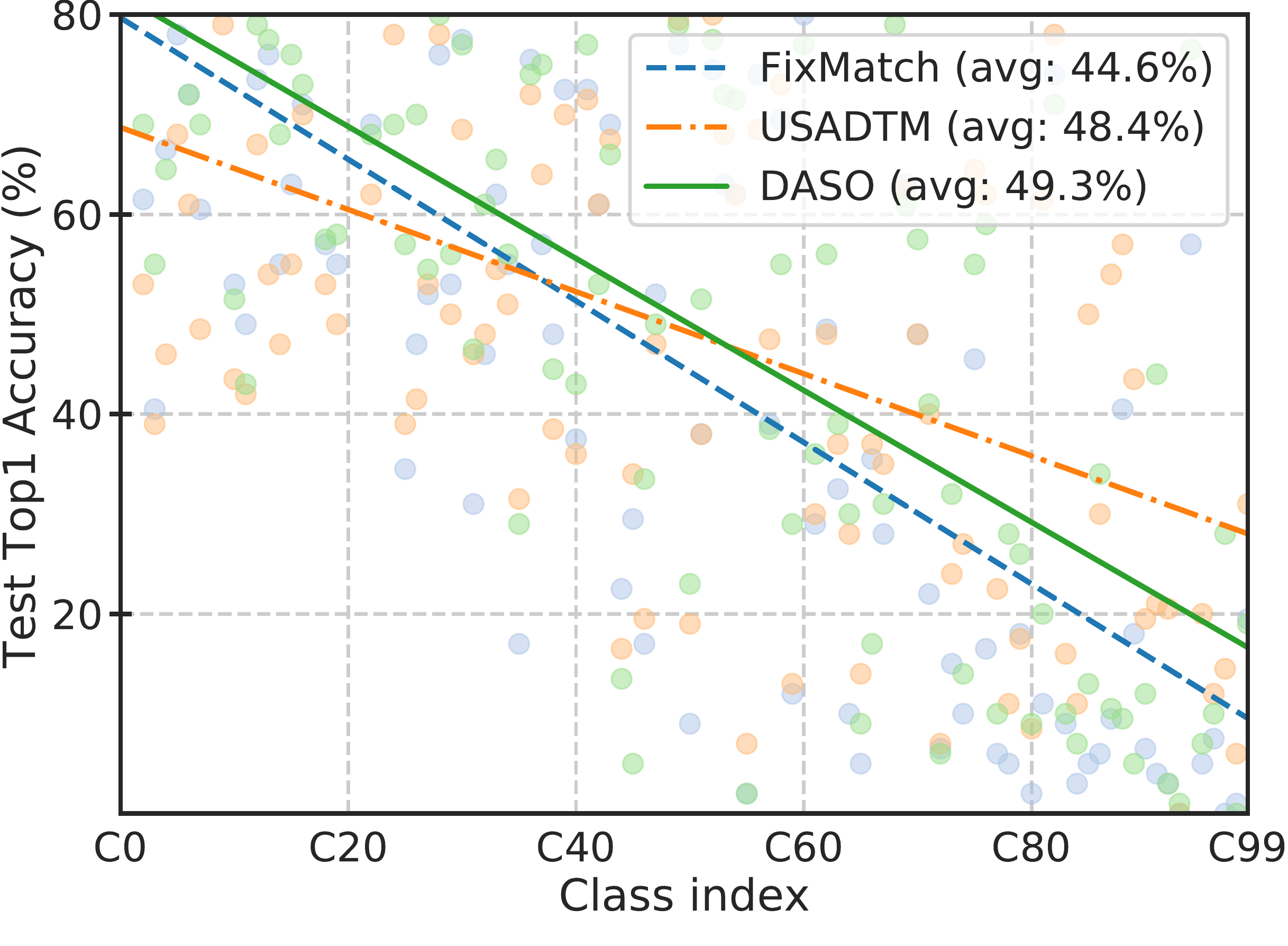}
    \label{fig:c100_acc_fud}
}
\caption{Analysis of bias in pseudo-labels. We consider FixMatch~\cite{sohn2020fixmatch} for linear pseudo-labels, USADTM~\cite{han2020unsupervised} for semantic pseudo-labels, and the proposed FixMatch w/ DASO trained on CIFAR100-LT with $N_1=50$ with $\gamma_l=\gamma_u=10$.}
\label{fig:analysis_c100_fud}
\vspace{-2mm}
\end{figure}

%% file: supple/figures/analysis_c10_mt.tex
\begin{figure}[ht]
\centering
\subfloat[][Recall of pseudo-labels]{
    \includegraphics[width=.30\linewidth]{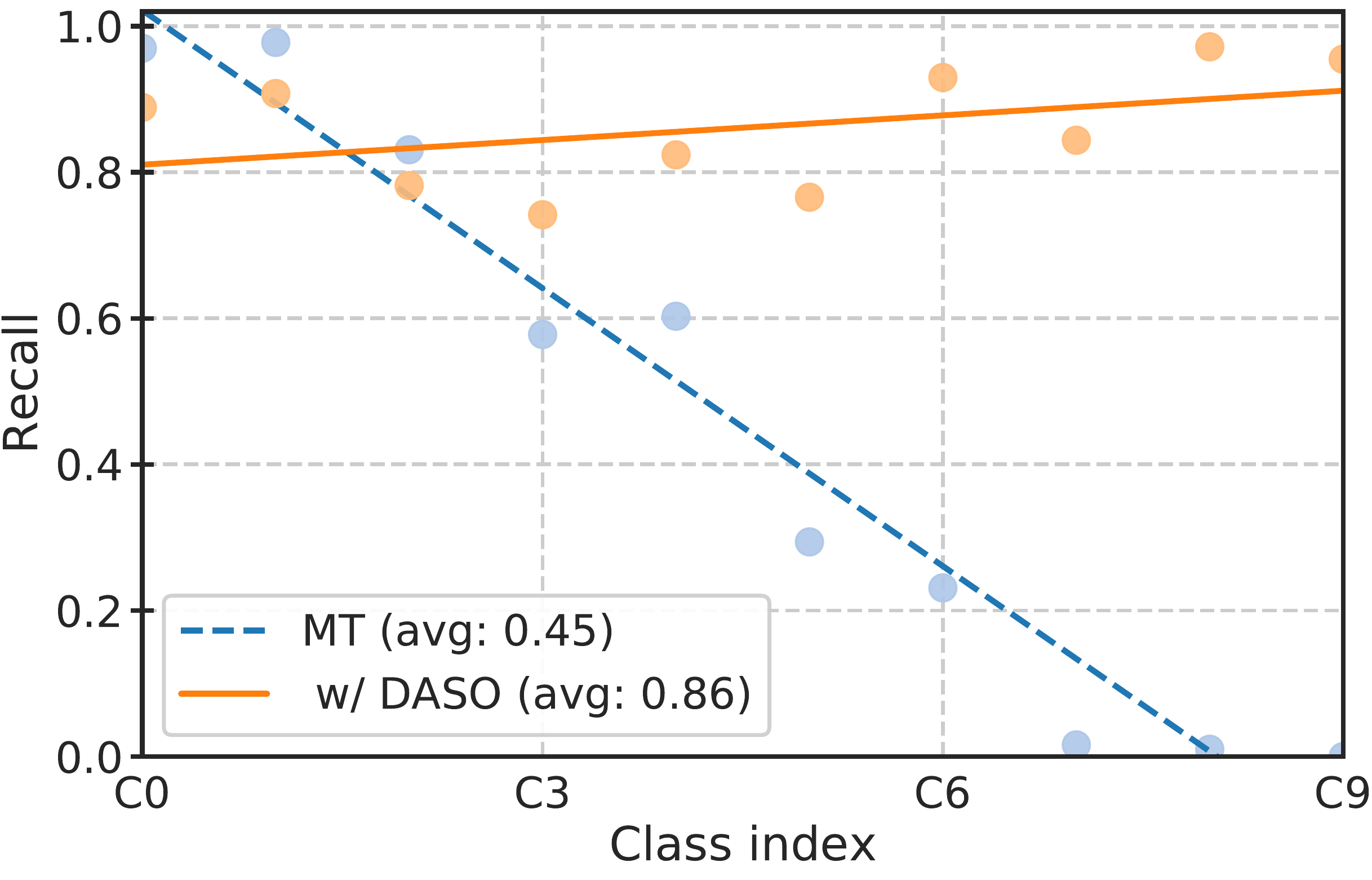}
    \label{fig:mt_recall}
}
\quad
\subfloat[][Precision of pseudo-labels]{
    \includegraphics[width=.30\linewidth]{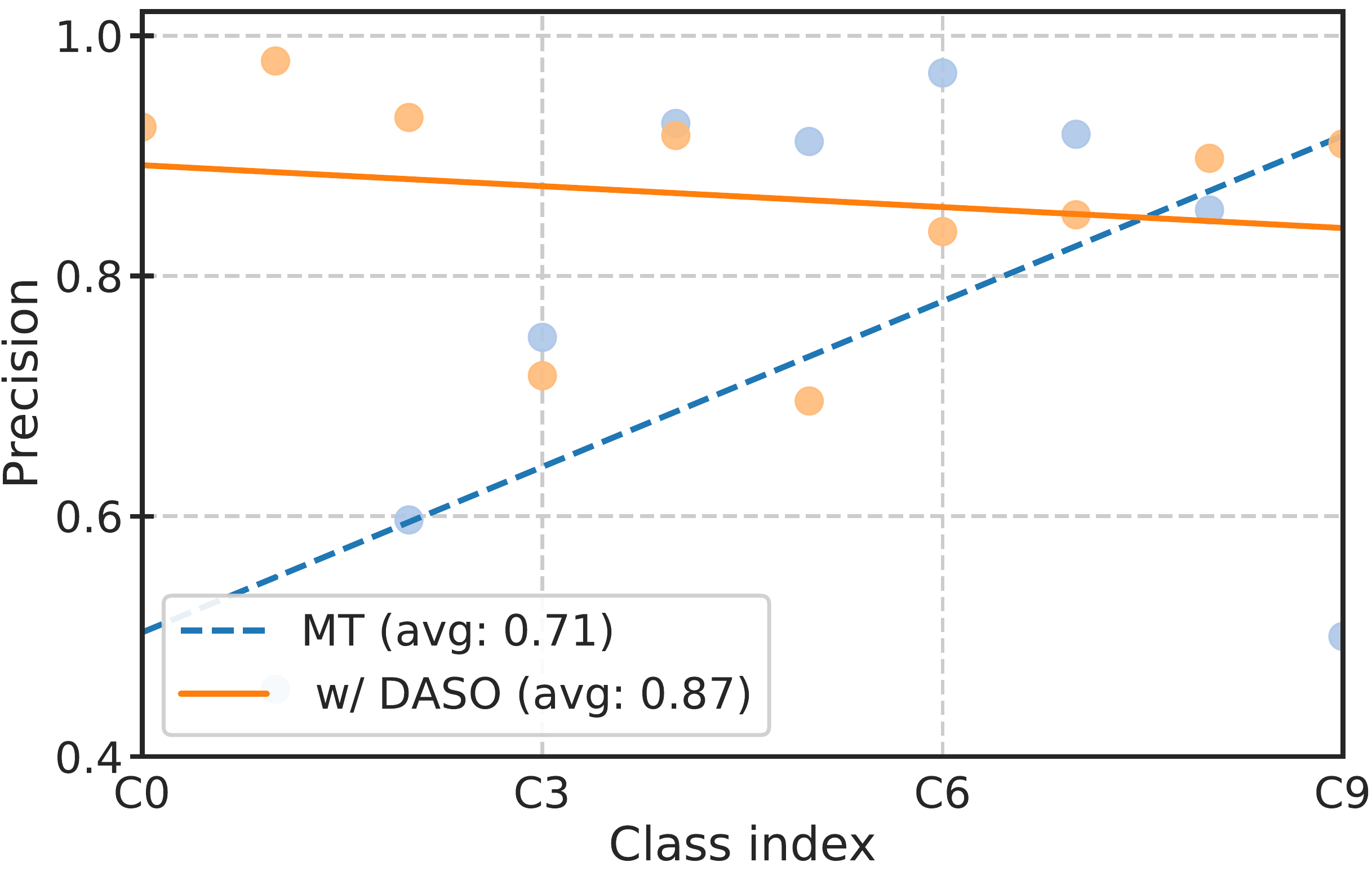}
    \label{fig:mt_prec}
}
\quad
\subfloat[][Class-wise test accuracy]{
    \includegraphics[width=.30\linewidth]{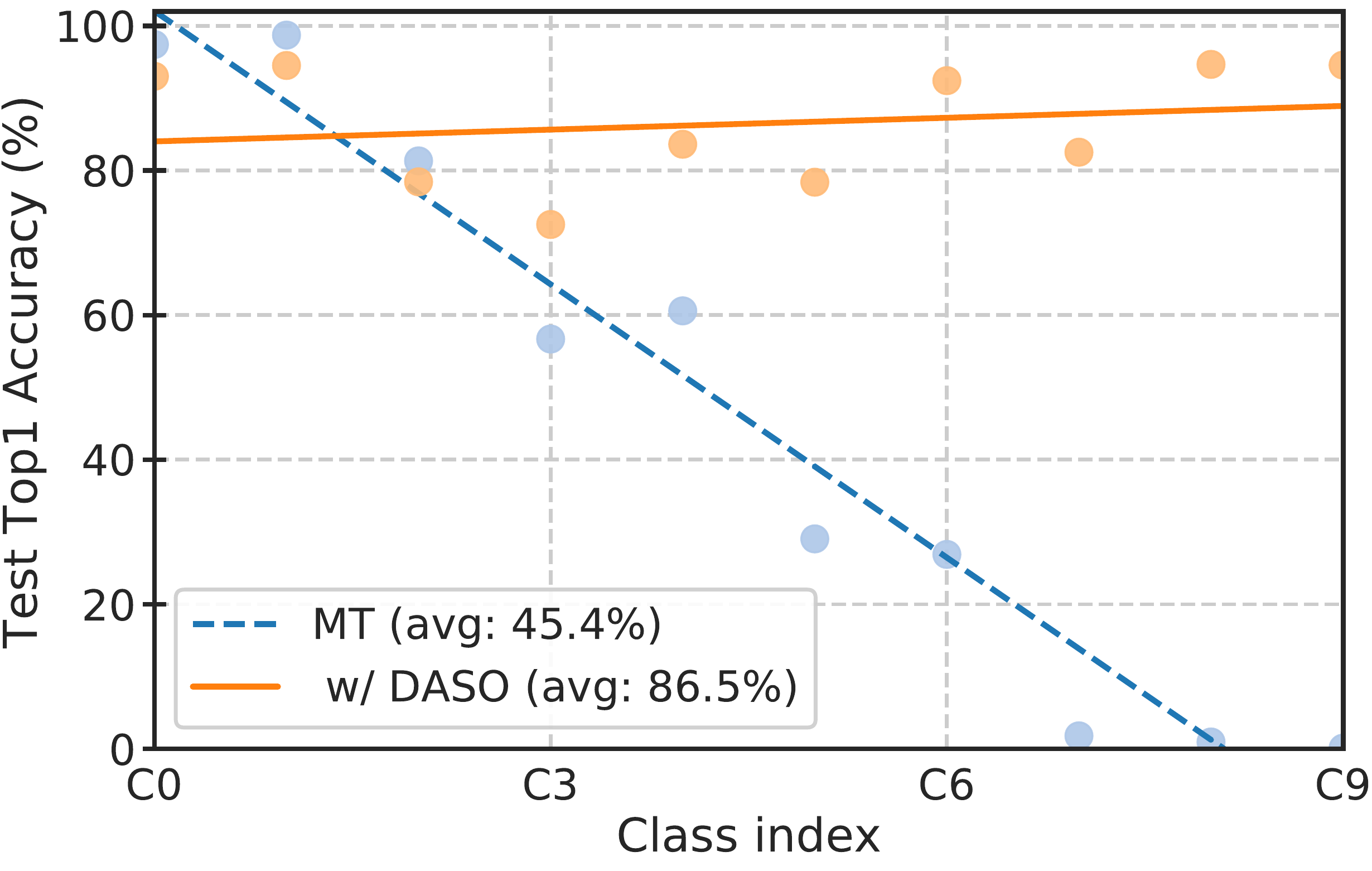}
    \label{fig:mt_acc}
}
\caption{Analysis of bias in pseudo-labels and test accuracy. We consider MeanTeacher (MT)~\cite{tarvainen2017mean}, and the proposed DASO applied to MT (MT w/ DASO) trained on CIFAR10-LT with $N_1=1500$ with $\gamma_l=100$ and $\gamma_u=1$.}
\label{fig:analysis_c10_mt}
\end{figure}

%% file: supple/figures/analysis_c10_mm.tex
\begin{figure}[ht]
\centering
\subfloat[][Recall of pseudo-labels]{
    \includegraphics[width=.30\linewidth]{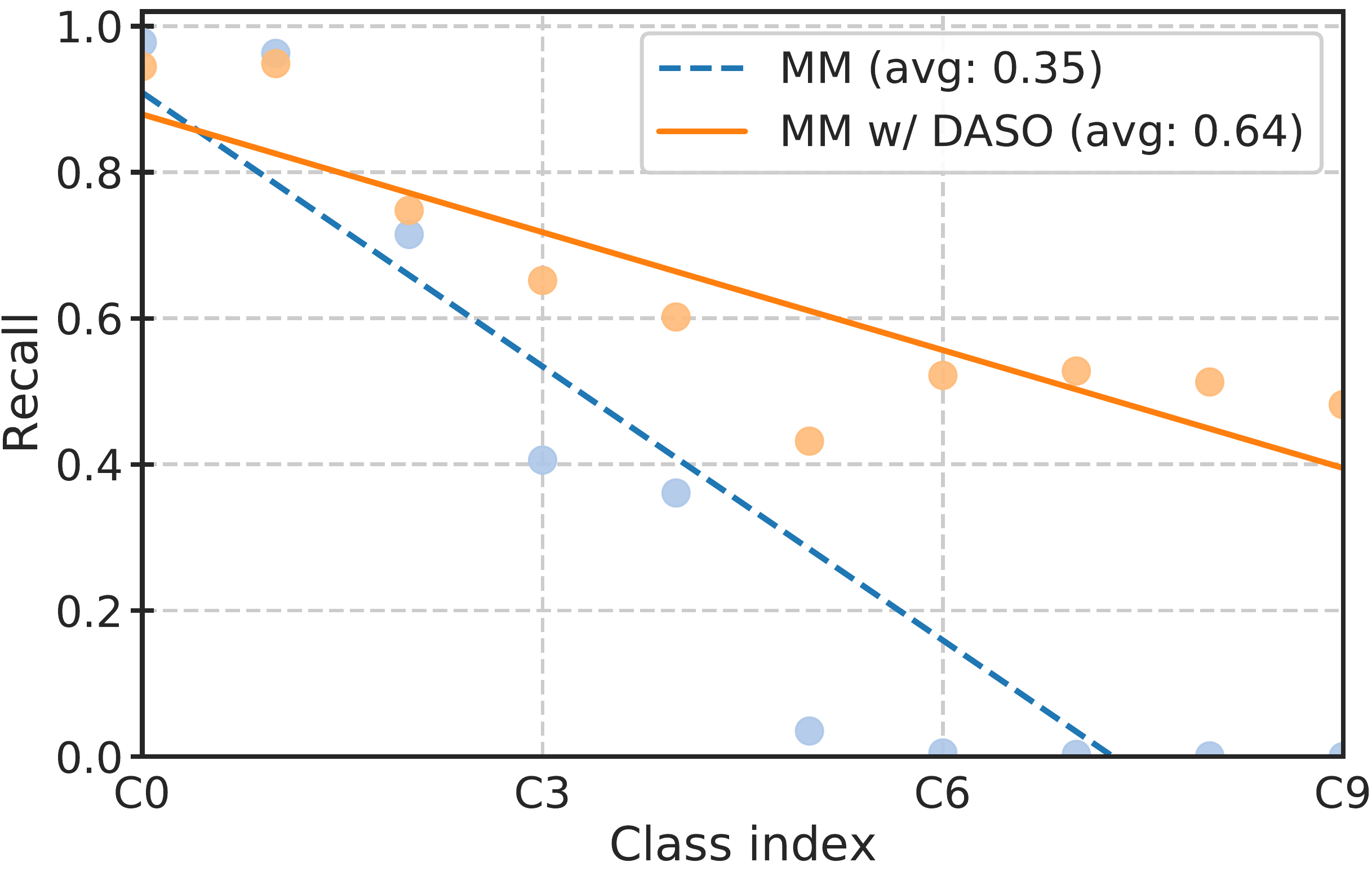}
    \label{fig:mm_recall}
}~
\subfloat[][Precision of pseudo-labels]{
    \includegraphics[width=.30\linewidth]{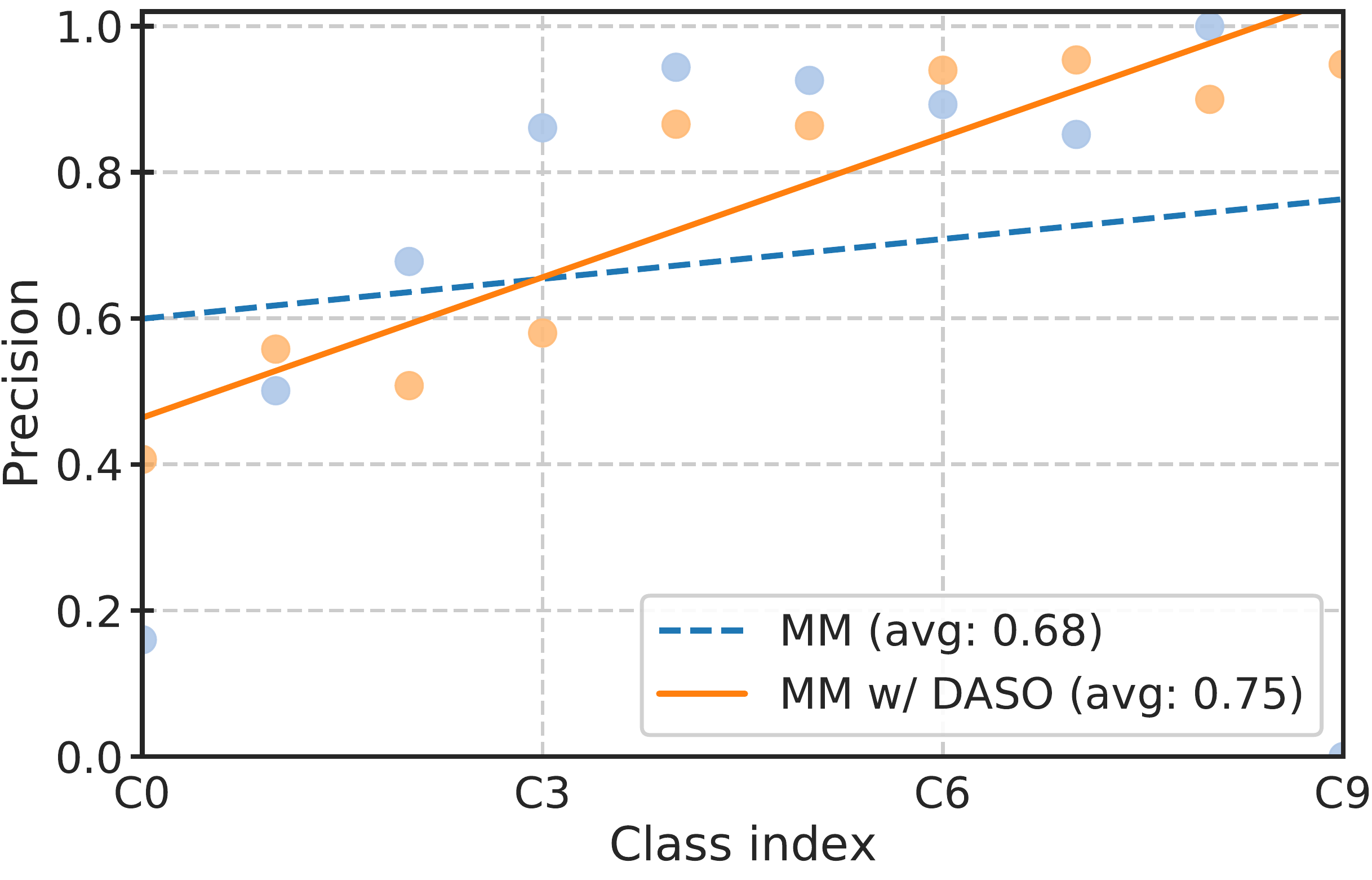}
    \label{fig:mm_prec}
}~
\subfloat[][Class-wise test accuracy]{
    \includegraphics[width=.30\linewidth]{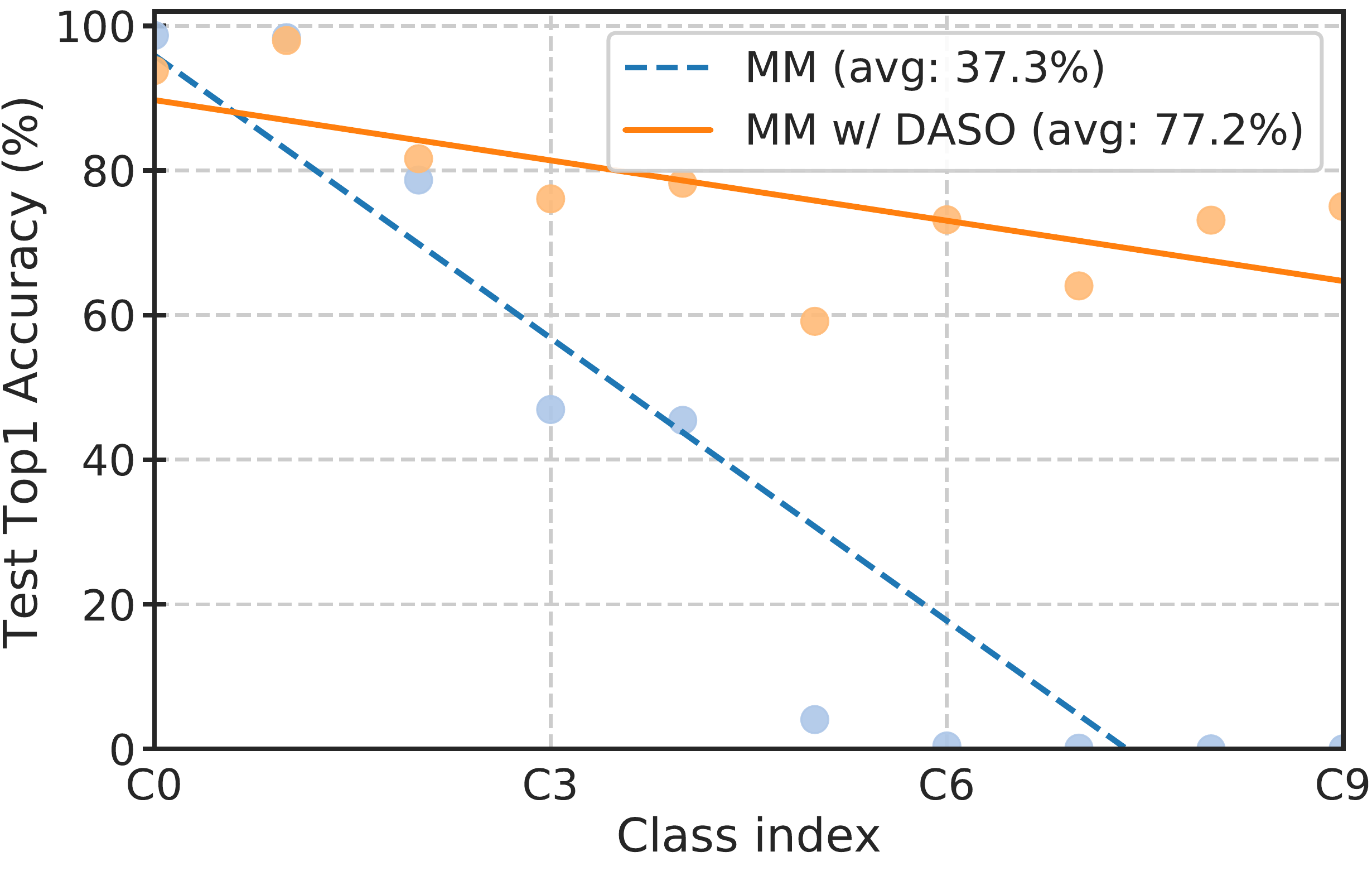}
    \label{fig:mm_acc}
}
\caption{Analysis of bias in pseudo-labels and test accuracy. We consider MixMatch (MM)~\cite{tarvainen2017mean}, and the proposed DASO applied to MM (MM w/ DASO) trained on CIFAR10-LT with $N_1=1500$ with $\gamma_l=100$ and $\gamma_u=1$.}
\label{fig:analysis_c10_mm}
\vspace{-3mm}
\end{figure}

%% file: supple/figures/train_curves.tex
\begin{figure}[ht]
\centering
\subfloat[][Train curves on CIFAR10-LT]{
    \includegraphics[width=.47\linewidth]{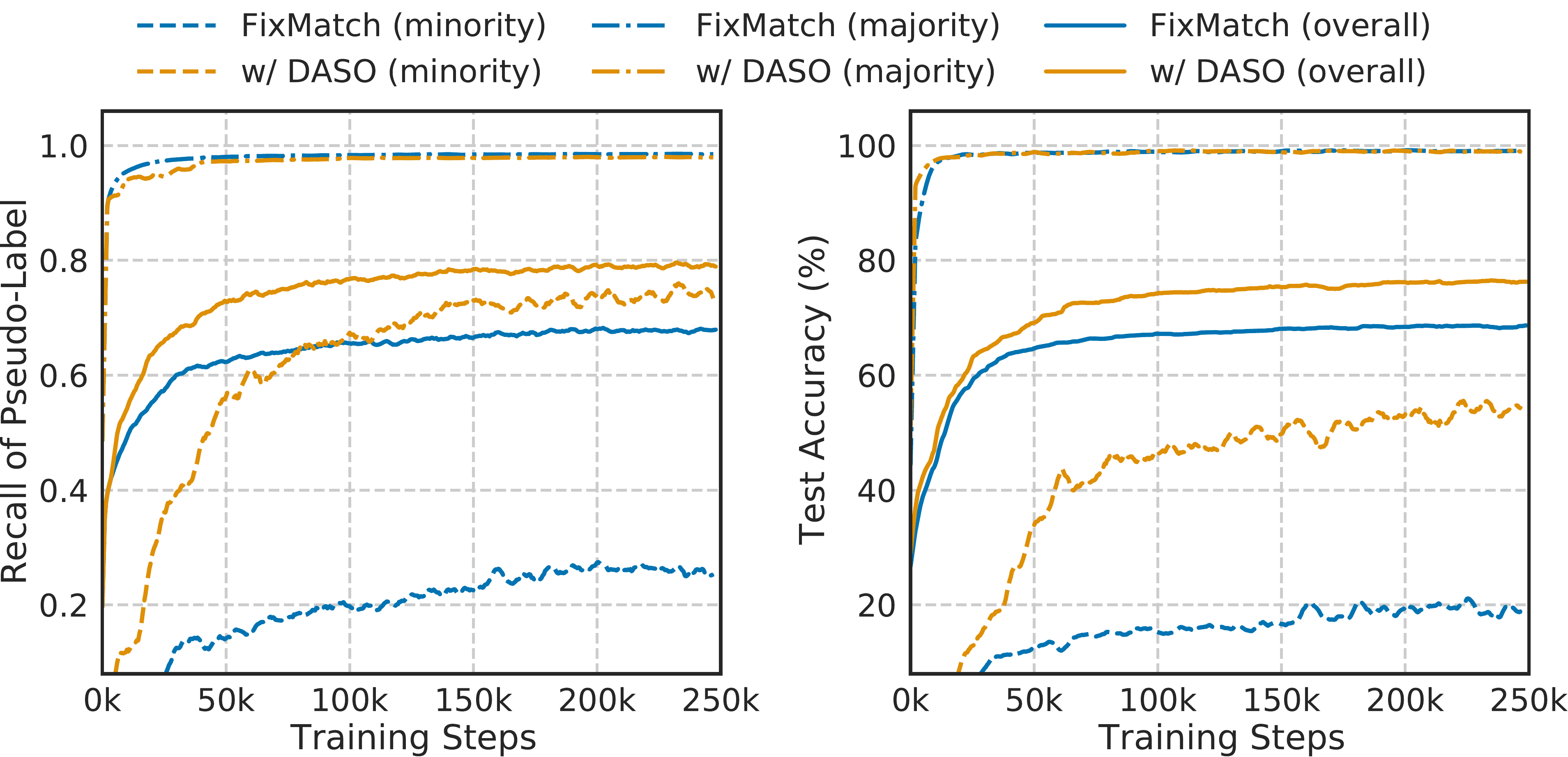}
    \label{fig:curve_c10}
}
\quad
\subfloat[][Train curves on CIFAR100-LT]{
    \includegraphics[width=.47\linewidth]{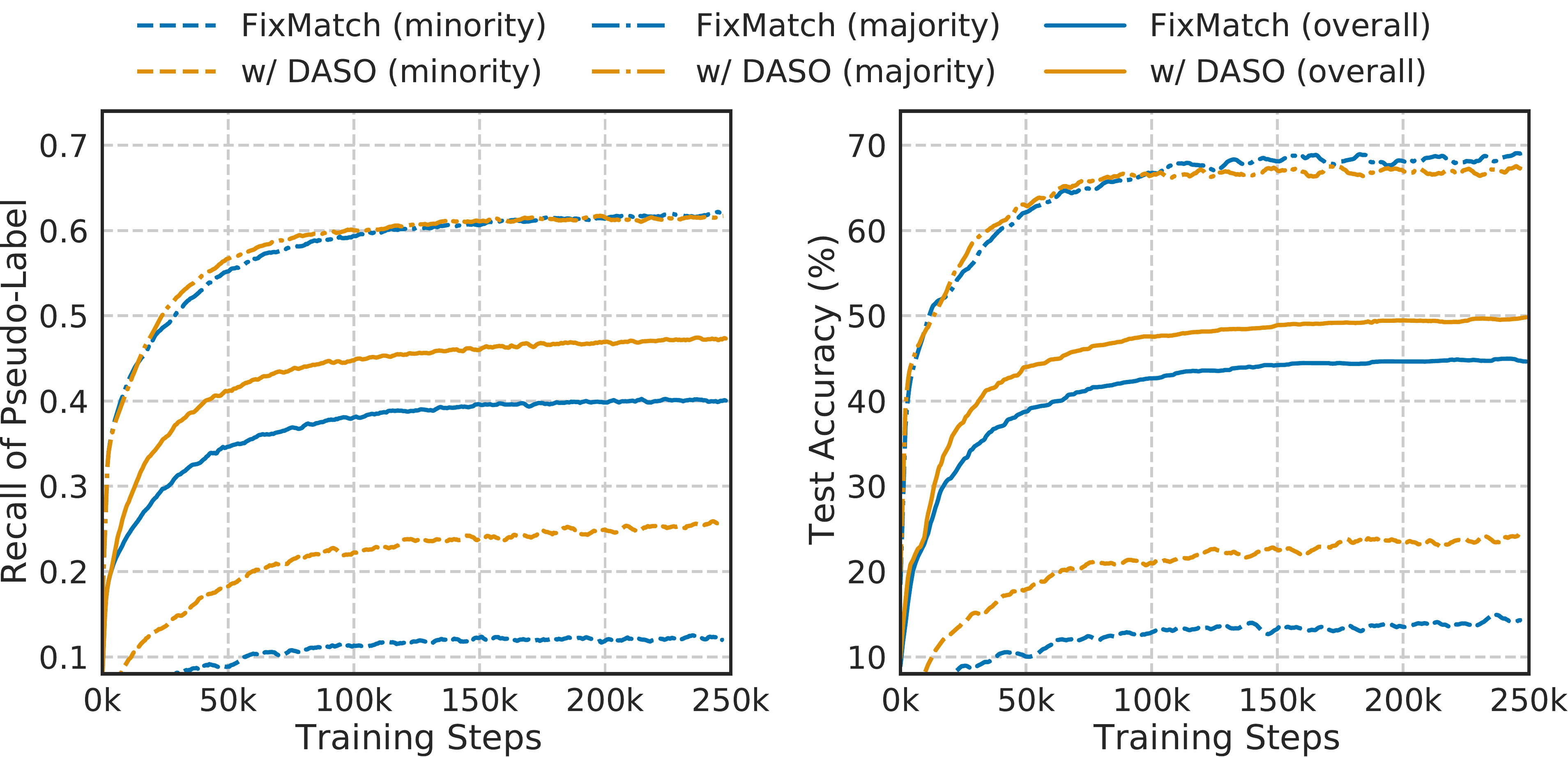}
    \label{fig:curve_c100}
}
\caption{Train curves for the recall and test accuracy values obtained from FixMatch and FixMatch w/ DASO (Ours). The training details are consistent from the main paper. DASO well reduces the biases on the tail classes, while preserving those from the head classes.}
\label{fig:train_curves}
\end{figure}

%% file: supple/figures/analysis_s_aves_ent.tex
\begin{figure}[ht]
\centering
\subfloat[][FixMatch~\cite{sohn2020fixmatch}]{
    \includegraphics[width=.32\linewidth]{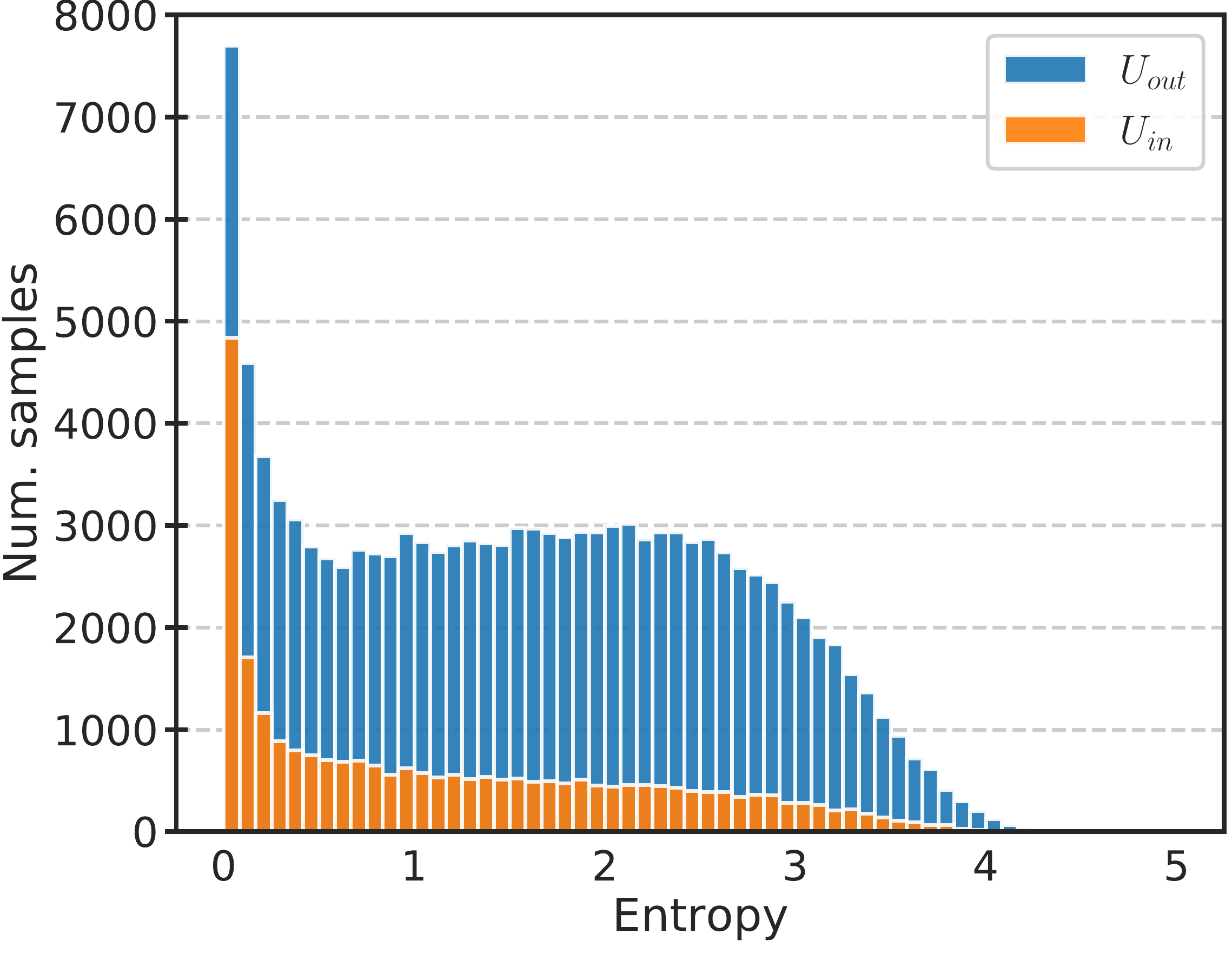}
    \label{fig:saves_ent_fm}
}\qquad
\subfloat[][FixMatch w/ DASO (Ours)]{
    \includegraphics[width=.32\linewidth]{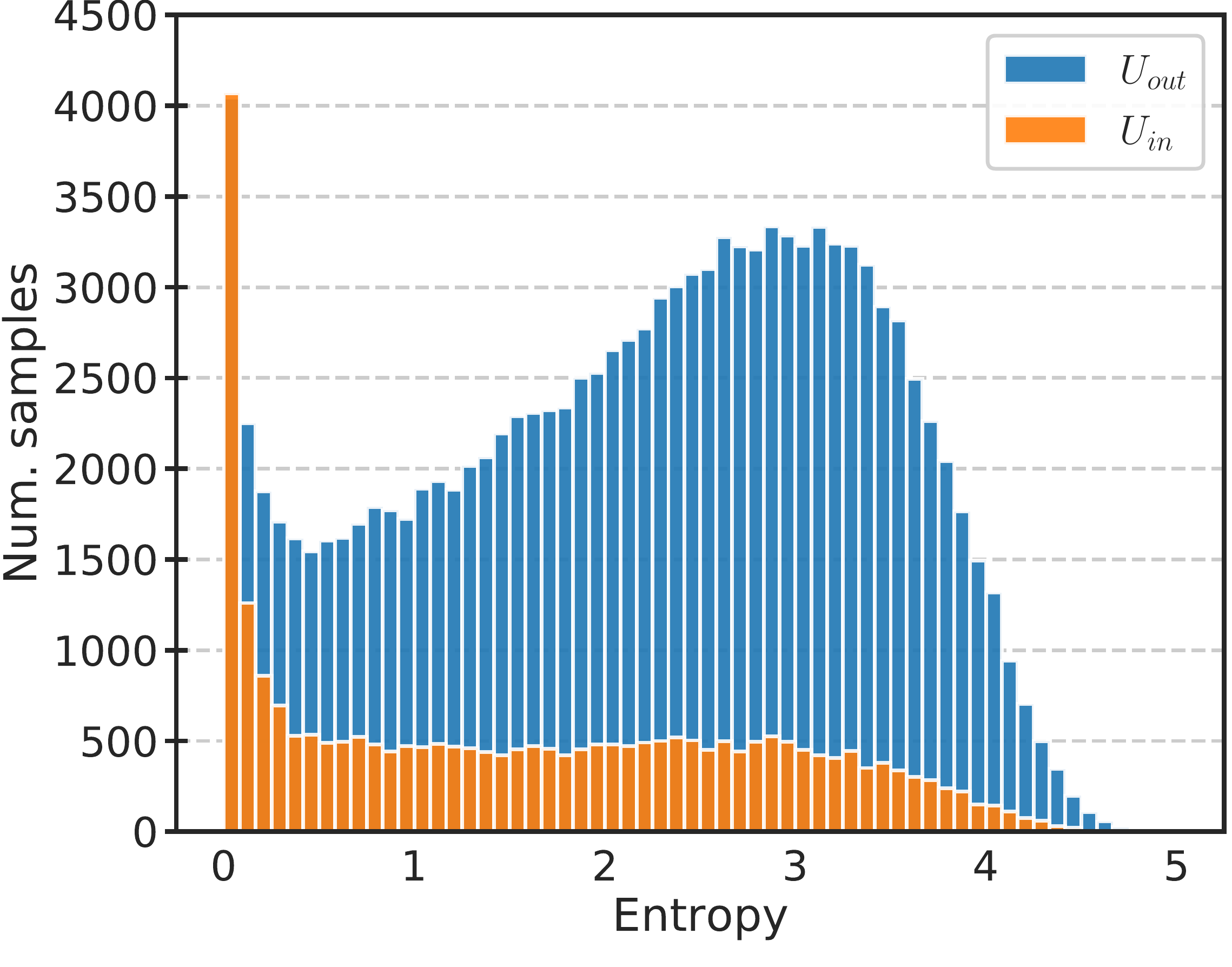}
    \label{fig:saves_ent_daso}
}
\caption{Comparisons of DASO and FixMatch~\cite{sohn2020fixmatch} on the distribution of entropy values from the predictions of samples in $\cU_{\text{in}}$ and $\cU_{\text{out}}$ of \emph{Semi-Aves} benchmark~\cite{su2021semisupervised}, respectively.
We observe that examples $\cU_{\text{in}}$ relatively remain in low-entropy (\eg, high-confidence) area, while those in $\cU_{\text{out}}$ are well pushed towards the high-entropy (\eg, low-confidence) area from DASO (ours).
}
\label{fig:analysis_saves_ent}
\end{figure}

%% file: supple/overall_framework.tex
\section{Overall Framework}
\begin{figure}[ht]
  \centering
   \includegraphics[width=0.9\linewidth]{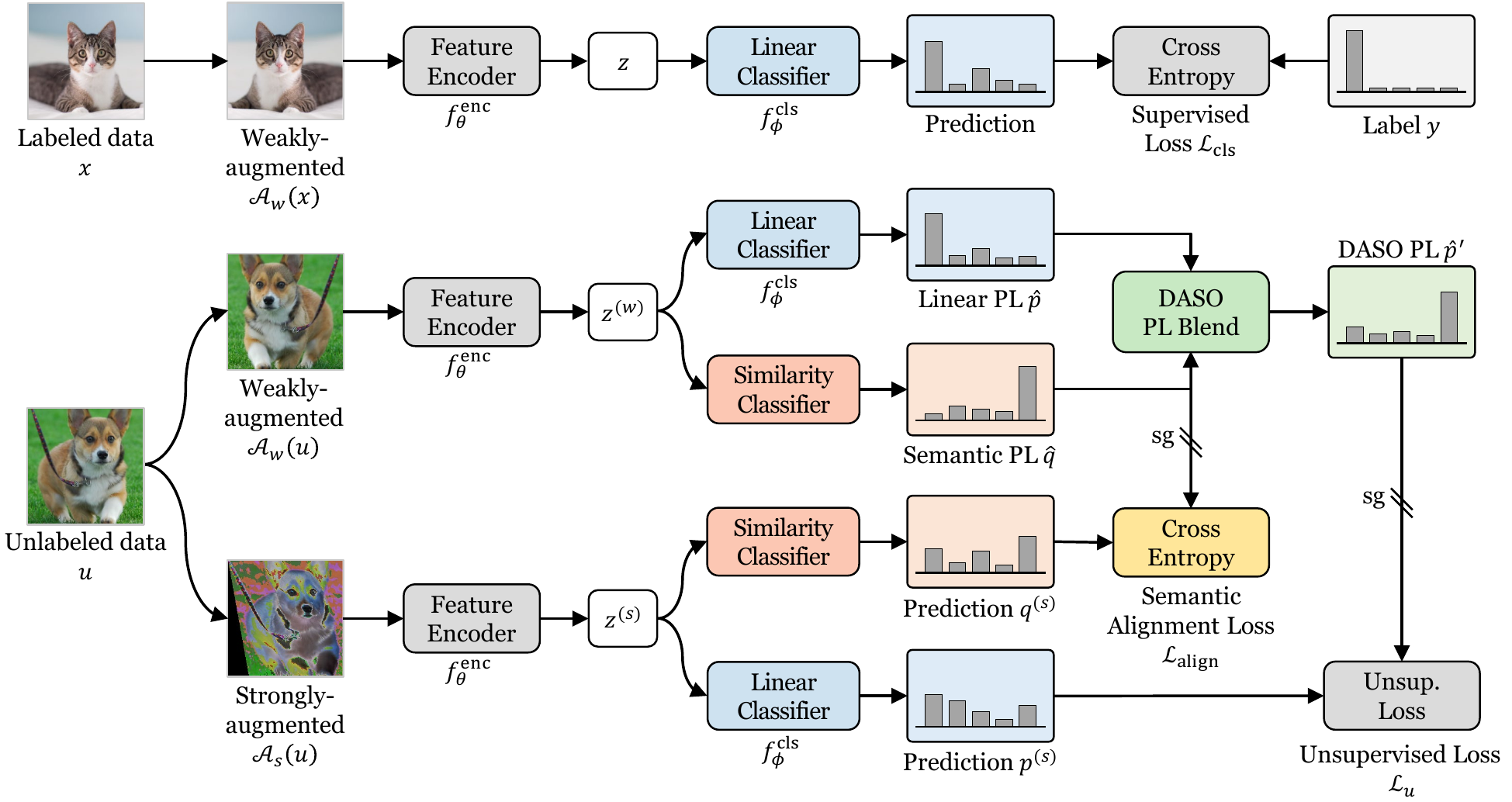}
   \caption{Overall framework of DASO including the blending of pseudo-labels (DASO PL Blend) and the semantic alignment loss ($\cL_{\text{align}}$).
   As explained in Sec.~{\color{red} 3.3} of the main paper, `balanced prototypes' for executing the similarity-based classifier are generated from EMA features of labeled data, which is omitted in this figure. 
   Two main components of DASO framework (blending of pseudo-labels and semantic alignment loss) can easily integrate with typical semi-supervised learning algorithms such as FixMatch~\cite{sohn2020fixmatch} and ReMixMatch~\cite{berthelot2020remixmatch} for debiasing pseudo-labels.
   Note that `sg' means stop-gradient operation. 
   }
   \label{fig:overall_framework}
\end{figure}

%% file: main_for_arxiv.bbl
\begin{thebibliography}{10}\itemsep=-1pt

\bibitem{ando2017deep}
Shin Ando and Chun~Yuan Huang.
\newblock Deep over-sampling framework for classifying imbalanced data.
\newblock In {\em Joint European Conference on Machine Learning and Knowledge
  Discovery in Databases}, pages 770--785, 2017.

\bibitem{arazo2020pseudo}
Eric Arazo, Diego Ortego, Paul Albert, Noel~E O’Connor, and Kevin McGuinness.
\newblock Pseudo-labeling and confirmation bias in deep semi-supervised
  learning.
\newblock In {\em International Joint Conference on Neural Networks (IJCNN)},
  pages 1--8, 2020.

\bibitem{bengio2015sharing}
Samy Bengio.
\newblock Sharing representations for long tail computer vision problems.
\newblock In {\em {ACM} International Conference on Multimodal Interaction},
  pages 1--1, 2015.

\bibitem{berthelot2020remixmatch}
David Berthelot, Nicholas Carlini, Ekin~D. Cubuk, Alex Kurakin, Kihyuk Sohn,
  Han Zhang, and Colin Raffel.
\newblock Remixmatch: Semi-supervised learning with distribution matching and
  augmentation anchoring.
\newblock In {\em International Conference on Learning Representations (ICLR)},
  2020.

\bibitem{berthelot2019mixmatch}
David Berthelot, Nicholas Carlini, Ian Goodfellow, Nicolas Papernot, Avital
  Oliver, and Colin~A Raffel.
\newblock Mixmatch: A holistic approach to semi-supervised learning.
\newblock In {\em Advances in Neural Information Processing Systems (NIPS)},
  volume~32, pages 5049--5059, 2019.

\bibitem{cao2019learning}
Kaidi Cao, Colin Wei, Adrien Gaidon, Nikos Arechiga, and Tengyu Ma.
\newblock Learning imbalanced datasets with label-distribution-aware margin
  loss.
\newblock In {\em Advances in Neural Information Processing Systems (NIPS)},
  volume~32, pages 1567--1578, 2019.

\bibitem{chapelle2009semi}
Olivier Chapelle, Bernhard Scholkopf, and Alexander Zien.
\newblock Semi-supervised learning.
\newblock {\em {IEEE} Transactions on Neural Networks}, 20(3):542--542, 2009.

\bibitem{chawla2002smote}
Nitesh~V Chawla, Kevin~W Bowyer, Lawrence~O Hall, and W~Philip Kegelmeyer.
\newblock Smote: synthetic minority over-sampling technique.
\newblock {\em Journal of artificial intelligence research}, 16:321--357, 2002.

\bibitem{chen2020semi}
Yanbei Chen, Xiatian Zhu, Wei Li, and Shaogang Gong.
\newblock Semi-supervised learning under class distribution mismatch.
\newblock In {\em AAAI Conference on Artificial Intelligence (AAAI)},
  volume~34, pages 3569--3576, 2020.

\bibitem{cho2021dealing}
Jae~Won Cho, Dong-Jin Kim, Jinsoo Choi, Yunjae Jung, and In~So Kweon.
\newblock Dealing with missing modalities in the visual question
  answer-difference prediction task through knowledge distillation.
\newblock In {\em Proceedings of the IEEE/CVF Conference on Computer Vision and
  Pattern Recognition}, 2021.

\bibitem{cho2021mcdal}
Jae~Won Cho, Dong-Jin Kim, Yunjae Jung, and In~So Kweon.
\newblock Mcdal: Maximum classifier discrepancy for active learning.
\newblock {\em IEEE transactions on neural networks and learning systems},
  2022.

\bibitem{coates2011analysis}
Adam Coates, Andrew Ng, and Honglak Lee.
\newblock An analysis of single-layer networks in unsupervised feature
  learning.
\newblock In {\em International Conference on Artificial Intelligence and
  Statistics (AISTATS)}, volume~15, pages 215--223, 2011.

\bibitem{cubuk2020randaugment}
Ekin~D Cubuk, Barret Zoph, Jonathon Shlens, and Quoc~V Le.
\newblock Randaugment: Practical automated data augmentation with a reduced
  search space.
\newblock In {\em Advances in Neural Information Processing Systems (NIPS)},
  2020.

\bibitem{cui2019class}
Yin Cui, Menglin Jia, Tsung-Yi Lin, Yang Song, and Serge Belongie.
\newblock Class-balanced loss based on effective number of samples.
\newblock In {\em {IEEE} Conference on Computer Vision and Pattern Recognition
  (CVPR)}, pages 9268--9277, 2019.

\bibitem{datta1997symbolic}
Piew Datta and Dennis Kibler.
\newblock Symbolic nearest mean classifiers.
\newblock In {\em AAAI Conference on Artificial Intelligence (AAAI)}, pages
  82--87, 1997.

\bibitem{deng2009imagenet}
Jia Deng, Wei Dong, Richard Socher, Li-Jia Li, Kai Li, and Li Fei-Fei.
\newblock Imagenet: A large-scale hierarchical image database.
\newblock In {\em {IEEE} Conference on Computer Vision and Pattern Recognition
  (CVPR)}, pages 248--255, 2009.

\bibitem{devries2017improved}
Terrance DeVries and Graham~W Taylor.
\newblock Improved regularization of convolutional neural networks with cutout.
\newblock {\em arXiv preprint arXiv:1708.04552}, 2017.

\bibitem{dong2018imbalanced}
Qi Dong, Shaogang Gong, and Xiatian Zhu.
\newblock Imbalanced deep learning by minority class incremental rectification.
\newblock {\em {IEEE} Transactions on Pattern Analysis and Machine Intelligence
  (TPAMI)}, 41(6):1367--1381, 2018.

\bibitem{grandvalet2005semi}
Yves Grandvalet and Yoshua Bengio.
\newblock Semi-supervised learning by entropy minimization.
\newblock In {\em Advances in Neural Information Processing Systems (NIPS)},
  volume~17, pages 281--296, 2005.

\bibitem{guo2020safe}
Lan-Zhe Guo, Zhen-Yu Zhang, Yuan Jiang, Yu-Feng Li, and Zhi-Hua Zhou.
\newblock Safe deep semi-supervised learning for unseen-class unlabeled data.
\newblock In {\em International Conference on Machine Learning (ICML)}, pages
  3897--3906, 2020.

\bibitem{gupta2019lvis}
Agrim Gupta, Piotr Dollar, and Ross Girshick.
\newblock Lvis: A dataset for large vocabulary instance segmentation.
\newblock In {\em {IEEE} Conference on Computer Vision and Pattern Recognition
  (CVPR)}, pages 5356--5364, 2019.

\bibitem{han2020unsupervised}
Tao Han, Junyu Gao, Yuan Yuan, and Qi Wang.
\newblock Unsupervised semantic aggregation and deformable template matching
  for semi-supervised learning.
\newblock In {\em Advances in Neural Information Processing Systems (NIPS)},
  volume~33, pages 9972--9982, 2020.

\bibitem{he2020momentum}
Kaiming He, Haoqi Fan, Yuxin Wu, Saining Xie, and Ross Girshick.
\newblock Momentum contrast for unsupervised visual representation learning.
\newblock In {\em {IEEE} Conference on Computer Vision and Pattern Recognition
  (CVPR)}, pages 9729--9738, 2020.

\bibitem{he2016deep}
Kaiming He, Xiangyu Zhang, Shaoqing Ren, and Jian Sun.
\newblock Deep residual learning for image recognition.
\newblock In {\em {IEEE} Conference on Computer Vision and Pattern Recognition
  (CVPR)}, pages 770--778, 2016.

\bibitem{hong2021disentangling}
Youngkyu Hong, Seungju Han, Kwanghee Choi, Seokjun Seo, Beomsu Kim, and Buru
  Chang.
\newblock Disentangling label distribution for long-tailed visual recognition.
\newblock In {\em {IEEE} Conference on Computer Vision and Pattern Recognition
  (CVPR)}, pages 6626--6636, June 2021.

\bibitem{hyun2020class}
Minsung Hyun, Jisoo Jeong, and Nojun Kwak.
\newblock Class-imbalanced semi-supervised learning.
\newblock {\em arXiv preprint arXiv:2002.06815}, 2020.

\bibitem{Kang2020Decoupling}
Bingyi Kang, Saining Xie, Marcus Rohrbach, Zhicheng Yan, Albert Gordo, Jiashi
  Feng, and Yannis Kalantidis.
\newblock Decoupling representation and classifier for long-tailed recognition.
\newblock In {\em International Conference on Learning Representations (ICLR)},
  2020.

\bibitem{kim2021single}
Dong-Jin Kim, Jae~Won Cho, Jinsoo Choi, Yunjae Jung, and In~So Kweon.
\newblock Single-modal entropy based active learning for visual question
  answering.
\newblock In {\em British Machine Vision Conference (BMVC)}, 2021.

\bibitem{kim2019image}
Dong-Jin Kim, Jinsoo Choi, Tae-Hyun Oh, and In~So Kweon.
\newblock Image captioning with very scarce supervised data: Adversarial
  semi-supervised learning approach.
\newblock In {\em Conference on Empirical Methods in Natural Language
  Processing (EMNLP)}, 2019.

\bibitem{kim2018disjoint}
Dong-Jin Kim, Jinsoo Choi, Tae-Hyun Oh, Youngjin Yoon, and In~So Kweon.
\newblock Disjoint multi-task learning between heterogeneous human-centric
  tasks.
\newblock In {\em IEEE Winter Conference on Applications of Computer Vision
  (WACV)}. IEEE, 2018.

\bibitem{kim2020detecting}
Dong-Jin Kim, Xiao Sun, Jinsoo Choi, Stephen Lin, and In~So Kweon.
\newblock Detecting human-object interactions with action co-occurrence priors.
\newblock In {\em European Conference on Computer Vision (ECCV)}, 2020.

\bibitem{kim2021acp++}
Dong-Jin Kim, Xiao Sun, Jinsoo Choi, Stephen Lin, and In~So Kweon.
\newblock Acp++: Action co-occurrence priors for human-object interaction
  detection.
\newblock {\em {IEEE} Transactions on Image Processing (TIP)}, 30:9150--9163,
  2021.

\bibitem{kim2020distribution}
Jaehyung Kim, Youngbum Hur, Sejun Park, Eunho Yang, Sung~Ju Hwang, and Jinwoo
  Shin.
\newblock Distribution aligning refinery of pseudo-label for imbalanced
  semi-supervised learning.
\newblock In {\em Advances in Neural Information Processing Systems (NIPS)},
  2020.

\bibitem{kim2020m2m}
Jaehyung Kim, Jongheon Jeong, and Jinwoo Shin.
\newblock M2m: Imbalanced classification via major-to-minor translation.
\newblock In {\em {IEEE} Conference on Computer Vision and Pattern Recognition
  (CVPR)}, pages 13896--13905, 2020.

\bibitem{krizhevsky2009learning}
Alex Krizhevsky, Geoffrey Hinton, et~al.
\newblock Learning multiple layers of features from tiny images.
\newblock {\em Technical report}, 2009.

\bibitem{kuo2020featmatch}
Chia{-}Wen Kuo, Chih{-}Yao Ma, Jia{-}Bin Huang, and Zsolt Kira.
\newblock Featmatch: Feature-based augmentation for semi-supervised learning.
\newblock In {\em European Conference on Computer Vision (ECCV)}, volume~18,
  pages 479--495, 2020.

\bibitem{laine2016temporal}
Samuli Laine and Timo Aila.
\newblock Temporal ensembling for semi-supervised learning.
\newblock In {\em International Conference on Learning Representations (ICLR)},
  2016.

\bibitem{lee2013pseudo}
Dong{-}Hyun Lee.
\newblock Pseudo-label: The simple and efficient semi-supervised learning
  method for deep neural networks.
\newblock In {\em Workshop on challenges in representation learning, ICML},
  2013.

\bibitem{lee2021auxiliary}
Hyuck Lee, Seungjae Shin, and Heeyoung Kim.
\newblock Abc: Auxiliary balanced classifier for class-imbalanced
  semi-supervised learning.
\newblock {\em arXiv preprint arXiv:2110.10368}, 2021.

\bibitem{li2020comatch}
Junnan Li, Caiming Xiong, and Steven Hoi.
\newblock Comatch: Semi-supervised learning with contrastive graph
  regularization.
\newblock In {\em {IEEE} International Conference on Computer Vision (ICCV)},
  2021.

\bibitem{li2021mopro}
Junnan Li, Caiming Xiong, and Steven Hoi.
\newblock Mopro: Webly supervised learning with momentum prototypes.
\newblock In {\em International Conference on Learning Representations (ICLR)},
  2021.

\bibitem{liu2020selectnet}
Yunru Liu, Tingran Gao, and Haizhao Yang.
\newblock Selectnet: Learning to sample from the wild for imbalanced data
  training.
\newblock In {\em Mathematical and Scientific Machine Learning}, volume 107,
  pages 193--206, 2020.

\bibitem{menon2021longtail}
Aditya~Krishna Menon, Sadeep Jayasumana, Ankit~Singh Rawat, Himanshu Jain,
  Andreas Veit, and Sanjiv Kumar.
\newblock Long-tail learning via logit adjustment.
\newblock In {\em International Conference on Learning Representations (ICLR)},
  2021.

\bibitem{miyato2018virtual}
Takeru Miyato, Shin-ichi Maeda, Masanori Koyama, and Shin Ishii.
\newblock Virtual adversarial training: a regularization method for supervised
  and semi-supervised learning.
\newblock {\em {IEEE} Transactions on Pattern Analysis and Machine Intelligence
  (TPAMI)}, 41(8):1979--1993, 2018.

\bibitem{oliver2018realistic}
Avital Oliver, Augustus Odena, Colin~A Raffel, Ekin~Dogus Cubuk, and Ian
  Goodfellow.
\newblock Realistic evaluation of deep semi-supervised learning algorithms.
\newblock In {\em Advances in Neural Information Processing Systems (NIPS)},
  volume~31, pages 3235--3246, 2018.

\bibitem{park2021opencos}
Jongjin Park, Sukmin Yun, Jongheon Jeong, and Jinwoo Shin.
\newblock Opencos: Contrastive semi-supervised learning for handling open-set
  unlabeled data.
\newblock {\em arXiv preprint arXiv:2107.08943}, 2021.

\bibitem{park2021influence}
Seulki Park, Jongin Lim, Younghan Jeon, and Jin~Young Choi.
\newblock Influence-balanced loss for imbalanced visual classification.
\newblock In {\em {IEEE} International Conference on Computer Vision (ICCV)},
  pages 735--744, 2021.

\bibitem{paszke2019pytorch}
Adam Paszke, Sam Gross, Francisco Massa, Adam Lerer, James Bradbury, Gregory
  Chanan, Trevor Killeen, Zeming Lin, Natalia Gimelshein, Luca Antiga, Alban
  Desmaison, Andreas Kopf, Edward Yang, Zachary DeVito, Martin Raison, Alykhan
  Tejani, Sasank Chilamkurthy, Benoit Steiner, Lu Fang, Junjie Bai, and Soumith
  Chintala.
\newblock Pytorch: An imperative style, high-performance deep learning library.
\newblock In {\em Advances in Neural Information Processing Systems (NIPS)},
  volume~32, page 8026–8037, 2019.

\bibitem{pham2020meta}
Hieu Pham, Qizhe Xie, Zihang Dai, and Quoc~V Le.
\newblock Meta pseudo labels.
\newblock In {\em {IEEE} Conference on Computer Vision and Pattern Recognition
  (CVPR)}, pages 11557--11568, 2021.

\bibitem{rebuffi2017icarl}
Sylvestre-Alvise Rebuffi, Alexander Kolesnikov, Georg Sperl, and Christoph~H
  Lampert.
\newblock icarl: Incremental classifier and representation learning.
\newblock In {\em Proceedings of the IEEE conference on Computer Vision and
  Pattern Recognition}, pages 2001--2010, 2017.

\bibitem{ren2020balanced}
Jiawei Ren, Cunjun Yu, shunan sheng, Xiao Ma, Haiyu Zhao, Shuai Yi, and
  hongsheng Li.
\newblock Balanced meta-softmax for long-tailed visual recognition.
\newblock In {\em Advances in Neural Information Processing Systems (NIPS)},
  volume~33, pages 4175--4186, 2020.

\bibitem{ren2018learning}
Mengye Ren, Wenyuan Zeng, Bin Yang, and Raquel Urtasun.
\newblock Learning to reweight examples for robust deep learning.
\newblock In {\em International Conference on Machine Learning (ICML)}, pages
  4334--4343. PMLR, 2018.

\bibitem{ren2020not}
Zhongzheng Ren, Raymond Yeh, and Alexander Schwing.
\newblock Not all unlabeled data are equal: Learning to weight data in
  semi-supervised learning.
\newblock In {\em Advances in Neural Information Processing Systems (NIPS)},
  volume~33, pages 21786--21797, 2020.

\bibitem{salakhutdinov2007learning}
Ruslan Salakhutdinov and Geoff Hinton.
\newblock Learning a nonlinear embedding by preserving class neighbourhood
  structure.
\newblock In {\em Artificial Intelligence and Statistics}, pages 412--419.
  PMLR, 2007.

\bibitem{shin2021labor}
Inkyu Shin, Dong-Jin Kim, Jae~Won Cho, Sanghyun Woo, KwanYong Park, and In~So
  Kweon.
\newblock Labor: Labeling only if required for domain adaptive semantic
  segmentation.
\newblock In {\em {IEEE} International Conference on Computer Vision (ICCV)},
  2021.

\bibitem{smith2020building}
Leslie~N Smith and Adam Conovaloff.
\newblock Building one-shot semi-supervised (boss) learning up to fully
  supervised performance.
\newblock {\em arXiv preprint arXiv:2006.09363}, 2020.

\bibitem{snell2017prototypical}
Jake Snell, Kevin Swersky, and Richard Zemel.
\newblock Prototypical networks for few-shot learning.
\newblock In {\em Advances in Neural Information Processing Systems (NIPS)},
  volume~30, pages 4077--4087, 2017.

\bibitem{sohn2020fixmatch}
Kihyuk Sohn, David Berthelot, Chun{-}Liang Li, Zizhao Zhang, Nicholas Carlini,
  Ekin~D Cubuk, Alex Kurakin, Han Zhang, and Colin Raffel.
\newblock Fixmatch: Simplifying semi-supervised learning with consistency and
  confidence.
\newblock In {\em Advances in Neural Information Processing Systems (NIPS)},
  2020.

\bibitem{su2021realistic}
Jong-Chyi Su, Zezhou Cheng, and Subhransu Maji.
\newblock A realistic evaluation of semi-supervised learning for fine-grained
  classification.
\newblock In {\em {IEEE} Conference on Computer Vision and Pattern Recognition
  (CVPR)}, 2021.

\bibitem{su2021semisupervised}
Jong-Chyi Su and Subhransu Maji.
\newblock The semi-supervised inaturalist-aves challenge at fgvc7 workshop,
  2021.

\bibitem{tarvainen2017mean}
Antti Tarvainen and Harri Valpola.
\newblock Mean teachers are better role models: Weight-averaged consistency
  targets improve semi-supervised deep learning results.
\newblock In {\em Advances in Neural Information Processing Systems (NIPS)},
  volume~30, pages 1195--1204, 2017.

\bibitem{van2008visualizing}
Laurens Van~der Maaten and Geoffrey Hinton.
\newblock Visualizing data using t-sne.
\newblock {\em Journal of machine learning research}, 9(11), 2008.

\bibitem{van2018inaturalist}
Grant Van~Horn, Oisin Mac~Aodha, Yang Song, Yin Cui, Chen Sun, Alex Shepard,
  Hartwig Adam, Pietro Perona, and Serge Belongie.
\newblock The inaturalist species classification and detection dataset.
\newblock In {\em {IEEE} Conference on Computer Vision and Pattern Recognition
  (CVPR)}, pages 8769--8778, 2018.

\bibitem{wang2021longtailed}
Xudong Wang, Long Lian, Zhongqi Miao, Ziwei Liu, and Stella Yu.
\newblock Long-tailed recognition by routing diverse distribution-aware
  experts.
\newblock In {\em International Conference on Learning Representations (ICLR)},
  2021.

\bibitem{wang2022debiased}
Xudong Wang, Zhirong Wu, Long Lian, and Stella~X Yu.
\newblock Debiased learning from naturally imbalanced pseudo-labels for
  zero-shot and semi-supervised learning.
\newblock {\em arXiv preprint arXiv:2201.01490}, 2022.

\bibitem{wei2021crest}
Chen Wei, Kihyuk Sohn, Clayton Mellina, Alan Yuille, and Fan Yang.
\newblock Crest: A class-rebalancing self-training framework for imbalanced
  semi-supervised learning.
\newblock In {\em {IEEE} Conference on Computer Vision and Pattern Recognition
  (CVPR)}, 2021.

\bibitem{xiang2020learning}
Liuyu Xiang, Guiguang Ding, and Jungong Han.
\newblock Learning from multiple experts: Self-paced knowledge distillation for
  long-tailed classification.
\newblock In {\em European Conference on Computer Vision (ECCV)}, pages
  247--263, 2020.

\bibitem{xie2019unsupervised}
Qizhe Xie, Zihang Dai, Eduard Hovy, Minh{-}Thang Luong, and Quoc~V Le.
\newblock Unsupervised data augmentation for consistency training.
\newblock In {\em Advances in Neural Information Processing Systems (NIPS)},
  2020.

\bibitem{xie2020self}
Qizhe Xie, Minh-Thang Luong, Eduard Hovy, and Quoc~V Le.
\newblock Self-training with noisy student improves imagenet classification.
\newblock In {\em {IEEE} Conference on Computer Vision and Pattern Recognition
  (CVPR)}, pages 10687--10698, 2020.

\bibitem{yalniz2019billion}
I~Zeki Yalniz, Herv{\'e} J{\'e}gou, Kan Chen, Manohar Paluri, and Dhruv
  Mahajan.
\newblock Billion-scale semi-supervised learning for image classification.
\newblock {\em arXiv preprint arXiv:1905.00546}, 2019.

\bibitem{yang2020rethinking}
Yuzhe Yang and Zhi Xu.
\newblock Rethinking the value of labels for improving class-imbalanced
  learning.
\newblock In {\em Advances in Neural Information Processing Systems (NIPS)},
  2020.

\bibitem{zagoruyko2016wider}
Sergey Zagoruyko and Nikos Komodakis.
\newblock Wide residual networks.
\newblock In {\em British Machine Vision Conference (BMVC)}, pages 87.1--87.12,
  2016.

\bibitem{zhang2018mixup}
Hongyi Zhang, Moustapha Cisse, Yann~N. Dauphin, and David Lopez{-}Paz.
\newblock mixup: Beyond empirical risk minimization.
\newblock In {\em International Conference on Learning Representations (ICLR)},
  2018.

\bibitem{zhou2020bbn}
Boyan Zhou, Quan Cui, Xiu-Shen Wei, and Zhao-Min Chen.
\newblock Bbn: Bilateral-branch network with cumulative learning for
  long-tailed visual recognition.
\newblock In {\em {IEEE} Conference on Computer Vision and Pattern Recognition
  (CVPR)}, pages 9719--9728, 2020.

\end{thebibliography}
